\documentclass[twoside]{article}

\usepackage[accepted]{aistats2026}
\usepackage[round]{natbib}

\usepackage{url}

\usepackage{graphicx}
\usepackage{wrapfig}
\usepackage{algorithm}
\usepackage{algpseudocode}
\usepackage{multirow}
\usepackage{soul}
\usepackage{amsmath}
\usepackage{diagbox}
\usepackage{booktabs}
\usepackage{bm}
\usepackage{amsthm}
\usepackage{amssymb}   
\usepackage{xcolor}
\usepackage{subcaption}
\usepackage{subfig}
\usepackage{flushend}

\newcommand{\boldhdr}[1]{\vspace{0.05cm}\noindent \textbf{#1.}}

\usepackage{enumitem}

\newcolumntype{?}{!{\vrule width 1pt}}
\theoremstyle{plain}

\newtheorem{lemma}{Lemma}[section]
\theoremstyle{remark}
\newtheorem{remark}{Remark}

\newcommand{\cvec}[1]{\boldsymbol{#1}}
\newcommand{\dotp}[2]{\langle #1, #2 \rangle}

\runningtitle{Understanding SAM’s Robustness to Noisy Labels through Gradient Down-weighting}

\begin{document}

%

%

\twocolumn[

\aistatstitle{Understanding SAM’s Robustness to Noisy Labels \\ through Gradient Down-weighting}

\aistatsauthor{ Hoang-Chau Luong$^{1,2\dagger}$ \And 
Quang-Thuc Nguyen$^{2\dagger}$ \And 
Dat Ba Tran$^{3}$ \And 
Minh-Triet Tran$^{2}$}

\aistatsaddress{ $^{1}$ Rochester Institute of Technology, Rochester, NY, USA \\
$^{2}$ University of Science - VNUHCM, Vietnam \\
$^{3}$ Rowan University, Glassboro, NJ, USA \\
$^{\dagger}$ These authors contributed equally}
]

\begin{abstract}
    Sharpness-Aware Minimization (SAM) was introduced to improve generalization by seeking flat minima, yet it also exhibits robustness to label noise, a phenomenon that remains only partially understood. Prior work has mainly attributed this effect to SAM’s tendency to prolong the learning of clean samples. In this work, we provide a complementary explanation by analyzing SAM at the element-wise level. We show that when noisy gradients dominate a parameter direction, their influence is reduced by the stronger amplification of clean gradients. This slows the memorization of noisy labels while sustaining clean learning, offering a more complete account of SAM’s robustness. Building on this insight, we propose SANER (Sharpness-Aware Noise-Explicit Reweighting), a simple variant of SAM that explicitly magnifies this down-weighting effect. Experiments on benchmark image classification tasks with noisy labels demonstrate that SANER significantly mitigates noisy-label memorization and improves generalization over both SAM and SGD. Moreover, since SANER is designed from the mechanism of SAM, it can also be seamlessly integrated into SAM-like variants, further boosting their robustness.
\end{abstract}

\section{INTRODUCTION}

Robust learning under label noise is a fundamental challenge in deep learning, as real-world datasets often contain annotation errors that severely hinder generalization. Well-known datasets such as CIFAR-10N~\citep{wei2022learning}, CIFAR-100N~\citep{wei2022learning}, and WebVision~\citep{li2017webvision} highlight this issue, where large-scale human labeling introduces substantial noise. Over-parameterized deep neural networks (DNNs) are particularly prone to memorizing these noisy labels~\citep{zhang2021understanding}, which makes it harder to learn useful patterns and leads to poor performance on clean test data~\citep{jiang2018mentornet,Nguyen2020SELF}. Additionally, because manually verifying labels at scale is costly, it is essential to develop training algorithms that robust to noisy labels while still leveraging clean supervision to ensure strong generalization.

Among recent optimization methods, SAM~\citep{foret2021sharpnessaware} has attracted attention not only for its ability to seek flat minima and improve generalization but also for its robustness to noisy labels. Moreover, several variants of SAM, though not explicitly developed to address label noise, also exhibit notable robustness in noisy-label scenarios~\citep{kwon2021asam, zhuang2022surrogate, li2024friendly, li2024enhancing}. \cite{baek2024why} observed that when trained with SAM under label noise, test accuracy peaks midway through training and does not improve with further epochs, indicating that its robustness cannot be attributed solely to flat-minima effects. Understanding this robustness is crucial, as it both deepens our theoretical understanding of noise-robust optimization and guides the design of more effective algorithms for learning with noisy labels.

\begin{figure*}[t]
\centering
    \begin{subfigure}{0.3\linewidth}
        \centering
        \includegraphics[width=\linewidth, height=34mm]{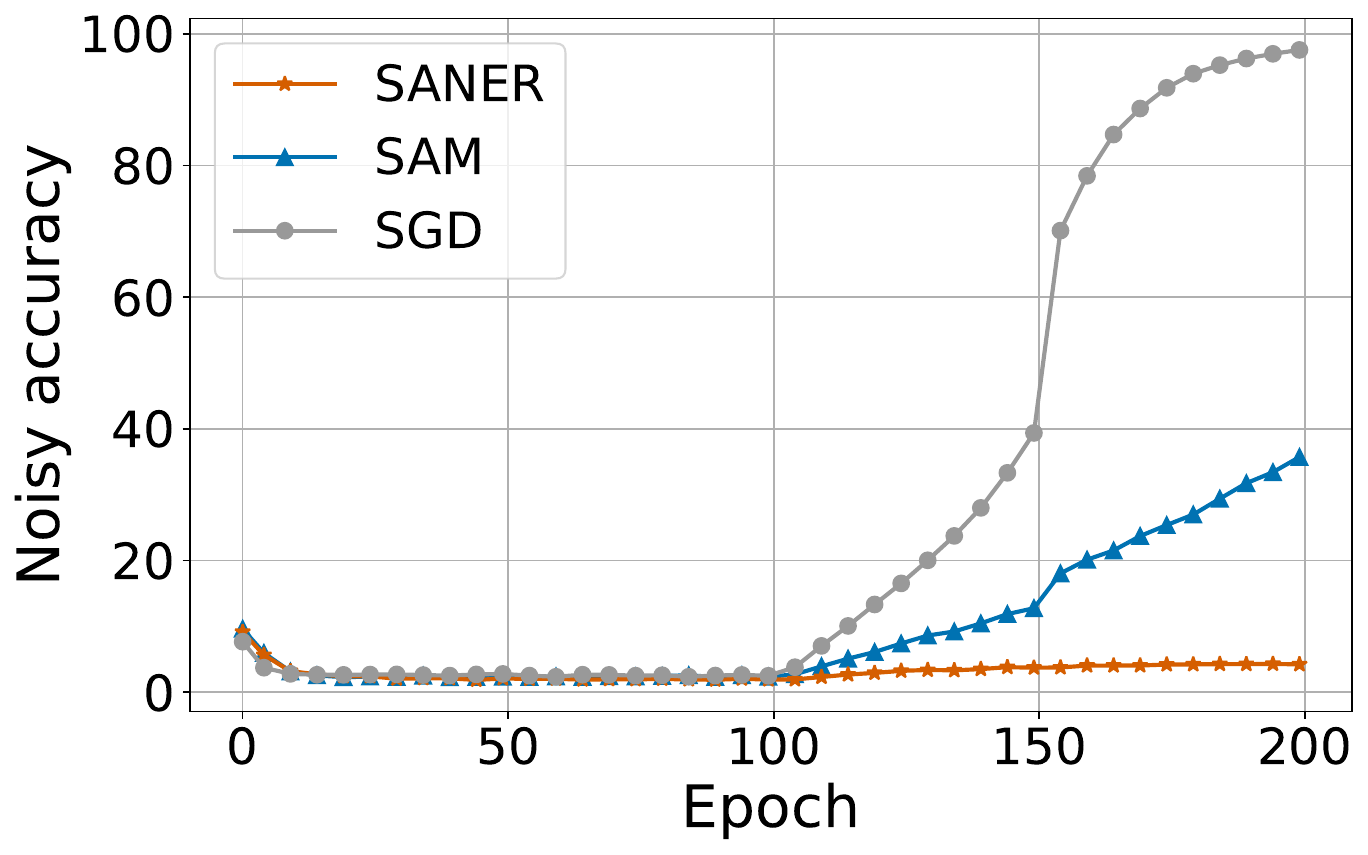}
        \caption{The noisy training accuracy}
        \label{fig:sam_noise_acc_025}
    \end{subfigure}
    \hfill
    \begin{subfigure}{0.3\linewidth}
        \centering
        \includegraphics[width=\linewidth, height=34mm]{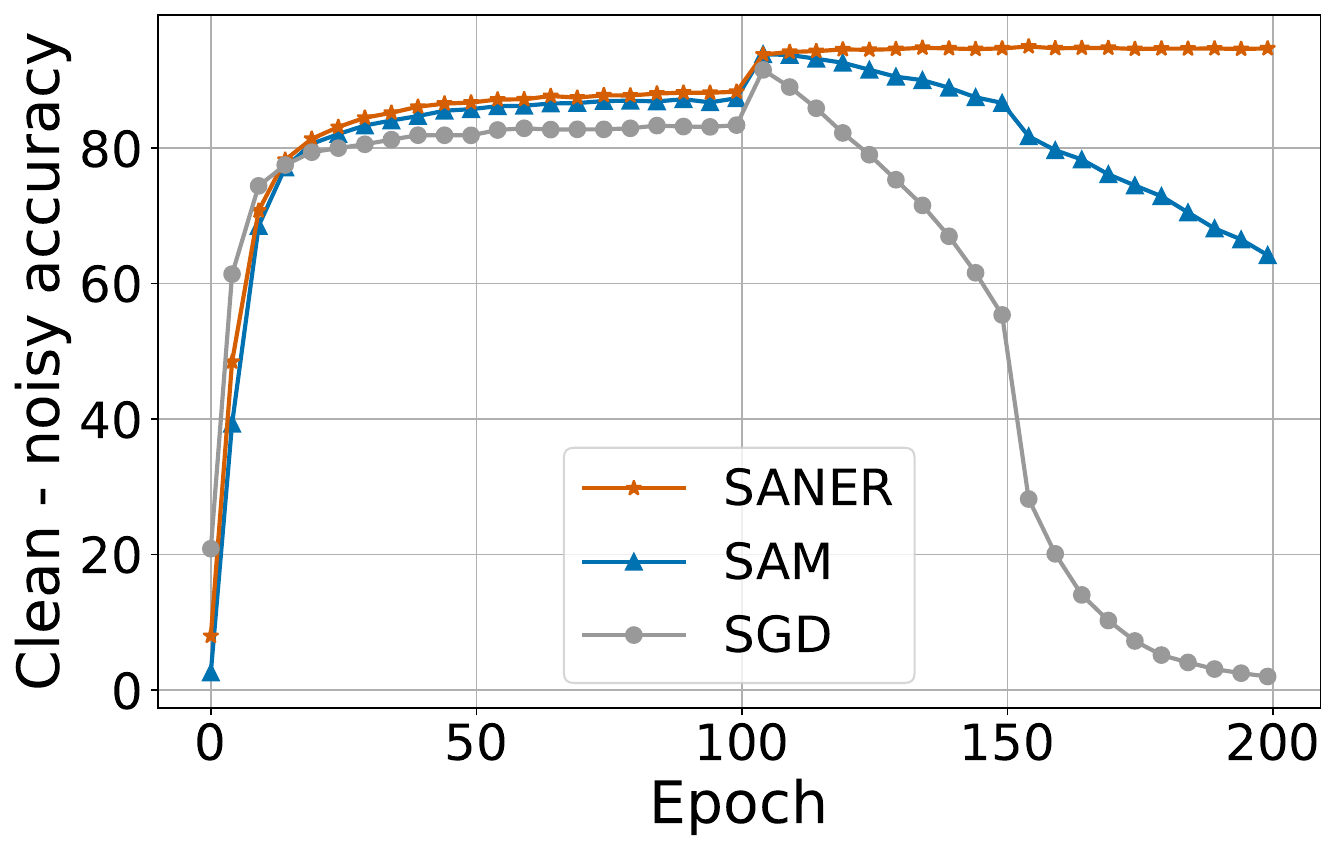}
        \caption{The gap between clean and noisy}
        \label{fig:gap_clean_noise_acc_025}
    \end{subfigure}
    \hfill
    \begin{subfigure}{0.3\linewidth}
        \centering
        \includegraphics[width=\linewidth, height=34mm]{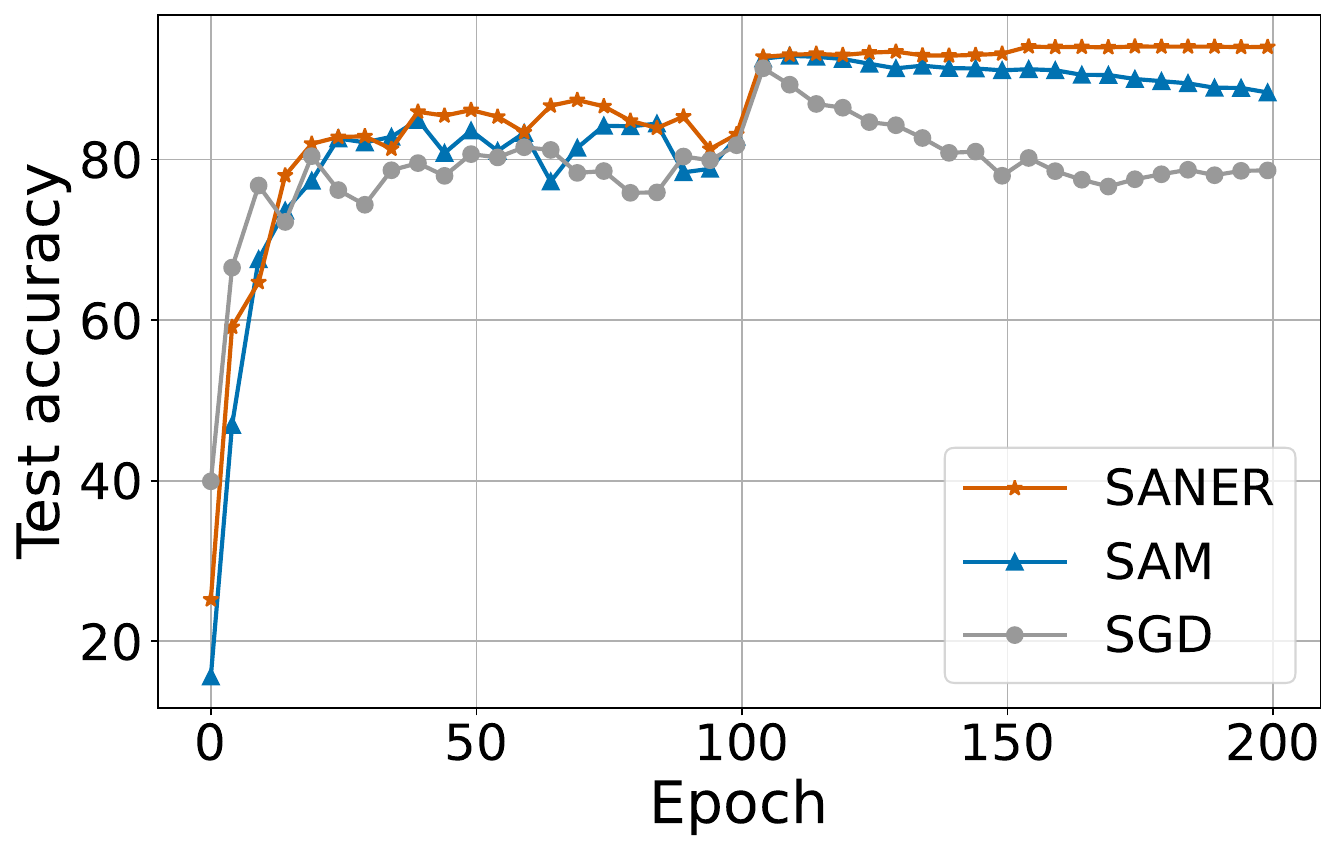}
        \caption{The test accuracy}
        \label{fig:test_acc_025}
    \end{subfigure}
    \caption{Performance comparison of SGD, SAM, and SANER (ours) on ResNet18 trained on CIFAR-10 with 25\% label noise. 
    Noise accuracy measures how much the model memorizes noisy labels, where lower values are better. 
    SAM reduces noisy label fitting (a) and increases the clean–noise accuracy gap (b), computed as clean accuracy minus noisy accuracy. 
    SANER further strengthens this effect, resulting in improved test accuracy (c).}
    \label{fig:sam_vs_sgd}
\end{figure*}

\cite{baek2024why} suggested that SAM's robustness to label noise arises primarily from its ability to prolong the learning of clean samples by up-weighting their gradients. While this explanation provides an important insight, it remains incomplete. In particular, SAM not only strengthens clean-sample gradients but also amplifies noisy-sample gradients, which raises the question of why it still achieves greater robustness than standard SGD. To address this gap, we analyze SAM at the element-wise gradient level and uncover characteristics that more fully explain its ability to slow down fitting to noisy labels.

Our theoretical analysis shows that during the transitional phase—when the model shifts from learning clean patterns to memorizing noisy labels—SAM down-weights element-wise gradients aligned with noisy supervision. This occurs because, when noisy gradients dominate a parameter direction, SAM rebalances the update. Particularly, it amplifies the contribution of clean gradients more strongly than that of noisy ones. This finding is significant because it complements SAM’s tendency to prolong clean learning, offering a more complete explanation of its robustness.

We further validate this mechanism in DNNs through a series of controlled experiments, which consistently support our theoretical findings. Motivated by these insights, we propose \textbf{SANER} (\textbf{S}harpness-\textbf{A}ware \textbf{N}oise-\textbf{E}xplicit \textbf{R}eweighting), a simple modification of SAM that magnifies this down-weighting phenomenon to more aggressively resist noisy-label memorization. As shown in Figure~\ref{fig:sam_vs_sgd}, SANER significantly reduces overfitting to noisy labels and achieves better generalization than both SAM and SGD.

In summary, our contributions are threefold:
\vspace{-1em}
\begin{itemize}[leftmargin=10pt]
    \item We provide a theoretical analysis showing that SAM down-weights specific gradient elements, thereby slowing the learning of noisy labels.
    \item We empirically validate this behavior in DNNs, showing that the down-weighted elements align closely with gradients induced by label noise, and that their removal degrades SAM's robustness.
    \item We introduce SANER, a simple modification of SAM that more strongly suppresses noisy gradient elements. Extensive experiments demonstrate that SANER consistently outperforms SAM across different datasets, noise types, and architectures. Moreover, SANER can be seamlessly integrated with other SAM-like optimizers, further improving their robustness to label noise.
\end{itemize}

\section{BACKGROUND \& RELATED WORKS}

\boldhdr{Sharpness-Aware Minimization}
SAM~\citep{foret2021sharpnessaware} was originally proposed to improve generalization by encouraging convergence to flat minima through a perturbation-based update. 
Since then, numerous variants have been developed and shown to improve performance in noisy-label settings~\citep{kwon2021asam, kim2022fisher, jiang2023an, li2024enhancing, li2024friendly}. 
SAM has also been studied from diverse perspectives, including implicit bias~\citep{andriushchenko2022towards}, training dynamics~\citep{bartlett2023dynamics}, Hessian regularization~\citep{wen2023how}, bias--variance trade-offs~\citep{behdin2023statistical}, and benign overfitting~\citep{chen2024does}. 
Its convergence has been analyzed in the framework of Inexact Gradient Descent~\citep{kmt23.1, kmt23.2}, where SAM’s perturbed gradient is treated as an approximation of the unperturbed gradient~\citep{khanh2024fundamental}. \cite{shin2023effects} showed that SAM benefits from over-parameterization under label noise, while \cite{baek2024why} linked its robustness to sample-wise gradients. We extend these analyses to the element-wise level, showing that SAM implicitly down-weights specific gradient components, thereby slowing noisy-label memorization.

\boldhdr{Label Noise}
Deep neural networks are prone to memorizing noisy labels, which can severely degrade generalization~\citep{zhang2021understanding}. 
To address this, a wide range of approaches have been developed. 
One line of work designs robust loss functions that are less sensitive to corrupted labels~\citep{zhang2018generalized, menon2020can, ma2020normalized, wei2023mitigating}. 
Another introduces sample reweighting or selection strategies to prioritize clean data and reduce the effect of noisy examples~\citep{liu2015classification, ren2018learning, jiang2018mentornet, wei2020metainfonet}. 
Additional methods employ regularization~\citep{tanaka2018joint, lukasik2020does, bai2021understanding, prog_noise_iclr2021, liu2022robust} or leverage semi-supervised, meta-learning, and self-supervised training paradigms~\citep{Nguyen2020SELF, Li2020DivideMix:, shu2019meta, li2022selective}. 
While effective, these techniques often depend on noise modeling or auxiliary procedures, which limit their scalability in practice. In contrast, our approach leverages the inherent behavior of the SAM optimizer to mitigate label-noise memorization.

\boldhdr{SAM algorithm} 
Let $f(\boldsymbol{x}_i; \boldsymbol{w})$ represent a mapping function from inputs $\boldsymbol{x_i} \in \mathbb{R}^d$ to outputs $\hat{y}_i \in \mathbb{R}$ which is parameterized by $\boldsymbol{w} \in \mathbb{R}^d$, and let $\ell(f(\boldsymbol{x}_i; \boldsymbol{w}), y_i)$ (shortened as $\ell_i(\boldsymbol{w})$) denote the loss function between the prediction $\hat{y}_i = f(\boldsymbol{x}_i; \boldsymbol{w})$ and the ground-truth label $y_i$.
To enhance generalization performance, SAM~\citep{foret2021sharpnessaware} is proposed to seek a flat minimum of the empirical training loss by minimizing the following robust objective:
\begin{equation}
    \min_{\boldsymbol{w}} \max_{ ||\boldsymbol{\epsilon}||_2 \leq \rho} L(\boldsymbol{w} + \boldsymbol{\epsilon}), \text{ where } 
    L(\boldsymbol{w}) = \frac{1}{n} \sum_{i=1}^n \ell_i(\boldsymbol{w}),
\end{equation}
where perturbation radius $\rho \in \mathbb{R}$ represents the strength of the adversarial weight perturbation $\boldsymbol{\epsilon} \in \mathbb{R}^d$. Intuitively, the objective seeks a robust solution such that within a neighbor region, the loss can remain stable under any $\boldsymbol{\epsilon}$-perturbation. SAM employs a first-order Taylor approximation of the loss to efficiently optimize this objective, which approximates the worst-case $\hat{\boldsymbol{\epsilon}}$ as follows 
\begin{equation}
\label{eq:sgd}
    \hat{\boldsymbol{\epsilon}} = \rho \frac{ \boldsymbol{g}^{\text{SGD}} } { || \boldsymbol{g}^{\text{SGD}} || } =  \rho \frac{ \nabla_{\boldsymbol{w}} L(\boldsymbol{w}) }{ || \nabla_{\boldsymbol{w}} L(\boldsymbol{w}) || }.
\end{equation}
SAM gradient is computed at the perturbed point $\boldsymbol{w} + \hat{\boldsymbol{\epsilon}}$, and the base optimizer (e.g., SGD) with a learning rate $\eta$ is used to update the model parameters:
\begin{equation}
\label{eq:sam_update}
    \boldsymbol{w} 
    = \boldsymbol{w} - \eta \boldsymbol{g}^{\text{SAM}}
    = \boldsymbol{w} - \eta \nabla_{\boldsymbol{w}} L(\boldsymbol{w}) \bigg|_{ \boldsymbol{w} + \hat{\boldsymbol{\epsilon}} }.
\end{equation}
This update steers the model parameters toward a perturbation-robust solution, requiring only one additional gradient computation per iteration.

\section{ANALYSIS ELEMENT-WISE GRADIENT BEHAVIOR OF SAM}
\label{sec:hypothesis_1}

\subsection{Problem Setup}
\label{sec:preliminaries}

We consider a binary classification setting. Given a training set of $n$ samples $[(\cvec{x}_i, y_i)]^{n}_{i=1}$, where $\cvec{x}_i \in \mathbb{R}^d$ is a feature vector and $y_i \in \{ 0, 1\}$ is the corresponding label, our objective is to learn model parameters $\cvec{w} \in \mathbb{R}^d$ that can give the correct prediction $\hat{y}_i = \sigma(\dotp{\cvec{w}}{\cvec{x}_i})$, where $\sigma(x) = 1/(1 + \exp(-x))$ by minimizing the empirical loss $L(\cvec{w}) = \frac{1}{n} \sum^{n}_{i=1} \ell(\cvec{w}, \cvec{x}_i, y_i)$. The binary cross-entropy loss is defined as
\begin{align}
    \ell(\cvec{w}, \cvec{x}_i, y_i) 
    &= H(y_i, \hat{y}_i) \\
    &= - y_i \log \hat{y}_i - (1 - y) \log (1 - \hat{y}_i).
\end{align}
To introduce label noise, we randomly flip the labels of a subset of training samples (e.g., from 0 to 1 or 1 to 0).

\boldhdr{SGD gradient} 
The gradient of the loss with respect to $\cvec{w}$ for a single sample $(\cvec{x}_i, y_i)$, denoted as $\nabla_{\cvec{w}} \ell(\cvec{w}; \cvec{x}_i, y_i)$, is commonly referred to as a sample-wise gradient. It is the update direction used in standard SGD and is computed as:
\begin{equation}
\label{eq:1_sgd}
    \nabla_{\cvec{w}} \ell(\cvec{w}, \cvec{x}_i, y_i) = \Big(\sigma \big( \dotp{\cvec{w}}{\cvec{x}_i} \big) - y_i \Big)\cvec{x}_i.
\end{equation}
\boldhdr{1-SAM gradient} 
We focus on 1-SAM variant of SAM, which computes an adversarial perturbation $\cvec{\epsilon}_i$ for each individual training sample rather than over the mini-batch. This variant has been shown to yield strong performance~\citep{foret2021sharpnessaware} and is used for theoretical analysis in~\cite{baek2024why}. The 1-SAM sample-wise gradient for the $i$-th sample is defined as
\begin{equation}
\label{eq:1_sam}
    \nabla_{\cvec{w}} \ell(\cvec{w} + \cvec{\epsilon}_i, \cvec{x}_i, y_i)  = \Big(\sigma \big( \dotp{\cvec{w} + \cvec{\epsilon}_i}{\cvec{x}_i} \big) - y_i \Big) \cvec{x}_i,
\end{equation}
where the adversarial perturbation is
\begin{equation}
    \cvec{\epsilon}_i = \rho  \frac{ \big( \sigma ( \dotp{\cvec{w}}{\cvec{x}_i}) - y_i \big)\cvec{x}_i }{ \big\| \big( \sigma ( \dotp{\cvec{w}}{\cvec{x}_i}  ) - y_i \big)\cvec{x}_i \big\| } = \rho (-1)^{y_i} \frac{\cvec{x}_i}{\|\cvec{x}_i\|},
\end{equation}
where the last equality follows from the fact that $ \sigma \big( \dotp{\cvec{w}}{\cvec{x}_i} \big) - y_i > 0$ if $y_i = 0$, and $ \sigma \big( \dotp{\cvec{w}}{\cvec{x}_i} \big) - y_i < 0$ if $y_i = 1$.

\subsection{Theoretical Analysis in Linear Models}
\label{sec:theory}

We theoretically show that SAM selectively down-weights specific element-wise gradients compared to standard SGD to mitigate noisy-label memorization.

\boldhdr{Down-weighted gradient elements} 
We consider a mini-batch consisting of two samples from the same true class, where one is correctly labeled and the other is corrupted due to label noise. Without loss of generality, let the clean sample be \( (\cvec{x}_c, y_c) \) with \( y_c = 0 \), and the noisy sample be \( (\cvec{x}_\eta, y_\eta) \) with \( y_\eta = 1 \). Because both samples are drawn from the same underlying class, we assume there exists at least one shared feature dimension \( j \) such that \( x_c^j = x_\eta^j \). We define the ratio between the 1-SAM and SGD gradients at dimension \( j \) as $g^{\rm SAM} / g^{\rm SGD}$, or equivalently
\begin{align*}
    r^j &= \frac{\nabla_{\cvec{w}} \ell(\cvec{w} + \cvec{\epsilon}_c, \cvec{x}_c, y_c) + \nabla_{\cvec{w}} \ell(\cvec{w} + \cvec{\epsilon}_\eta, \cvec{x}_\eta, y_\eta)}{\nabla_{\cvec{w}} \ell(\cvec{w}, \cvec{x}_c, y_c) + \nabla_{\cvec{w}} \ell(\cvec{w}, \cvec{x}_\eta, y_\eta)} \\
    &= \frac{ 1 - \sigma\big( \dotp{\cvec{w}}{\cvec{x}_{c}} + \rho \|\cvec{x}_{c}\| \big) - \sigma\big( \dotp{\cvec{w}}{\cvec{x}_{\eta}} - \rho \|\cvec{x}_{\eta}\| \big) }{ 1 - \sigma\big(\dotp{\cvec{w}}{\cvec{x}_{c}}\big) - \sigma\big(\dotp{\cvec{w}}{\cvec{x}_{\eta}}\big)}.
\end{align*}
We say that the gradient at dimension \( j \) is down-weighted by SAM if \( 0 < r^j < 1 \), indicating that SAM reduces the gradient magnitude relative to SGD.

\boldhdr{The transitional phase} 
Our analysis focuses on the transitional phase of training, during which the model shifts from fitting clean samples to memorizing noisy labels. This phase is particularly important because the difference between SAM and SGD becomes most evident here, as illustrated in Figure~\ref{fig:sam_vs_sgd}: after the 100th epoch, once the clean samples are well learned, both methods begin to fit noisy labels, but the rate at which SAM memorizes noisy labels is significantly lower than that of SGD. Prior work~\citep{NEURIPS2020_ea89621b} has shown that in linear models, the early training dynamics are dominated by the majority of clean samples, and thus clean data are typically learned first. At this stage, the model can predict the correct true class for both clean and noisy samples. Since both \( \cvec{x}_c \) and \( \cvec{x}_\eta \) belong to the true class \( y = 0 \), the linear model outputs probabilities less than $0.5$, or equivalently \( \dotp{\cvec{w}}{\cvec{x}_c} < 0 \) and \( \dotp{\cvec{w}}{\cvec{x}_\eta} < 0 \).

\vspace{0.3cm}
\begin{remark}
\label{rm:learn_noise} 
    In the transitional phase, noisy gradient dominate the mini-batch gradient. Specifically, the gradients of the clean sample \( \boldsymbol{x}_c \) and the noisy sample \( \boldsymbol{x}_\eta \) take opposite signs, since \( \sigma ( \langle \boldsymbol{w}, \boldsymbol{x}_c \rangle) ( \sigma ( \langle \boldsymbol{w}, \boldsymbol{x}_\eta \rangle) - 1 ) < 0 \). Furthermore, the noisy sample induces a larger gradient magnitude than the clean sample: \( |\sigma ( \langle \boldsymbol{w}, \boldsymbol{x}_c \rangle)| < | \sigma ( \langle \boldsymbol{w}, \boldsymbol{x}_\eta \rangle ) - 1 | \). As a result, the mini-batch gradient is biased toward learning the noisy sample, which drives the model to fit incorrect labels.
\end{remark}

\vspace{0.2cm}
\begin{lemma}
\label{lemma:r_j}
    Let $C > 0$ be a constant and \( \sigma(x) = \frac{1}{1+e^{-x}} \) denote the sigmoid function. For any negative real numbers $z_1, z_2$ such that \( z_1 > z_2 - C \), we have 
    \begin{equation}
    \label{eq:less_than_1}
        0 < \frac{1 - \sigma(z_1 + C) - \sigma(z_2 - C)}{1 - \sigma(z_1) - \sigma(z_2) } < 1.
    \end{equation}
\end{lemma}
The proof of Lemma~\ref{lemma:r_j} is provided in the Appendix~\ref{appendix:proof_3_1}.

\vspace{0.3cm}
\begin{remark}
    SAM slows down the learning of noisy labels.
    Let \( z_1 = \dotp{\cvec{w}}{\cvec{x}_c} \), \( z_2 = \dotp{\cvec{w}}{\cvec{x}_\eta} \), and \( C = \rho \| \cvec{x}_\eta \| \). Assume both samples yield similar confidence predictions and have equal norms \( \| \cvec{x}_c \| = \| \cvec{x}_\eta \| \). Then the condition \( z_1 > z_2 - C \) holds, and Lemma~\ref{lemma:r_j} guarantees that \( 0 < r^j < 1 \). This confirms that SAM down-weights the noisy gradient relative to SGD. Combining this with Remark~\ref{rm:learn_noise}, which establishes that SGD gradient is biased toward overfitting to noisy labels, we conclude that SAM mitigates this bias by reducing the influence of such gradients.
\end{remark}

\noindent\textit{Discussion.} Eq.~\eqref{eq:less_than_1} can be reformulated to provide an intuitive interpretation. The condition \( r^j < 1 \) holds if: 
\begin{equation}
    ( \sigma( z_2 - C ) - 1 ) - ( \sigma(z_2) - 1 ) < \sigma( z_1 + C ) - \sigma(z_1).
\end{equation}
Here, the left-hand side represents the change in the noisy sample's gradient induced by SAM, while the right-hand side reflects the corresponding change for the clean sample. This shows that SAM amplifies the clean gradient more than the noisy one, thereby reducing the influence of noise-driven gradients in the aggregated mini-batch gradient and giving rise to its down-weighting behavior.

\subsection{Empirical Validation in DNNs}

\begin{figure*}[t]
\centering
    \begin{subfigure}{0.23\linewidth}
        \includegraphics[height=25mm]{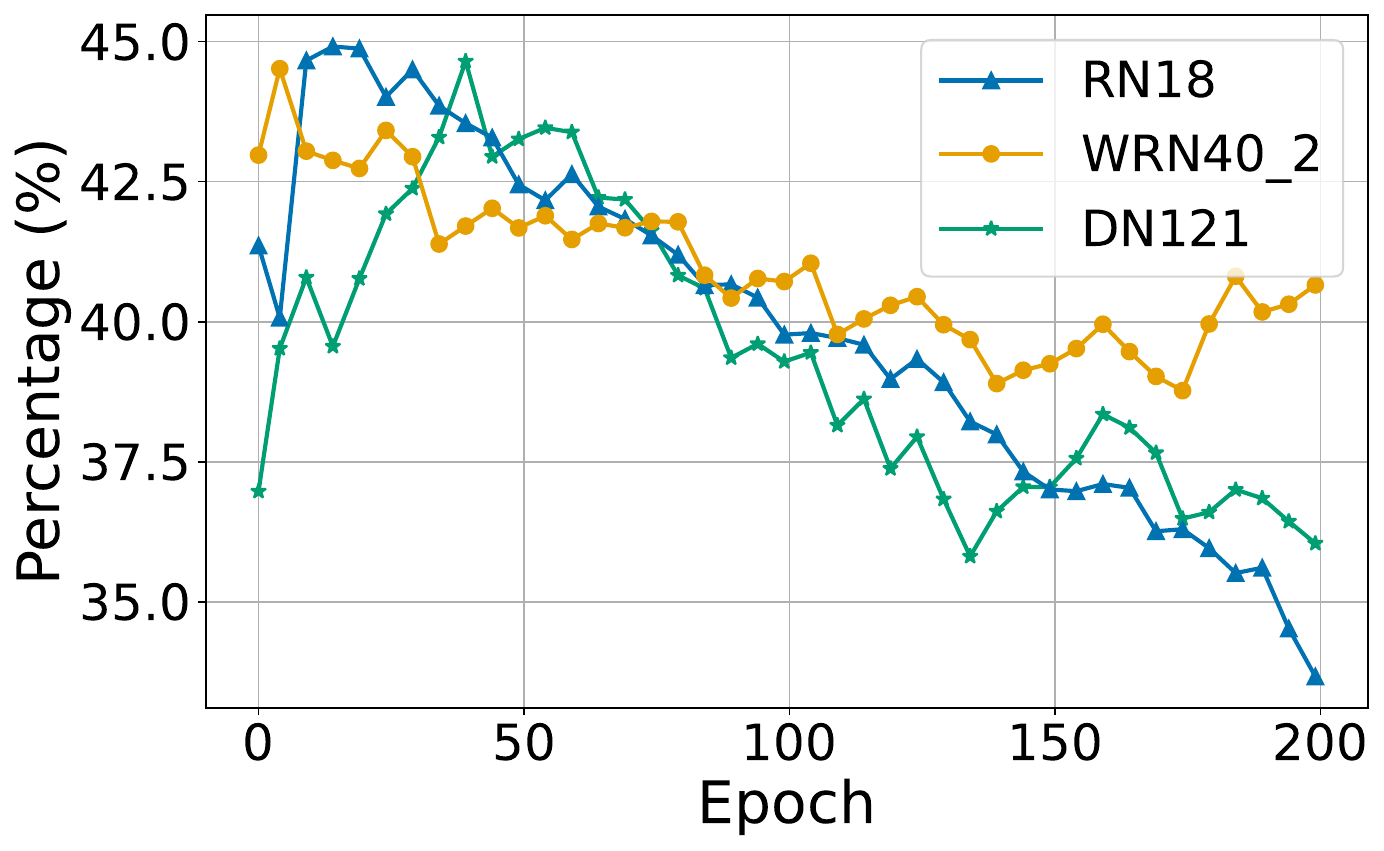}
        \caption{}
        \label{fig:distribution}
    \end{subfigure}
    \hfill
    \begin{subfigure}{0.23\linewidth}
        \includegraphics[height=25mm]{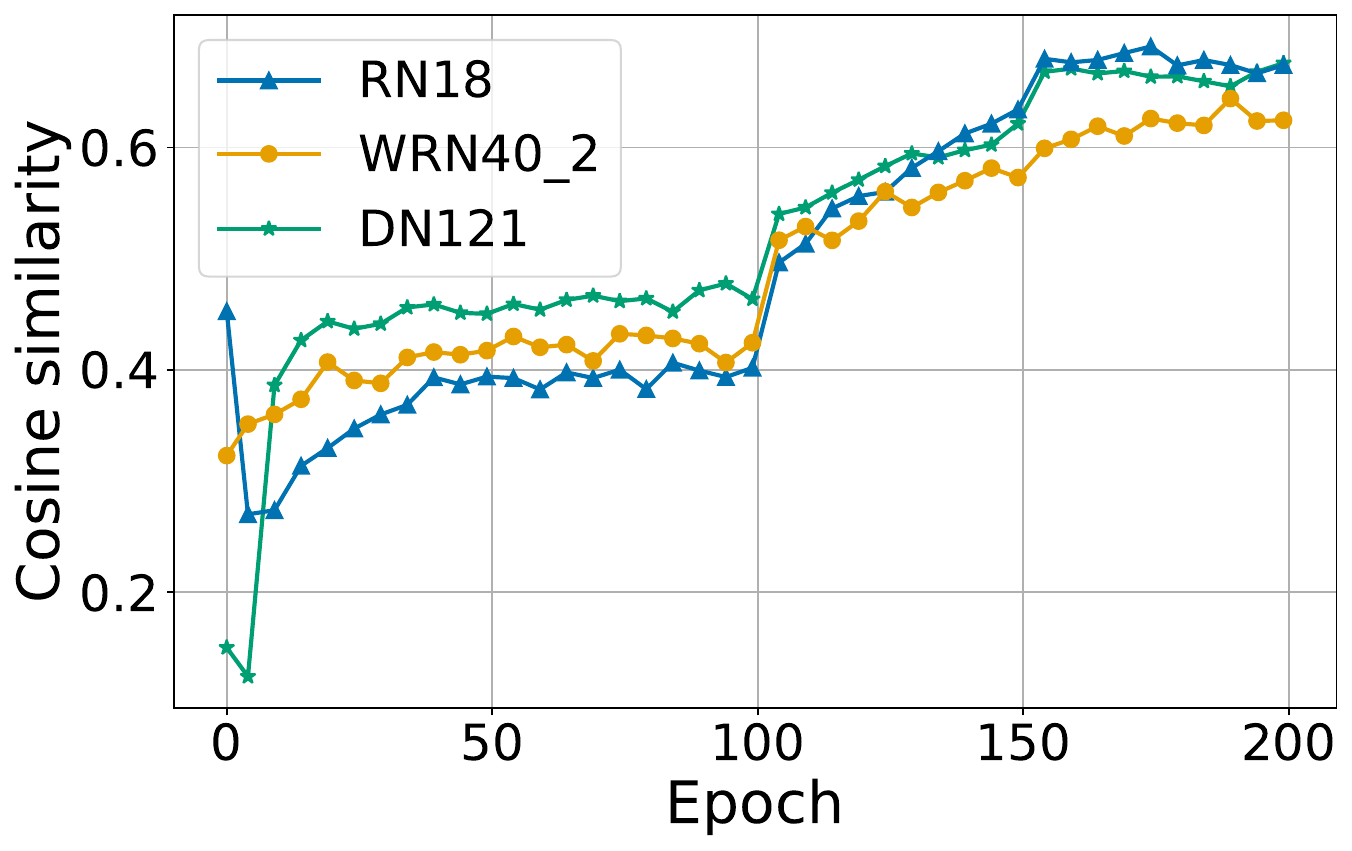}
        \caption{}
        \label{fig:cosine_sim_resnet18_cifar10}
    \end{subfigure}
    \hfill
    \begin{subfigure}{0.23\linewidth}
        \includegraphics[height=25mm]{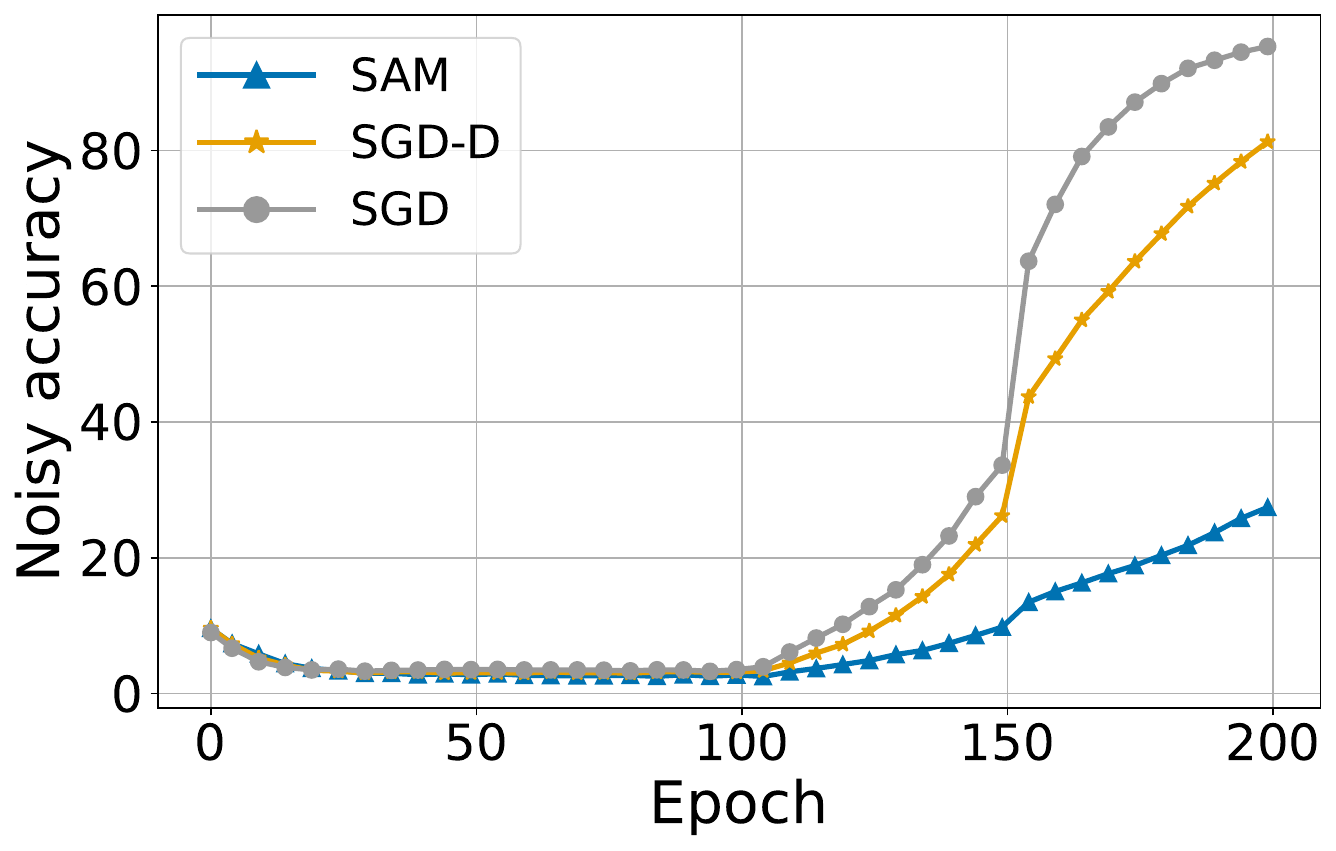}
        \caption{}
        \label{fig:noise_acc_only}
    \end{subfigure}
    \hfill
    \begin{subfigure}{0.23\linewidth}
        \includegraphics[height=25mm]{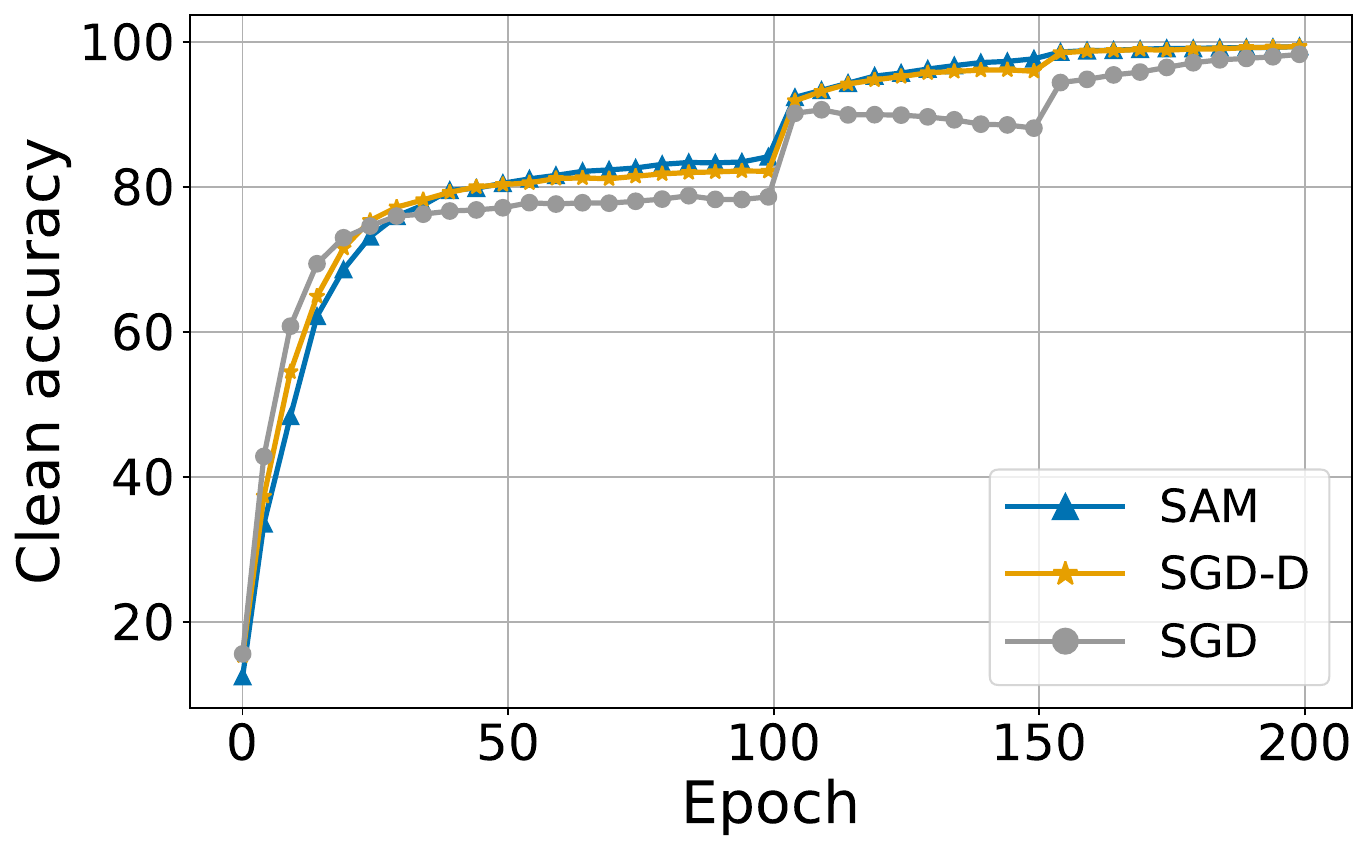}
        \caption{}
        \label{fig:clean_acc_only}
    \end{subfigure}
    \vspace{-0.5em}
    \caption{(a) Percentage (\%) of down-weighted elements $\mathcal{S}^{dw}$. 
             (b) Cosine similarity between $g^{\mathcal{S}^{dw}}$ and $g^{\text{noise}}$. 
             (c) Noisy accuracy of SGD, SAM, and a SAM variant without gradient down-weighting (SGD-D). (d) Clean accuracy of SGD, SAM, and a SAM variant without gradient down-weighting (SGD-D).
             }
    \label{fig:samwo_noise_acc}
\end{figure*}

We empirically validate the theoretical analysis from Section~\ref{sec:theory} regarding SAM's gradient down-weighting behavior in DNNs. All experiments are conducted on CIFAR-10 with 50\% symmetric label noise, using ResNet18 (RN18), WideResNet40-2 (WRN40\_2), and DenseNet121 (DN121).

\textbf{SAM consistently down-weights a significant portion of gradient elements.} To assess the occurrence of SAM's down-weighting behavior, we compute the element-wise ratio between SAM and SGD gradients, $r_i = g_i^{\text{SAM}} / g_i^{\text{SGD}}$, for each update. Let \( d \) be the number of network parameters, we then define the set of down-weighted elements as
\begin{equation}
\label{def:dw_elements}
    \mathcal{S}^{dw} = \{ i \in \{1, 2, \ldots, d\} \mid 0 < r_i < 1 \},
\end{equation}
As shown in Figure~\ref{fig:distribution}, SAM consistently down-weights a substantial fraction of gradient elements (35–45\%) throughout training across three different architectures, confirming that this is a systematic and nontrivial behavior.

\textbf{Down-weighted gradient elements align with gradients from noisy labels.} To validate that down-weighted elements correspond to updates induced by noisy labels, we measure the cosine similarity between (i) the SGD gradient computed on the entire mini-batch but projected onto the elements in $\mathcal{S}^{dw}$, denoted as $\boldsymbol{g}^{\mathcal{S}^{dw}}$, and (ii) the SGD gradient computed using only the noisy examples within the same mini-batch, denoted as $ \boldsymbol{g}^{\text{noise}}$. Specifically, we define the down-weighted SGD gradient at each element $i$ as
\begin{equation*}
    g_i^{\mathcal{S}^{dw}} = \begin{cases}
        g_i^{\text{SGD}} & \text{if } i \in \mathcal{S}^{dw}, \\
        0 & \text{otherwise}.
    \end{cases}
\end{equation*}
Figure~\ref{fig:cosine_sim_resnet18_cifar10} shows that this similarity increases over time, peaking around $\sim$0.7 between epochs 150 and 175. This indicates that the elements down-weighted by SAM are closely aligned with gradients from noisy examples, supporting our claim that SAM mitigates noise-driven gradients at the element-wise level.

\textbf{Isolating the effect of down-weighting on SAM's robustness.} 
To evaluate the importance of SAM’s down-weighting mechanism, we conduct an ablation study by creating a variant that nullifies this behavior. Specifically, we construct a variant, \textit{SAM without down-weighting} (SGD-D), in which gradient elements in $\mathcal{S}^{dw}$ are replaced with their SGD counterparts:
\begin{equation*}
    g_i^\text{SGD-D} = \begin{cases}
        g_i^\text{SGD}, & \text{if } i \in \mathcal{S}^{dw}, \\
        g_i^\text{SAM}, & \text{otherwise}.
    \end{cases}
\end{equation*}
As shown in Figure~\ref{fig:noise_acc_only}, this variant substantially accelerates noisy-label memorization, reaching nearly 80\% by the final epoch, whereas SAM remains around 30\% at the same stage. Although SGD-D also exhibits a pronounced increase in memorizing noisy labels, its performance on clean samples remains comparable to SAM. This contrast highlights that SAM's resistance to noisy labels depends critically on its selective down-weighting of noise-aligned gradient elements.


\section{ENHANCING NOISE ROBUSTNESS IN SAM VIA GRADIENT REWEIGHTING}
\label{sec:proposed_methods}

Building on our analysis in the previous section, we introduce 
\textbf{SANER}, an optimizer designed to enhance SAM's robustness to label noise. 
The core insight is that the gradient elements implicitly down-weighted by SAM are those most responsible for noisy-label memorization. 
SANER leverages this finding by applying an explicit reweighting scheme that further suppresses these gradient components. 
We provide extensive experimental validation, showing that SANER consistently outperforms both SAM and SGD across various setups. 
In addition, SANER improves the performance of other SAM-like variants, while preserving their advantage in challenging overfitting scenarios.

\subsection{The SANER Algorithm}
\label{subsec:proposed_methods}

SANER improves robustness by explicitly identifying and re-weighting the gradient elements that SAM naturally down-weights. Specifically, we introduce a binary mask vector $\boldsymbol{m} \in \{0,1\}^d$ to identify these elements and apply a scaling hyperparameter $\alpha \in \mathbb{R}$ to adjust their contribution. The binary mask for the $i$-th index is computed as
\begin{align}
    m_i &=
    \begin{cases} 
        1, & \text{if } 0 < r_i < 1, \\
        0, & \text{otherwise},
    \end{cases}
    \quad i = 1, \ldots, d.
\end{align}
Then, the SANER update is thus given by:
\begin{align}
    \boldsymbol{g}^{\text{SANER}} 
    &= \big( \mathbf{1} - \alpha\boldsymbol{m} \big) \odot \boldsymbol{g}^{\text{SAM}}.
\end{align}
It is important to note that SANER \textit{incurs the same computational cost as SAM}, since it does not require any additional gradient evaluations. The complete procedure is summarized in Algorithm~\ref{algo:SANER}.

\begin{algorithm}[t]
\caption{SANER}
\label{algo:SANER}
\begin{algorithmic}[1]
    \State \textbf{Input:} Learning rate $\eta$, initial parameters $\boldsymbol{w}_0$, iterations $T$, perturbation size $\rho$, noise control $\alpha$
    \State Initialize model parameters: $\boldsymbol{w} = \boldsymbol{w}_0$
    \For{$t = 0$ to $T$}
        \State Sample a mini-batch of $m$ training examples to calculate gradient: $\{\boldsymbol{x}^{(1)}, \dots, \boldsymbol{x}^{(m)}\}$
        \State Compute the SGD gradient: $\boldsymbol{g}^{\text{SGD}} = \nabla_{\boldsymbol{w}} L(\boldsymbol{w})$
        \State Compute the SAM gradient: $\boldsymbol{g}^{\text{SAM}}$ via Eq.~\eqref{eq:sam_update}
        \State Calculate the gradient ratio: $\boldsymbol{r} = \boldsymbol{g}^{\text{SAM}} / \boldsymbol{g}^{\text{SGD}}$ (element-wise division)
        \State Compute $\boldsymbol{m}$ and $\boldsymbol{g}^{\text{SANER}}$ via Eq.~\ref{eq:compute_maskB} and~\ref{eq:compute_SANER}
        \State Update parameters: $\boldsymbol{w} = \boldsymbol{w} - \eta \boldsymbol{g}^{\text{SANER}}$
    \EndFor
    \State \textbf{Output:} Learned parameters $\boldsymbol{w}$
\end{algorithmic}
\end{algorithm}

\textbf{Setting $\alpha > 0$ reduces noisy fitting.} 
Our theoretical and empirical analyses in Section~\ref{sec:hypothesis_1} show that SAM’s down-weighted gradient elements slow the learning of noisy labels. 
Building on this, we hypothesize that explicitly amplifying this effect through a scaling parameter $\alpha$ can further improve robustness. 
To test this, we evaluate SANER with different values of $\alpha \in \{-0.50, -0.25, 0.00 \, (\text{SAM}), 0.25, 0.50\}$ on ResNet18 with CIFAR-10. 
We report both the noisy training accuracy and the true accuracy on noisy samples, i.e., accuracy with respect to their clean labels.  

\begin{figure}[t]
\centering
    \begin{subfigure}{0.48\linewidth}
        \centering
        \includegraphics[width=\linewidth]{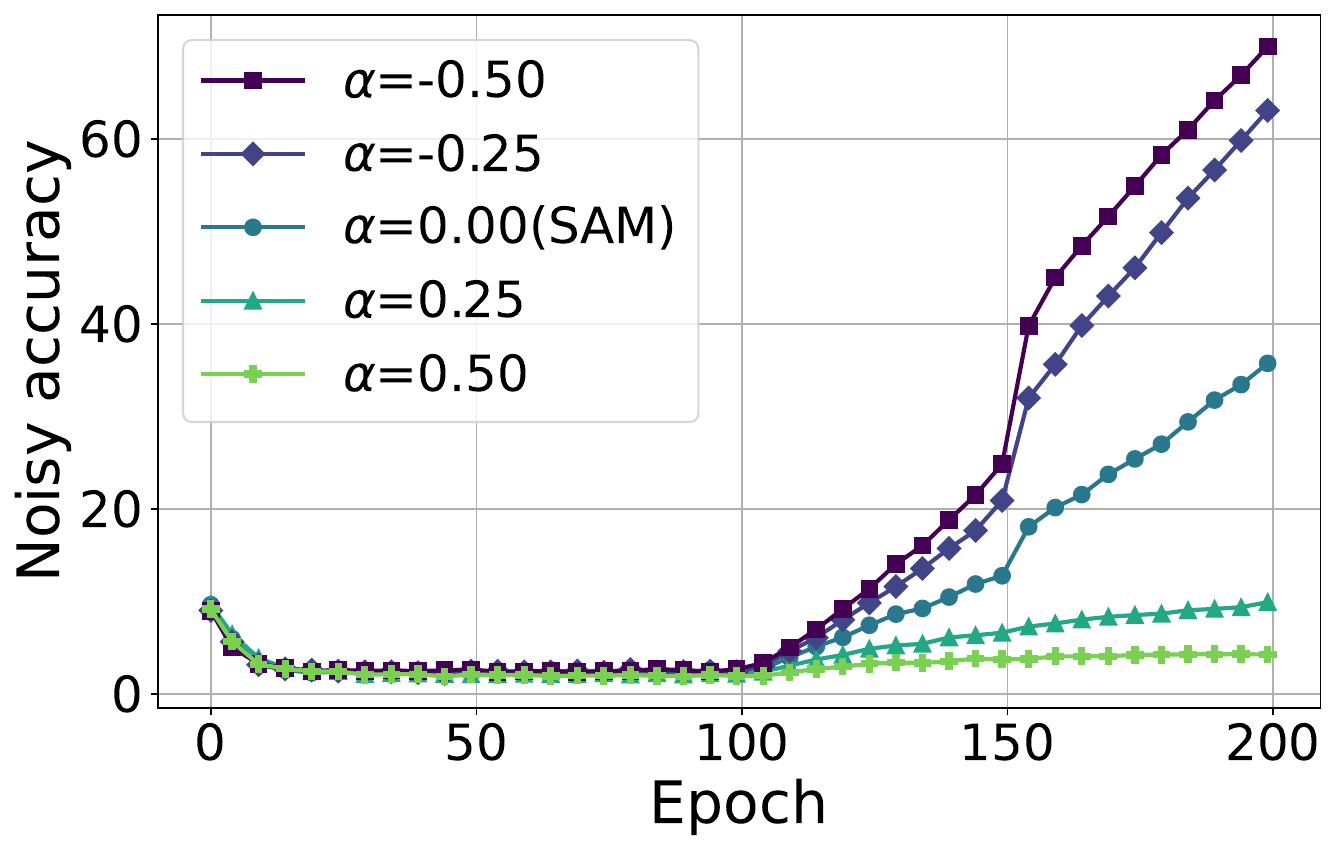}
        \caption{Noisy accuracy.}
        \label{fig:samen_vary_alpha}
    \end{subfigure}
    \hfill
    \begin{subfigure}{0.48\linewidth}
        \centering
        \includegraphics[width=\linewidth]{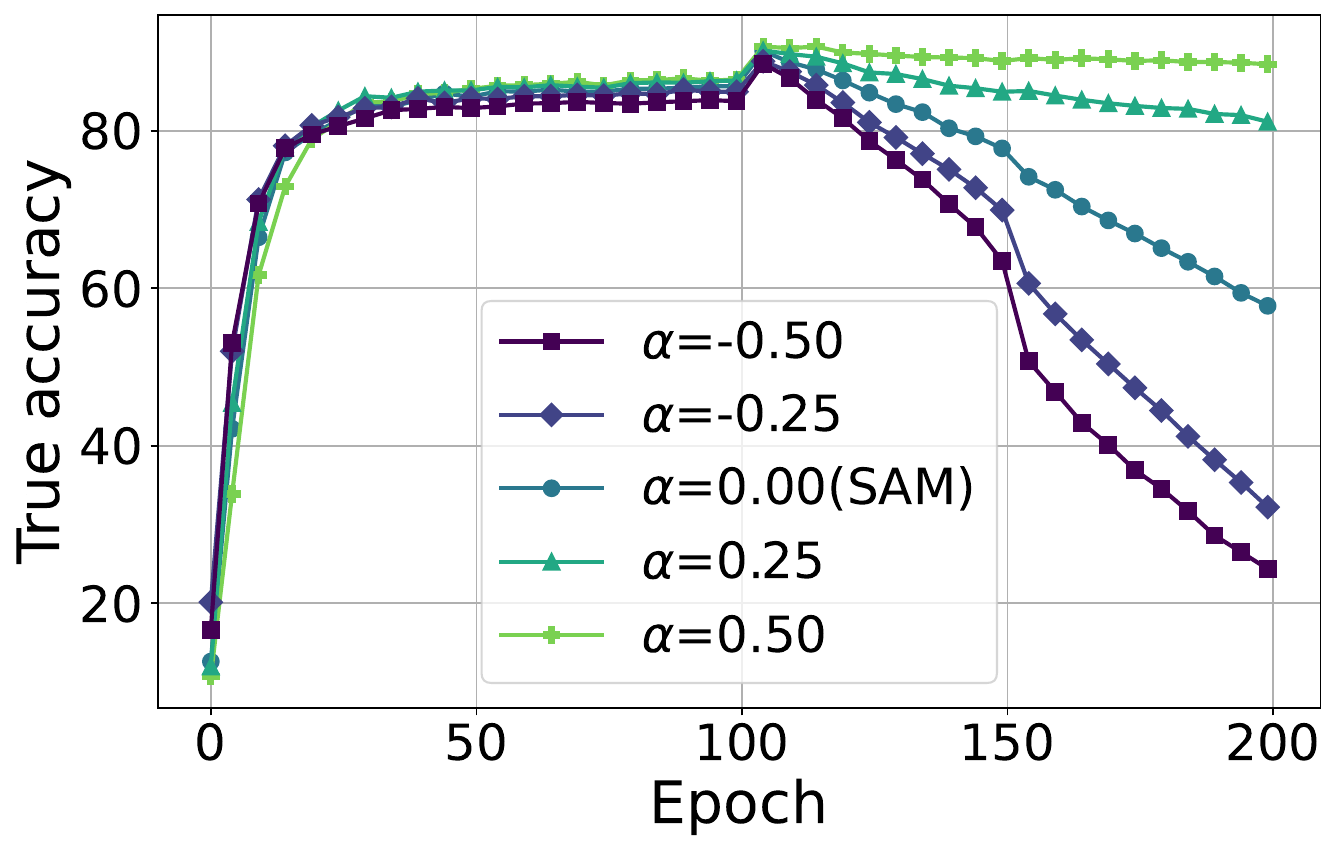}
        \caption{True accuracy on noisy.}
        \label{fig:samen_true_acc_noise}
    \end{subfigure}
    \caption{(a) Effect of hyperparameter $\alpha$ on noisy accuracy and lower values of $\alpha$ enhance noise resistance. (b) shows the true accuracy on noisy samples (measured by the true labels of noisy samples).}
    \vspace{-1.5em}
    \label{fig:samen_noise_acc_true_acc_noise}
\end{figure}

Figure~\ref{fig:samen_noise_acc_true_acc_noise} shows that values $\alpha < 0$ increase noisy fitting compared to SAM and degrade performance on the true labels of noisy samples. 
In contrast, $\alpha > 0$ suppresses noisy fitting and better preserves performance on these samples. 
The results further demonstrate that $\alpha$ in our proposed method is strongly correlated with the degree of memorization of noisy labels.

\subsection{Main Results}
\label{subsec:experimental_results}

We conduct experiments on several datasets, including CIFAR-10/100 \citep{krizhevsky2009learning}, Tiny-ImageNet \citep{Le2015TinyIV}, and Mini-Webvision \citep{li2017webvision}. To evaluate robustness under different conditions, we consider four types of label noise: symmetric noise, asymmetric noise \citep{zhang2018generalized}, instance-dependent noise \citep{xia2020part}, and real-world noise. More details on these noise types are provided in the Appendix~\ref{appendix:noise_type}.

\textbf{Training details.} 
All models were trained for 200 epochs from scratch using SGD with a momentum of 0.9, a weight decay of $5 \times 10^{-4}$, and a batch size of 128. 
We applied standard data augmentation (random crops and horizontal flips). 
The learning rate was initialized to 0.1 and reduced by a factor of 10 at epochs 100 and 150, following prior works \citep{andriushchenko2022towards, shin2023effects}.   

For computing the SAM gradient, we set the perturbation radius $\rho = 0.1$ in all experiments, following \cite{foret2021sharpnessaware}. 
Our noise control hyperparameter $\alpha$ was set to 0.5, which provided stable performance across various settings from the set $\{0.10, 0.25, 0.50, 0.75, 0.9\}$. 
We linearly increase $\alpha$ from 0.0 to its target value over the first 50 epochs to avoid abrupt changes during early training. 
All results are reported as the mean and standard deviation of the best test accuracy, averaged over three runs with different random seeds.

\textbf{CIFAR results.} We evaluate SANER on ResNet18 \citep{he2016deep} under various noise types and rates, with results summarized in Table~\ref{table:combined}. SANER consistently outperforms both SGD and SAM across all settings. On CIFAR-10, it achieves an average improvement of about 1\% over SAM, with a maximum gain of 2.7\%. On CIFAR-100, the improvement is more substantial—averaging 3\% and reaching up to 8\%. These results demonstrate that SANER effectively mitigates overfitting to noisy labels. The improvements on CIFAR-100 are particularly notable, as its larger number of classes makes it more susceptible to noise \citep{han2018co}, highlighting SANER's robustness in more challenging settings.

\begin{table*}[t]
\centering
\caption{Test accuracy comparison of SAM and SANER across different noise types and rates, trained on CIFAR-10 and CIFAR-100 with ResNet18. Bold values indicate the highest test accuracy. The teal values with an uparrow ($\uparrow$) indicate the improvement of SANER over SAM.}
    \begin{tabular}{|l|c|c|c|c|c|}
    \hline
    \multirow{2}{*}{\small{Type}} & \multirow{2}{*}{Noise rate} & \multicolumn{2}{c|}{CIFAR-10} & \multicolumn{2}{c|}{CIFAR-100} \\ \cline{3-6} 
                                  &                             & SAM & SANER & SAM & SANER \\ \hline
    \small{Symmetric noise} & 25\% & $93.05_{\pm 0.17}$ & $\textbf{94.08}_{\pm 0.11} \textcolor{teal}{\text{\tiny{($\uparrow$1.03)}}}$ & $69.68_{\pm 0.07}$ & $\textbf{72.90}_{\pm 0.21} \textcolor{teal}{\text{\tiny{($\uparrow$3.22)}}}$ \\
                                  & 50\% & $88.82_{\pm 0.08}$ & $\textbf{90.60}_{\pm 0.36} \textcolor{teal}{\text{\tiny{($\uparrow$1.78)}}}$ & $61.17_{\pm 0.14}$ & $\textbf{66.34}_{\pm 0.11} \textcolor{teal}{\text{\tiny{($\uparrow$5.17)}}}$ \\ \hline
    \small{Asymmetric noise} & 25\% & $94.75_{\pm 0.28}$ & $\textbf{94.83}_{\pm 0.14} \textcolor{teal}{\text{\tiny{($\uparrow$0.08)}}}$ & $71.57_{\pm 0.30}$ & $\textbf{74.64}_{\pm 0.13} \textcolor{teal}{\text{\tiny{($\uparrow$3.07)}}}$ \\
                                  & 50\% & $81.94_{\pm 0.71}$ & $\textbf{82.25}_{\pm 1.43} \textcolor{teal}{\text{\tiny{($\uparrow$0.31)}}}$ & $39.11_{\pm 0.50}$ & $\textbf{40.05}_{\pm 0.51} \textcolor{teal}{\text{\tiny{($\uparrow$0.94)}}}$ \\ \hline
    \small{Dependent noise} & 25\% & $92.84_{\pm 0.18}$ & $\textbf{93.67}_{\pm 0.30} \textcolor{teal}{\text{\tiny{($\uparrow$0.83)}}}$ & $69.46_{\pm 0.24}$ & $\textbf{72.93}_{\pm 0.29} \textcolor{teal}{\text{\tiny{($\uparrow$3.47)}}}$ \\
                                  & 50\% & $87.32_{\pm 1.17}$ & $\textbf{90.01}_{\pm 0.62} \textcolor{teal}{\text{\tiny{($\uparrow$2.69)}}}$ & $58.71_{\pm 0.69}$ & $\textbf{66.72}_{\pm 0.75} \textcolor{teal}{\text{\tiny{($\uparrow$8.01)}}}$ \\ \hline
    \small{Real label noise}      & -    & $86.33_{\pm 0.07}$ & $\textbf{87.89}_{\pm 0.12} \textcolor{teal}{\text{\tiny{($\uparrow$1.56)}}}$ & $62.74_{\pm 0.59}$ & $\textbf{64.75}_{\pm 0.30} \textcolor{teal}{\text{\tiny{($\uparrow$2.01)}}}$ \\ \hline
    \end{tabular}
\label{table:combined}
\end{table*}

\textbf{Tiny-ImageNet and Mini-WebVision results.} Tiny-ImageNet \citep{Le2015TinyIV}, a subset of ImageNet \citep{deng2009imagenet}, contains 100,000 color images of size $64 \times 64$ across 200 classes. We introduce 25\% symmetric label noise for this dataset. For Mini-WebVision, we follow the ``Mini'' setting from \citep{jiang2018mentornet}, selecting the first 50 classes from the Google-resized subset and evaluating on the corresponding 50 classes from the clean ImageNet 2012 validation set \citep{ILSVRC15}. As shown in Table~\ref{table:mini_webvision}, SANER outperforms SAM by approximately 4\% on Tiny-ImageNet and improves test accuracy by 3\% on Mini-WebVision.

\textbf{Different architectures.} We evaluate SGD, SAM, and SANER on CIFAR-100 across different architectures including ResNet34 \citep{he2016deep}, DenseNet121 \citep{huang2017densely}, WideResNet40-2 and WideResNet28-10 \citep{zagoruyko2017wideresidualnetworks} to assess SANER's adaptability across models. Table~\ref{table:architecture} shows that SANER consistently surpasses SAM across all settings, highlighting the role of SAM's element-wise down-weighting in enhancing robustness across diverse architectures.

\textbf{Integration with SAM-like optimizers.} Beyond designing SANER specifically for SAM, we extend our evaluation to other optimizers built upon SAM’s theoretical foundation to examine the general effectiveness of SANER. To this end, we evaluate SANER's effectiveness when integrated into SAM-like optimizers on CIFAR-10 and CIFAR-100 using ResNet18. Specifically, we compare standard versions of ASAM \citep{kwon2021asam}, GSAM \citep{zhuang2022surrogate}, FSAM \citep{li2024friendly}, and VaSSO \citep{li2024enhancing} with the SANER-enhanced versions. Detailed implementation settings and integration procedures are provided in Appendix~\ref{appendix-subsec:sam-variants}.

As shown in Table~\ref{table:sam_like}, SANER consistently improves test accuracy across all noise levels. The gains are particularly notable on CIFAR-100 under 50\% symmetric noise, where SANER boosts performance by 4–5\%. These results validate the shared behavior among SAM-like optimizers regarding down-weighted gradient elements.

\begin{table}[t]
\centering
\caption{The test accuracy on the Tiny-ImageNet (Tiny-IN) for different models and the Top-1 validation accuracy on the clean ImageNet 2012 validation set for ResNet18 trained on Mini-WebVision (Mini-WV). Bold values indicate the highest test accuracy.}
\resizebox{1.0\columnwidth}{!}{%
    \begin{tabular}{|l|c|c|c|c|}
    \hline
    Dataset & Architecture  & SGD & SAM & SANER \\ \hline
    \multirow{3}{*}{ \small{Tiny-IN} }
    & ResNet18  & $56.50$ & $57.60$  & \bm{$61.60$} \\
    & ResNet34  & $56.82$ & $59.30$  & \bm{$63.22$} \\ 
    & \small{WRN28-10}  & $57.94$ & $59.84$  & \bm{$64.08$} \\ 
    \hline
    \multirow{1}{*}{ \small{Mini-WV} }
    & ResNet18       & $64.96$ & $67.48$  & \bm{$70.84$} \\ \hline
    \end{tabular}
    }
\label{table:mini_webvision}
\end{table}

\subsection{Ablation Studies}

\begin{table}[t]
\centering
\caption{Test accuracy comparison of different architectures using SGD, SAM, and SANER on CIFAR-100 (Symmetric noise). Bold values indicate the highest test accuracy and $\gamma$ denotes the noise rate.}
\resizebox{1.0\columnwidth}{!}{%
    \begin{tabular}{|l|c|c|c|c|}
    \hline
    Model              & $\gamma$ & SGD & SAM & SANER \\ \hline
    \multirow{2}{*}{\small{ResNet34}} & 25\% & $69.07_{\pm 0.53}$ & $71.10_{\pm 0.83}$ & $\textbf{74.02}_{\pm 0.22}$ \\ 
                              & 50\% & $59.73_{\pm 1.26}$ & $62.49_{\pm 1.18}$ & $\textbf{67.26}_{\pm 0.28}$  \\ \hline 
    \multirow{2}{*}{\small{DN121}}    & 25\% & $69.13_{\pm 0.48}$ & $71.61_{\pm 0.49}$ & $\textbf{73.89}_{\pm 0.64}$ \\ 
                              & 50\% & $58.19_{\pm 1.20}$ & $60.74_{\pm 0.72}$ & $\textbf{64.26}_{\pm 0.62}$ \\ \hline
    \multirow{2}{*}{\small{WRN40-2}}  & 25\% & $67.81_{\pm 0.27}$ & $69.75_{\pm 0.26}$ & $\textbf{70.35}_{\pm 0.10}$ \\ 
                              & 50\% & $60.51_{\pm 0.18}$ & $62.58_{\pm 0.35}$ & $\textbf{64.71}_{\pm 0.55}$ \\ \hline
    \multirow{2}{*}{\small{WRN28-10}} & 25\% & $70.78_{\pm 0.20}$ & $72.56_{\pm 0.18}$ & $\textbf{76.20}_{\pm 0.41}$ \\ 
                              & 50\% & $61.94_{\pm 0.49}$ & $64.12_{\pm 0.30}$ & $\textbf{70.80}_{\pm 0.28}$ \\ \hline
    \end{tabular}
    }
\label{table:architecture}
\vspace{-1em}
\end{table}

\begin{table*}[t]
\centering
\caption{Test accuracy comparison of different SAM-like optimizers with and without SANER integration on ResNet18 and CIFAR-10/CIFAR-100 (Symmetric noise). Bold values indicate the highest test accuracy. The teal values in parentheses ($\uparrow$) show the improvement from integrating SANER.}
    \begin{tabular}{|l|c|c|c|c|c|}
    \hline
    \multirow{2}{*}{\small{Variants}} & \multirow{2}{*}{Noise rate} & \multicolumn{2}{c|}{CIFAR-10} & \multicolumn{2}{c|}{CIFAR-100} \\ \cline{3-6} 
                                      &                             & Original & +SANER & Original & +SANER \\ \hline
    \multirow{2}{*}{ASAM}             & 25\%                        & $92.88_{\pm 0.13}$   & $\textbf{92.96}_{\pm 0.06} \textcolor{teal}{\text{\tiny{($\uparrow$0.08)}}}$   & $70.67_{\pm 0.40}$   & $\textbf{72.44}_{\pm 0.10} \textcolor{teal}{\text{\tiny{($\uparrow$1.77)}}}$   \\ 
                                      & 50\%                        & $88.70_{\pm 0.18}$   & $\textbf{88.80}_{\pm 0.10} \textcolor{teal}{\text{\tiny{($\uparrow$0.10)}}}$   & $63.04_{\pm 0.25}$   & $\textbf{66.62}_{\pm 0.13} \textcolor{teal}{\text{\tiny{($\uparrow$3.58)}}}$   \\ \hline
    \multirow{2}{*}{GSAM}             & 25\%                        & $93.10_{\pm 0.12}$   & $\textbf{94.09}_{\pm 0.16} \textcolor{teal}{\text{\tiny{($\uparrow$0.99)}}}$   & $69.65_{\pm 0.39}$   & $\textbf{72.97}_{\pm 0.27} \textcolor{teal}{\text{\tiny{($\uparrow$3.32)}}}$   \\ 
                                      & 50\%                        & $88.71_{\pm 0.15}$   & $\textbf{90.69}_{\pm 0.17} \textcolor{teal}{\text{\tiny{($\uparrow$1.98)}}}$   & $61.25_{\pm 0.33}$   & $\textbf{66.19}_{\pm 0.15} \textcolor{teal}{\text{\tiny{($\uparrow$4.94)}}}$   \\ \hline
    \multirow{2}{*}{FSAM}             & 25\%                        & $92.93_{\pm 0.08}$   & $\textbf{94.00}_{\pm 0.15} \textcolor{teal}{\text{\tiny{($\uparrow$1.07)}}}$   & $69.49_{\pm 0.35}$   & $\textbf{72.94}_{\pm 0.58} \textcolor{teal}{\text{\tiny{($\uparrow$3.45)}}}$   \\ 
                                      & 50\%                        & $88.71_{\pm 0.13}$   & $\textbf{90.47}_{\pm 0.01} \textcolor{teal}{\text{\tiny{($\uparrow$1.76)}}}$   & $61.24_{\pm 0.32}$   & $\textbf{66.25}_{\pm 0.15} \textcolor{teal}{\text{\tiny{($\uparrow$5.01)}}}$   \\ \hline
    \multirow{2}{*}{VaSSO}            & 25\%                        & $92.35_{\pm 0.12}$   & $\textbf{93.31}_{\pm 0.32} \textcolor{teal}{\text{\tiny{($\uparrow$0.96)}}}$   & $68.86_{\pm 0.18}$   & $\textbf{72.43}_{\pm 0.46} \textcolor{teal}{\text{\tiny{($\uparrow$3.57)}}}$   \\ 
                                      & 50\%                        & $87.93_{\pm 0.06}$   & $\textbf{89.66}_{\pm 0.57} \textcolor{teal}{\text{\tiny{($\uparrow$1.73)}}}$   & $60.46_{\pm 0.05}$   & $\textbf{65.55}_{\pm 0.51} \textcolor{teal}{\text{\tiny{($\uparrow$5.09)}}}$   \\ \hline
    \end{tabular}
\label{table:sam_like}
\end{table*}

\begin{figure}[ht]
\centering
    \begin{subfigure}{0.48\linewidth}
        \centering
        \includegraphics[height=3.05cm,width=3.95cm]{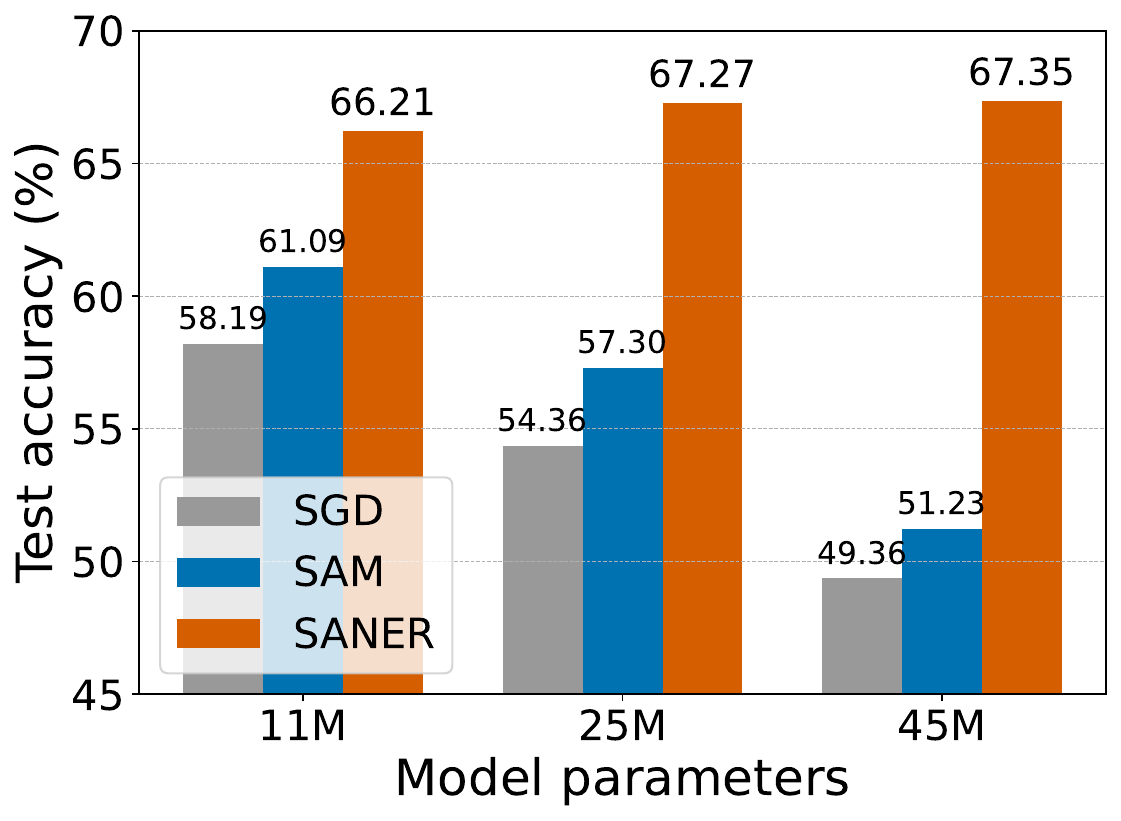}
        \caption{}
        \label{fig:width}
    \end{subfigure}
    \hfill
    \begin{subfigure}{0.48\linewidth}
        \centering
        \includegraphics[height=3cm, width=3.95cm]{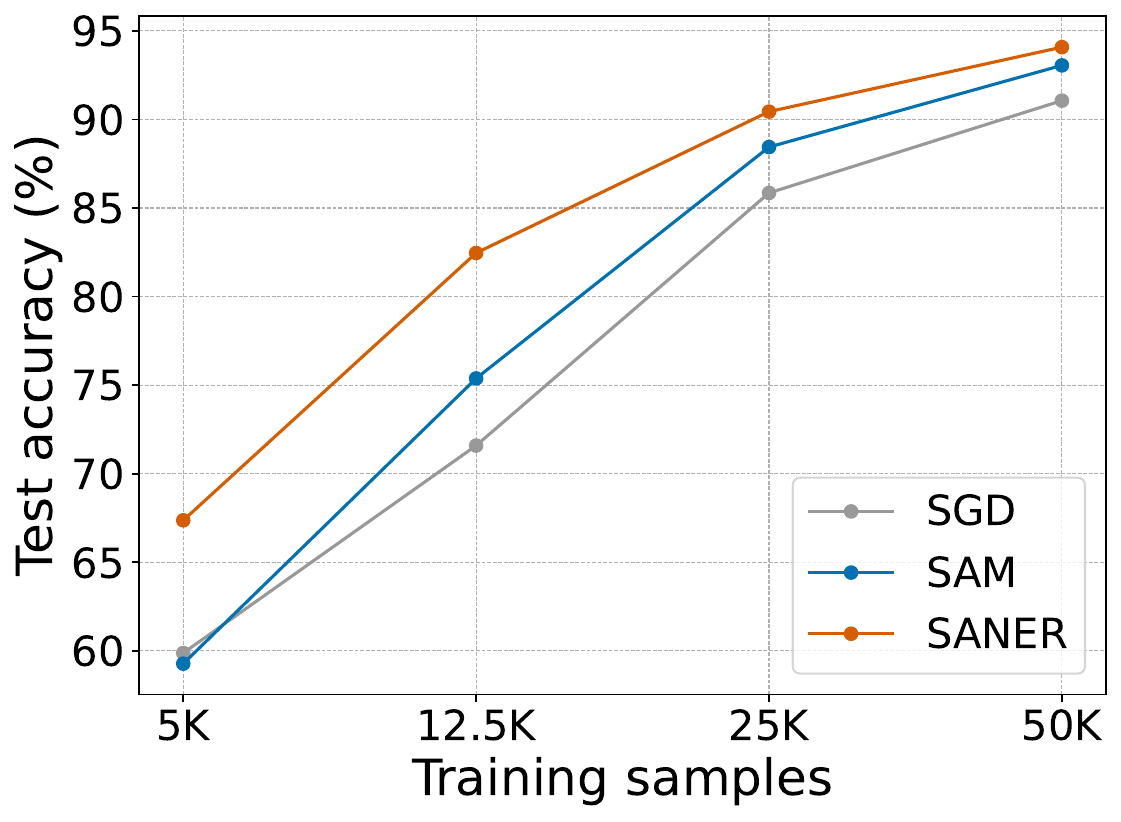}
        \caption{}
        \label{fig:vary_training_set}
    \end{subfigure}
    \caption{Test accuracy comparison of ResNet18 under different conditions and noise levels: (a) increasing layer width with 50\% label noise and (b) varying CIFAR-10 training set size. SANER consistently outperforms other methods across all settings.}
    \label{fig:}
\end{figure}

\textbf{Various overfitting scenarios.} We further evaluate SANER under two challenging settings known to induce overfitting: training wider models and using smaller datasets, following the protocol from \citep{Nakkiran2020Deep}.

\textit{Increased model width.} We observe that as model capacity increases, the performance of both SGD and SAM degrades, consistent with prior findings on overfitting noisy labels \citep{Belkin_2019, Nakkiran2020Deep, zhang2021understanding}. In contrast, SANER benefits from the increased capacity, substantially widening its performance gap over SAM in highly over-parameterized regimes, as shown in Figure~\ref{fig:width}.

\textit{Limited dataset size.} Overfitting also becomes more pronounced when the dataset size is small relative to model capacity. We evaluate SGD, SAM, and SANER on CIFAR-10 with training set sizes reduced to 10\%, 25\%, and 50\%, while retaining the full test set. We omit CIFAR-100 due to its limited samples per class, which causes instability under further reduction. As shown in Figure~\ref{fig:vary_training_set}, SANER consistently outperforms both baselines, with a significant gain (7\%) over SAM when trained on only 12,500 examples.

These experiments reveal a clear trend: the performance gap between SANER and SAM widens as conditions become more prone to overfitting. Whether through increased model capacity or reduced data, SANER’s advantage grows, showcasing its effectiveness in preventing the memorization of noisy labels.

\begin{table}[t]
\centering
\caption{Test accuracy of ResNet18 under 25\% symmetric label noise across various SAM perturbation radii ($\rho$). Bold values highlight the highest accuracy.}
\resizebox{1.0\columnwidth}{!}{
\begin{tabular}{|l|l|c|c|c|c|}
\hline
\multirow{2}{*}{Dataset} & \multirow{2}{*}{Opt} & \multicolumn{4}{c|}{$\rho$} \\ \cline{3-6}
                    & & $0.05$ & $0.10$ & $0.15$ & $0.20$ \\ \hline
\multirow{2}{*}{\small{CIFAR-10}} 
                    & \small{SAM}   & $92.15_{\pm 0.27}$ & $93.05_{\pm 0.17}$ & $93.78_{\pm 0.22}$ & $93.90_{\pm 0.20}$  \\
                    & \small{SANER} & $\textbf{93.02}_{\pm 0.25}$ & $\textbf{94.08}_{\pm 0.11}$  &  \bm{$94.27_{\pm 0.18}$} & \bm{$94.30_{\pm 0.05}$} \\ \hline
\multirow{2}{*}{\small{CIFAR-100}} & 
                    \small{SAM}  & $68.59_{\pm 0.27}$ & $69.68_{\pm 0.07}$ & $70.34_{\pm 0.26}$ & $71.19_{\pm 0.29}$ \\ 
                    & \small{SANER} & $\textbf{72.10}_{\pm 0.40}$ & $\textbf{72.90}_{\pm 0.21}$ & \bm{$73.21_{\pm 0.31}$} & $\textbf{73.42}_{\pm 0.26}$ \\ \hline 
\end{tabular}
}
\label{tab:saner_rho}
\end{table}

\begin{table}[t]
\centering
\caption{Test accuracy comparison of SAM and SANER trained on clean datasets using ResNet18. Bold values highlight the highest test accuracy.}
\resizebox{\columnwidth}{!}{
\begin{tabular}{|l|c|c|c|c|}
\hline
Dataset & SGD   & SAM & SANER  \\ \hline
\small{CIFAR-10}  & $95.18_{\pm 0.09}$ & $96.04_{\pm 0.04}$ & $\textbf{96.06}_{\pm 0.12}$   \\ \hline
\small{CIFAR-100} & $78.06_{\pm 0.09}$ & $79.19_{\pm 0.22}$ & $\textbf{79.63}_{\pm 0.36}$   \\ \hline
\end{tabular}
}
\label{table:saner_clean_data}
\end{table}
\begin{table*}[ht]
\centering
\caption{Test accuracy comparison of SAM and SANER integrated with Bootstrap (BS) and CoTeaching (CT) on CIFAR-10/100 with symmetric label noise using ResNet18. Bold values indicate the best performance under each setting and $\gamma$ denotes the noise rate.}
\begin{tabular}{|l|c||c|c||c|c||c|c|}
\hline
\multirow{2}{*}{Dataset} & \multirow{2}{*}{$\gamma$} 
& \multicolumn{2}{c||}{Baseline} 
& \multicolumn{2}{c||}{Bootstrap (BS)} 
& \multicolumn{2}{c|}{CoTeaching (CT)} \\ \cline{3-8}

& & SAM & SANER & SAM+BS & SANER+BS & SAM+CT & SANER+CT \\ \hline

\multirow{2}{*}{\small{CIFAR-10}} 
& 25\% & $93.05_{\pm 0.17}$ & $94.08_{\pm 0.11}$ & $93.35_{\pm 0.07}$ & $\bm{94.49}_{\pm 0.09}$ & $92.86_{\pm 0.17}$ & $93.41_{\pm 0.27}$ \\
& 50\% & $88.82_{\pm 0.08}$ & $90.60_{\pm 0.36}$ & $89.57_{\pm 0.49}$ & $\bm{90.98}_{\pm 0.49}$ & $86.89_{\pm 0.16}$ & $88.18_{\pm 0.23}$ \\ \hline

\multirow{2}{*}{\small{CIFAR-100}} 
& 25\% & $69.68_{\pm 0.07}$ & $72.90_{\pm 0.21}$ & $70.81_{\pm 0.18}$ & $73.27_{\pm 0.29}$ & $72.46_{\pm 0.31}$ & $\bm{73.89}_{\pm 0.38}$ \\
& 50\% & $61.17_{\pm 0.14}$ & $66.34_{\pm 0.11}$ & $64.08_{\pm 0.42}$ & $\bm{66.81}_{\pm 0.11}$ & $61.35_{\pm 0.45}$ & $64.26_{\pm 0.27}$ \\ \hline
\end{tabular}
\label{table:robust_methods_combined}
\end{table*}

\textbf{Integrating noise-robust training strategies with SAM and SANER.}
As an optimization technique, SANER can, in principle, be integrated into a wide range of label-noise learning algorithms. We evaluate the compatibility of SANER with existing noise-robust learning strategies, including hard bootstrapping~\citep{reed2014training} (BS) and CoTeaching~\citep{han2018co}.

As shown in Table~\ref{table:robust_methods_combined}, integrating bootstrapping noticeably improves the performance of standard SAM, while its benefit for SANER is comparatively smaller. This suggests that SANER already provides strong robustness to label noise, leaving limited room for further gains from bootstrapping. For CoTeaching, the improvements are less consistent, particularly for SAM under higher noise levels. Nevertheless, SANER consistently achieves stronger performance than SAM across nearly all settings, both with and without additional noise-robust strategies, demonstrating that SANER serves as a reliable and complementary optimization framework for noisy-label training.

\textbf{Impact of SAM's $\rho$ on SANER performance.}
The perturbation radius $\rho$ in SAM is a key hyperparameter that directly affects generalization performance \citep{foret2021sharpnessaware}. To understand how SANER interacts with this factor, we evaluate its performance across a range of $\rho$ values (from 0.05 to 0.20), as recommended by \cite{foret2021sharpnessaware}. 

As shown in Table~\ref{tab:saner_rho}, SANER consistently outperforms SAM for all tested radii. While increasing $\rho$ generally improves both methods, the gains introduced by SANER remain complementary and substantial. This effect is particularly evident on CIFAR-100, where the worst performance of SANER (72.57\%) still surpasses the best result achieved by SAM (70.86\%). Overall, these findings demonstrate that SANER provides a fundamental robustness improvement beyond what can be achieved by tuning $\rho$ alone.

\textbf{Noise-free scenarios.}
In addition to noisy settings, we evaluate SANER under noise-free conditions to examine its behavior when label noise is absent. Specifically, we test whether the proposed mechanism remains well-aligned with standard learning without introducing performance degradation. 

As shown in Table~\ref{table:saner_clean_data}, SANER performs on par with SAM in clean settings, without noticeable improvement. These results indicate that SANER preserves the baseline performance of SAM while avoiding any negative impact in the absence of noise. Overall, this demonstrates that SANER is a safe and broadly applicable optimization strategy across both noisy and noise-free scenarios.





\section{CONCLUSION}
In this work, we investigated the mechanisms behind SAM's robustness to label noise, identifying that element-wise gradient down-weighting is a key factor in mitigating the memorization of noisy samples during transitional phase. This down-weighting is supported by an up-weighting of the clean samples' gradients. Building on this insight, we proposed SANER, a simple modification that enhances this mechanism at no additional computational cost. Our experiments demonstrated that SANER consistently outperforms standard SAM across various datasets, noise types, and challenging overfitting scenarios. Moreover, our modification can be integrated into SAM-variants and shows consistent improvement. While our current theory is based on linear models, these strong empirical results suggest promising directions for future work, such as extending the analysis to deep models and exploring adaptive re-weighting mechanisms.

\bibliographystyle{plainnat}
\bibliography{main}
\section*{Checklist}

\begin{enumerate}

  \item For all models and algorithms presented, check if you include:
  \begin{enumerate}
    \item A clear description of the mathematical setting, assumptions, algorithm, and/or model. \textcolor{blue}{[Yes]}. See Section~\ref{sec:theory}.
    \item An analysis of the properties and complexity (time, space, sample size) of any algorithm. \textcolor{blue}{[Yes]}. See Section~\ref{subsec:proposed_methods}.
  \end{enumerate}

  \item For any theoretical claim, check if you include:
  \begin{enumerate}
    \item Statements of the full set of assumptions of all theoretical results. \textcolor{blue}{[Yes]}. See Section~\ref{sec:theory}.
    \item Complete proofs of all theoretical results. \textcolor{blue}{[Yes]}. See Appendix~\ref{appendix:proof_3_1}.
    \item Clear explanations of any assumptions. \textcolor{blue}{[Yes]}. See Section~\ref{sec:theory}.    
  \end{enumerate}

  \item For all figures and tables that present empirical results, check if you include:
  \begin{enumerate}
    \item The code, data, and instructions needed to reproduce the main experimental results (either in the supplemental material or as a URL). \textcolor{blue}{[Yes]}. See Section~\ref{subsec:experimental_results}.
    \item All the training details (e.g., data splits, hyperparameters, how they were chosen). \textcolor{blue}{[Yes]}. See Section~\ref{subsec:experimental_results}.
    \item A clear definition of the specific measure or statistics and error bars (e.g., with respect to the random seed after running experiments multiple times). \textcolor{blue}{[Yes]}
    \item A description of the computing infrastructure used. (e.g., type of GPUs, internal cluster, or cloud provider). \textcolor{blue}{[Yes]}. See Section~\ref{appendix:implementation}.
  \end{enumerate}

  \item If you are using existing assets (e.g., code, data, models) or curating/releasing new assets, check if you include:
  \begin{enumerate}
    \item Citations of the creator If your work uses existing assets. \textcolor{blue}{[Yes]}
    \item The license information of the assets, if applicable. \textcolor{blue}{[Not Applicable]}
    \item New assets either in the supplemental material or as a URL, if applicable. \textcolor{blue}{[Not Applicable]}
    \item Information about consent from data providers/curators. \textcolor{blue}{[Not Applicable]}
    \item Discussion of sensible content if applicable, e.g., personally identifiable information or offensive content. \textcolor{blue}{[Not Applicable]}
  \end{enumerate}

  \item If you used crowdsourcing or conducted research with human subjects, check if you include:
  \begin{enumerate}
    \item The full text of instructions given to participants and screenshots. \textcolor{blue}{[Not Applicable]}
    \item Descriptions of potential participant risks, with links to Institutional Review Board (IRB) approvals if applicable. \textcolor{blue}{[Not Applicable]}
    \item The estimated hourly wage paid to participants and the total amount spent on participant compensation. \textcolor{blue}{[Not Applicable]}
  \end{enumerate}

\end{enumerate}

\appendix

\onecolumn
\aistatstitle{Supplementary Materials}

\section{MISSING PROOFS}

\subsection{Proof of Lemma 3.1}
\label{appendix:proof_3_1}
\begin{proof}
     Let $ h: \mathbb{R} \to \mathbb{R} $ be defined as $h(z) = \sigma(z + C) - \sigma(z)$ for all $z \in \mathbb{R}$. It follows from the derivative of sigmoid function that
    \begin{align*}
        h'(z) 
        &= \sigma(z+C)\bigl(1-\sigma(z+C)\bigr) - \sigma(z)\bigl(1-\sigma(z)\bigr) \\
        &= \bigl[\sigma(z+C)-\sigma(z)\bigr]\Bigl[1-\sigma(z+C)-\sigma(z)\Bigr].
    \end{align*}
    As $\sigma(z+C)-\sigma(z) > 0$ for all $z \in \mathbb{R}$ due to the increasing property of $\sigma$ and $C > 0$, the inequality $h'(z) > 0$ is equivalent to $\sigma(z+C)+\sigma(z) < 1$, which means
    \[
        \frac{1}{1+e^{-(z+C)}} + \frac{1}{1+e^{-z}} < 1.
    \]
    It follows that $e^{-2z-C} > 1$, or equivalently $z < -\frac{C}{2}$, which means the function $h$ is strictly increasing on $(-\infty, -\frac{C}{2})$. Using the property $\sigma(z) = 1 - \sigma(-z)$ for all $z \in \mathbb{R}$ of the sigmoid function, we deduce that
    \begin{align}
    \label{eq:symmetric}
        h \big(z \big) &= \sigma \big( z + C \big) - \sigma \big( z \big) \notag \\
        &= \sigma \big( - z \big) - \sigma \big( - z - C  \big) = h \big(  -z - C \big),
    \end{align}
    for all $z \in \mathbb{R}$. Since $z_1, z_2 < 0$, and $z_1 > z_2 - C$, if $z_1 < \frac{-C}{2}$, then $h(z_1) > h(-z_2-C) = h(z_2)$ due to the increasing property of $h$ on $(-\infty, \frac{-C}{2})$. Otherwise, $-\frac{C}{2} \leq z_1 < 0$ implies that $-\frac{C}{2} \geq -z_1 - C > -C$. Combining this with \eqref{eq:symmetric}, the increasing property of $h$ on $(-\infty, \frac{-C}{2})$ and $z_2 < 0$, we have 
    \begin{align*}
        h(z_1) = h(-z_1 - C) > h(-C) > h(z_2-C).
    \end{align*}
     Using this with $h(z) = \sigma(z + C) - \sigma(z)$, we have $\sigma(z_1 + C) - \sigma(z_1) > \sigma(z_2) - \sigma(z_2 - C)$, or equivalently
    \begin{equation}
    \label{eq:ineq}
         1 - \sigma(z_1) - \sigma(z_2) > 1 - \sigma(z_1 + C) - \sigma(z_2 - C).
    \end{equation}
    Using the property $\sigma(z) < 0.5$ for all $z < 0$ and $z_1, z_2 < 0$, we deduce that $1 - \sigma(z_1) - \sigma(z_2) > 0$. Combining this with \eqref{eq:ineq}, we have 
    \begin{equation*}
        \frac{1 - \sigma(z_1 + C) - \sigma(z_2 - C)}{1 -\sigma(z_1) - \sigma(z_2)} < 1.
    \end{equation*}
    We also aim to show that $1 - \sigma(z_1 + C) - \sigma(z_2 - C) = \sigma(-z_1 - C) - \sigma(z_2 - C) > 0$, which is equivalent to 
    \begin{equation*}
        \frac{1}{ 1 + \exp(z_1 + C)} - \frac{1}{1 + \exp(-z_2 + C)} > 0.
    \end{equation*}
    This inequality further simplifies to $\exp(-z_2) - \exp(z_1) > 0$, which holds because $-z_2 > 0 > z_1$. Combining this with $1 - \sigma(z_1) - \sigma(z_2) > 0$, we have 
    \begin{equation*}
        0 < \frac{1 - \sigma(z_1 + C) - \sigma(z_2 - C)}{1 -\sigma(z_1) - \sigma(z_2)},
    \end{equation*}    
    which verifies the lemma.
\end{proof}

\section{IMPLEMENTATION DETAILS}
\label{appendix:implementation}
\label{appendix:noise_type}
In this paper, we use four types of noise as follows:
\begin{enumerate}
    \item Symmetric noise: Each label is flipped to any other class with equal probability noise rate $\gamma$. 
    
    \item Asymmetric noise: Labels are flipped to similar, but not identical classes \citep{zhang2018generalized}. For CIFAR-10, we generate asymmetric noisy labels by mapping specific classes to their most similar counterparts: TRUCK to AUTOMOBILE, BIRD to AIRPLANE, DEER to HORSE, CAT to DOG with noise rate $\gamma$, and leaving other labels unchanged. For CIFAR-100, each class is shifted circularly to the next class with noise rate $\gamma$.
    
    \item Instance-dependent noise: The mislabeling probability of each instance depends on its input features. In our experiments, we use instance-dependent noise from PDN~\citep{xia2020part} with noisy rate $\gamma$, where the noise is synthesized based on DNN prediction errors.
    
    \item Real-world noise: Labels are taken from the mislabeling of real-world human annotations. For CIFAR datasets, we use the ``Worst" label set from CIFAR-10N and the ``Fine" label set from CIFAR-100N \citep{wei2022learning}.
\end{enumerate}

All experiments are mainly conducted on an Ubuntu Linux machine equipped with an NVIDIA RTX 3090 GPU with 24GB of memory.

\section{ADDITIONAL EXPERIMENTS}

\subsection{Integration with SAM Variants}
\label{appendix-subsec:sam-variants}

\textbf{Experimental setup.} To evaluate the effect of SANER on SAM-based optimizers, we conducted experiments on CIFAR-10 and CIFAR-100 using ResNet18. The SAM variants used include ASAM \citep{kwon2021asam}, GSAM \citep{zhuang2022surrogate}, FSAM \citep{li2024friendly}, and VaSSO \citep{li2024enhancing}, tested both with and without SANER integration. The models were trained with label noise levels of 25\% and 50\%, and the SANER hyperparameter $\alpha = 0.5$ for all experiments as we setup when comparing with SAM. All other training configurations were kept consistent for fair comparison between methods.

\textbf{Modification of SAM-based optimizers.} SANER was integrated into the SAM-variants by modifying the update rules. Specifically, we compute a mask vector $\boldsymbol{m}$ at index $i$ as
\begin{align}
\label{eq:compute_maskB}
    m_i &=
    \begin{cases} 
        1, & \text{if } 0 < r_i < 1, \\
        0, & \text{otherwise},
    \end{cases}
    \quad i = 1, \ldots, d,
\end{align}
then replace $\boldsymbol{r} = \boldsymbol{g}^{\text{SAM*}}/{\boldsymbol{g}^{\text{SGD}}}$ and $\boldsymbol{g}^{\text{SAM*}}$ as follows:
\begin{align}
    \label{eq:compute_SANER}
    \boldsymbol{g}^{\text{SANER*}} 
    &= \big( \mathbf{1} - \alpha\boldsymbol{m} \big) \odot \boldsymbol{g}^{\text{SAM*}} 
\end{align}
where $\boldsymbol{g}^{\text{SAM*}}$ refers to the gradient of the specific SAM variant and $\boldsymbol{g}^{\text{SANER*}}$ denotes the modified gradient under SANER integration. 

\begin{figure*}[ht]
    \centering
    \begin{subfigure}{\linewidth}
        \centering
        \includegraphics[width=0.24\linewidth]{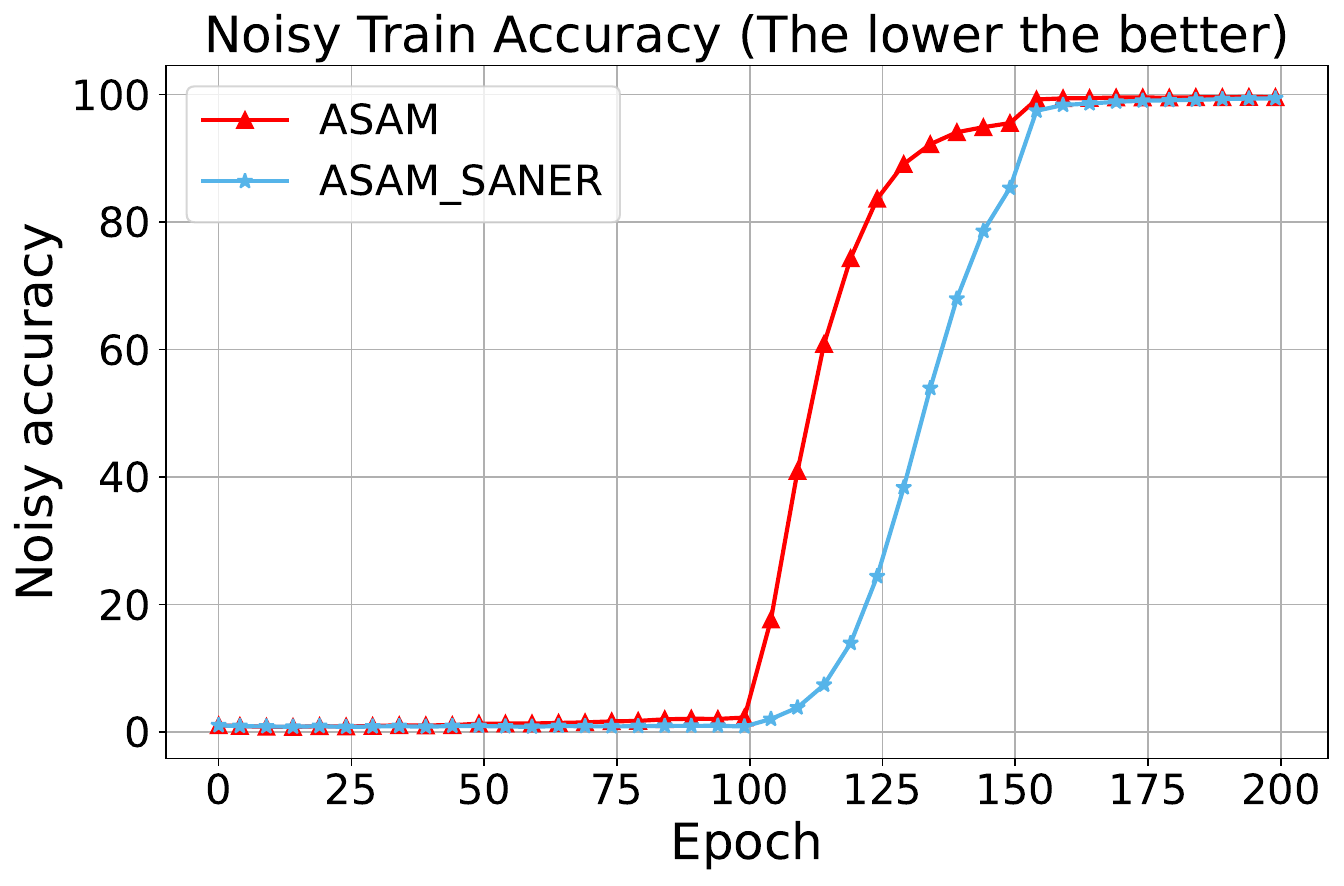}
        \includegraphics[width=0.24\linewidth]{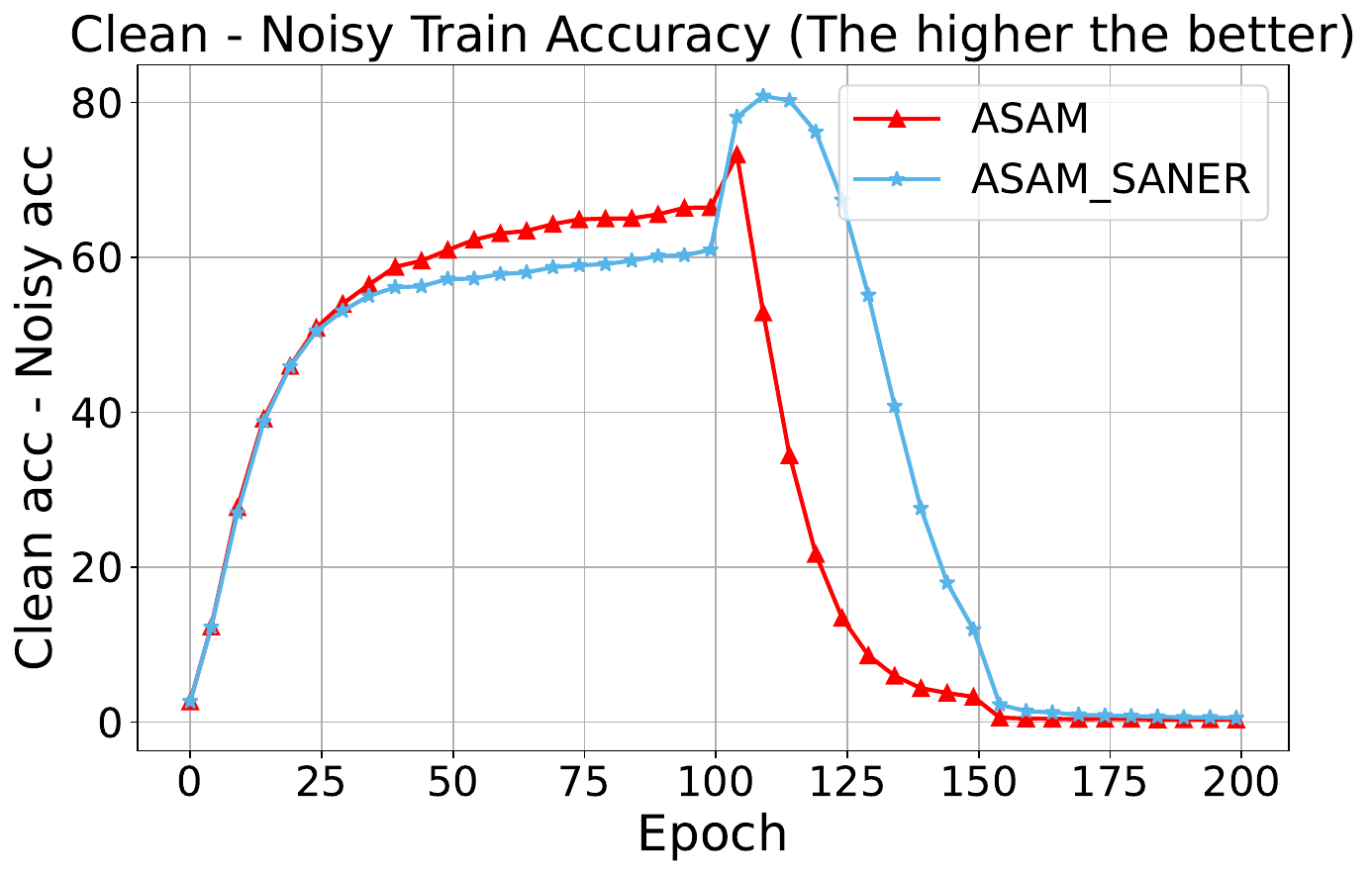}
        \includegraphics[width=0.24\linewidth]{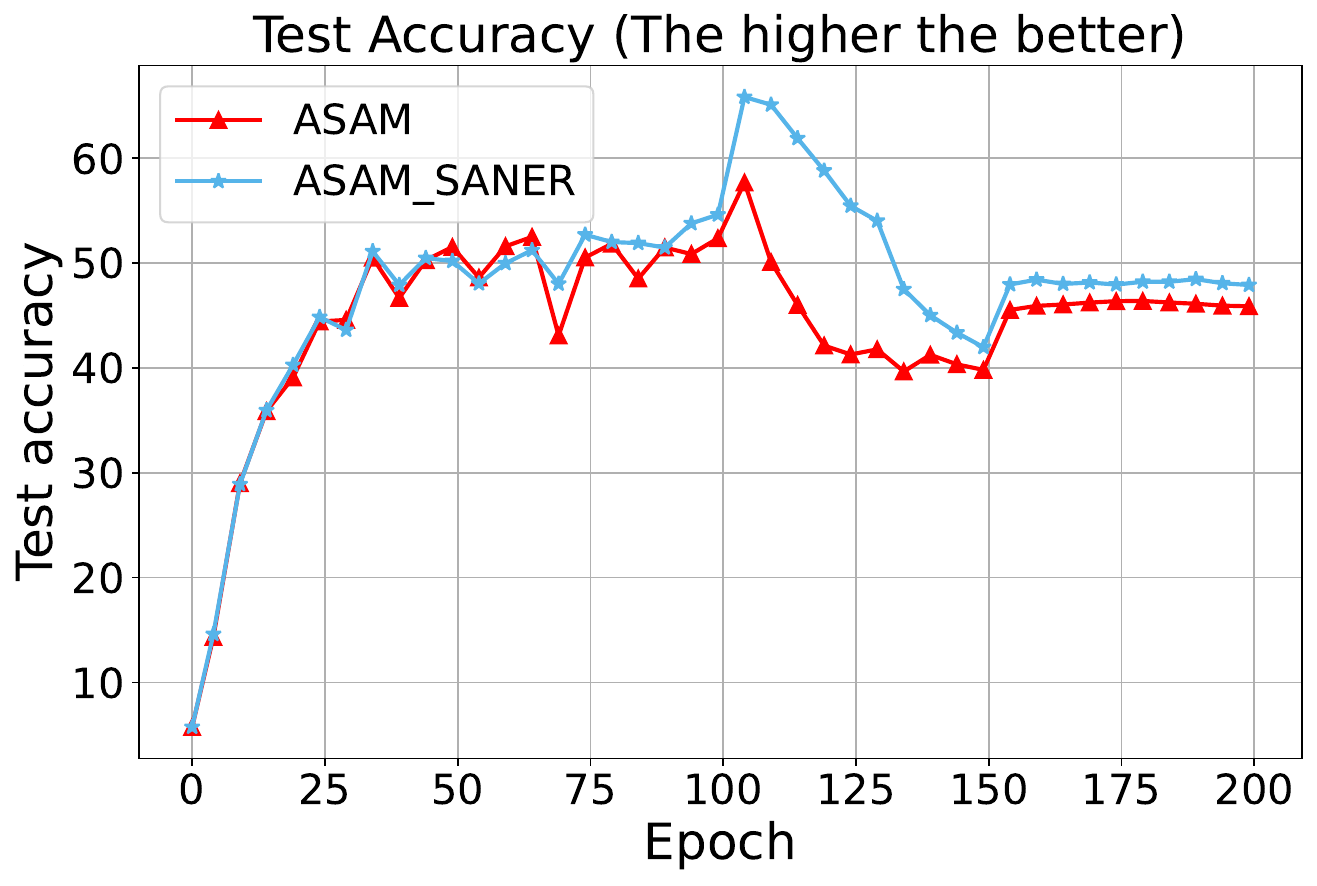}
        \includegraphics[width=0.24\linewidth]{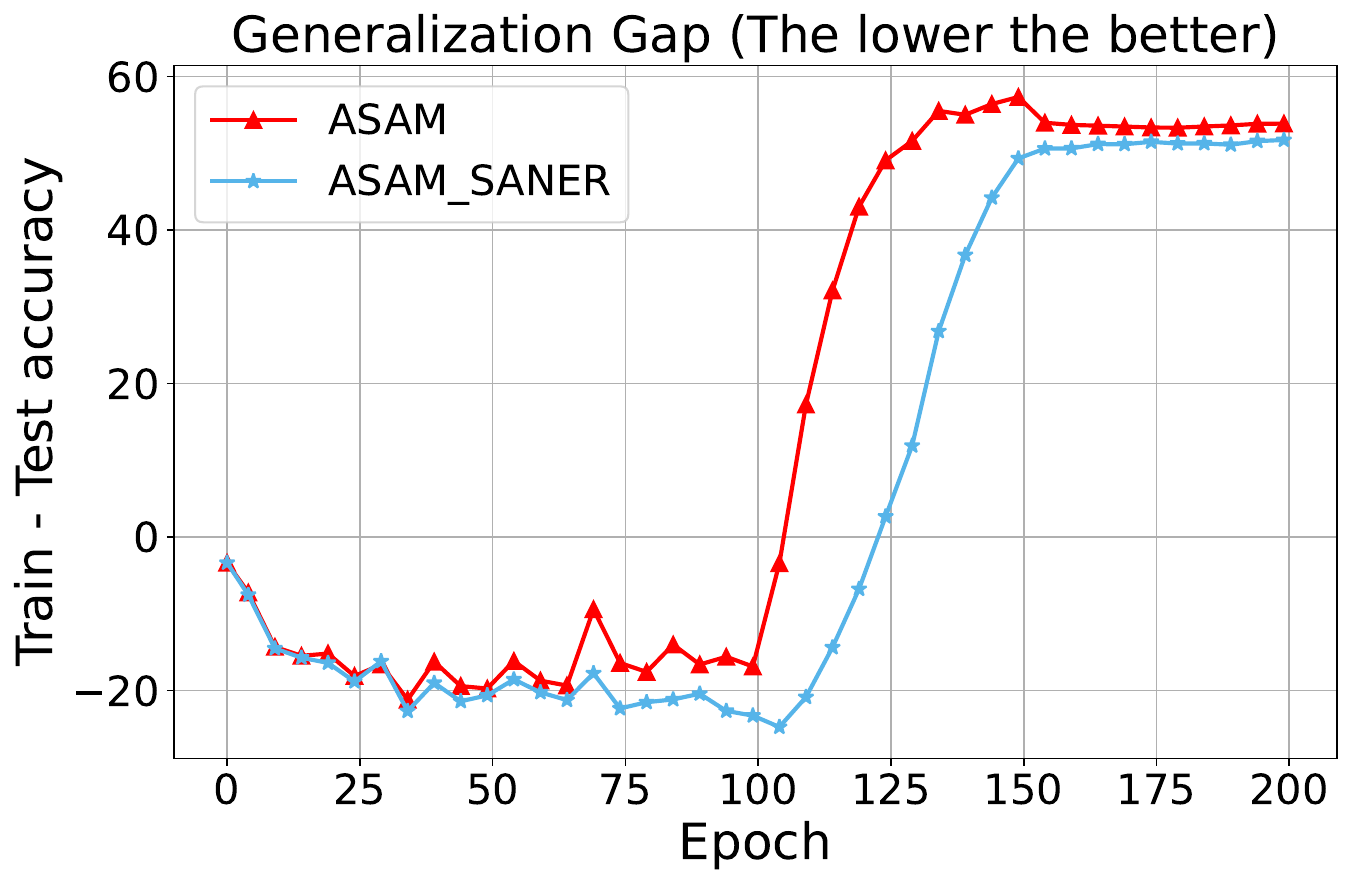}
        \caption{ASAM}
    \end{subfigure}
    \vspace{3pt}
    
    \begin{subfigure}{\linewidth}
        \centering
        \includegraphics[width=0.24\linewidth]{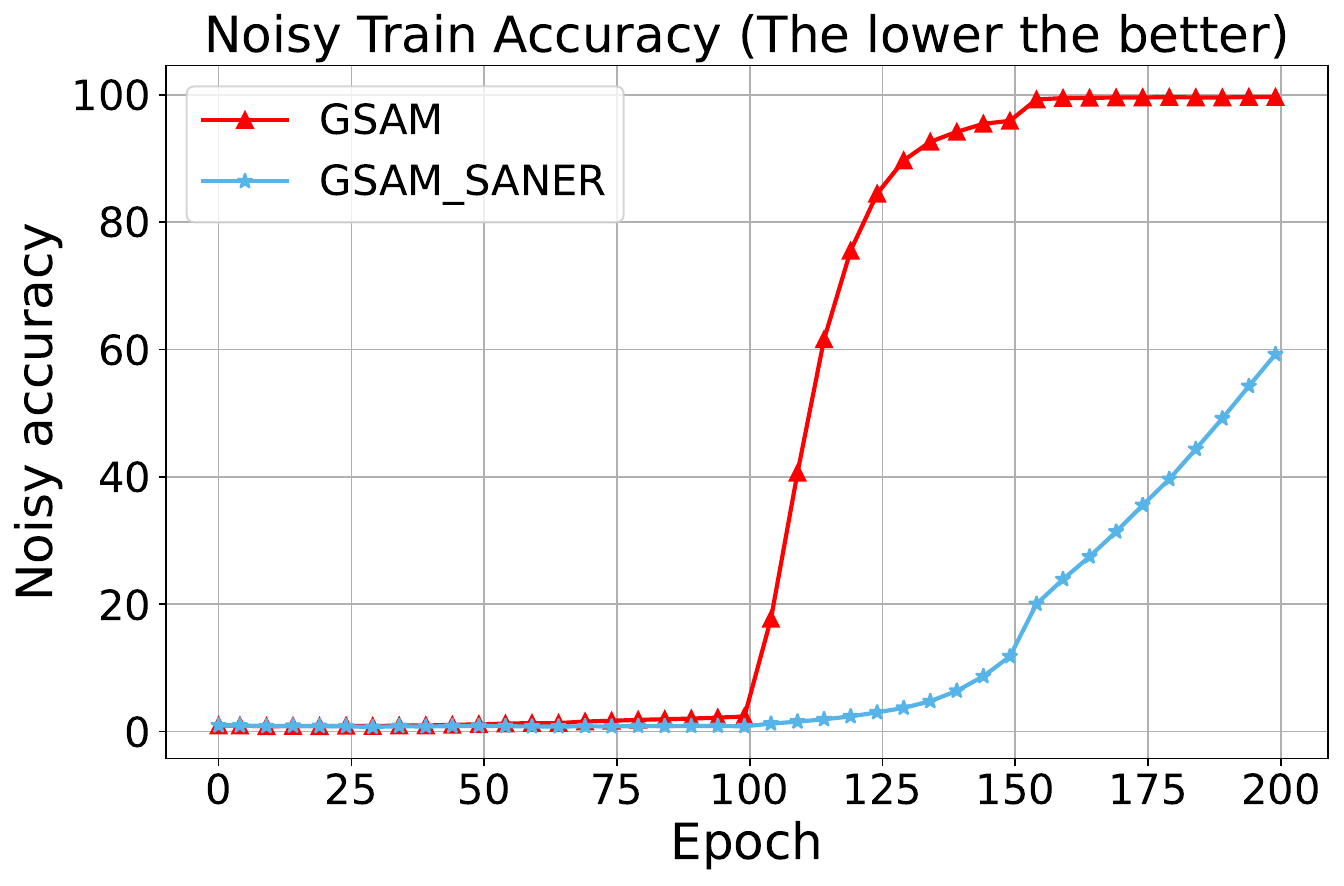}
        \includegraphics[width=0.24\linewidth]{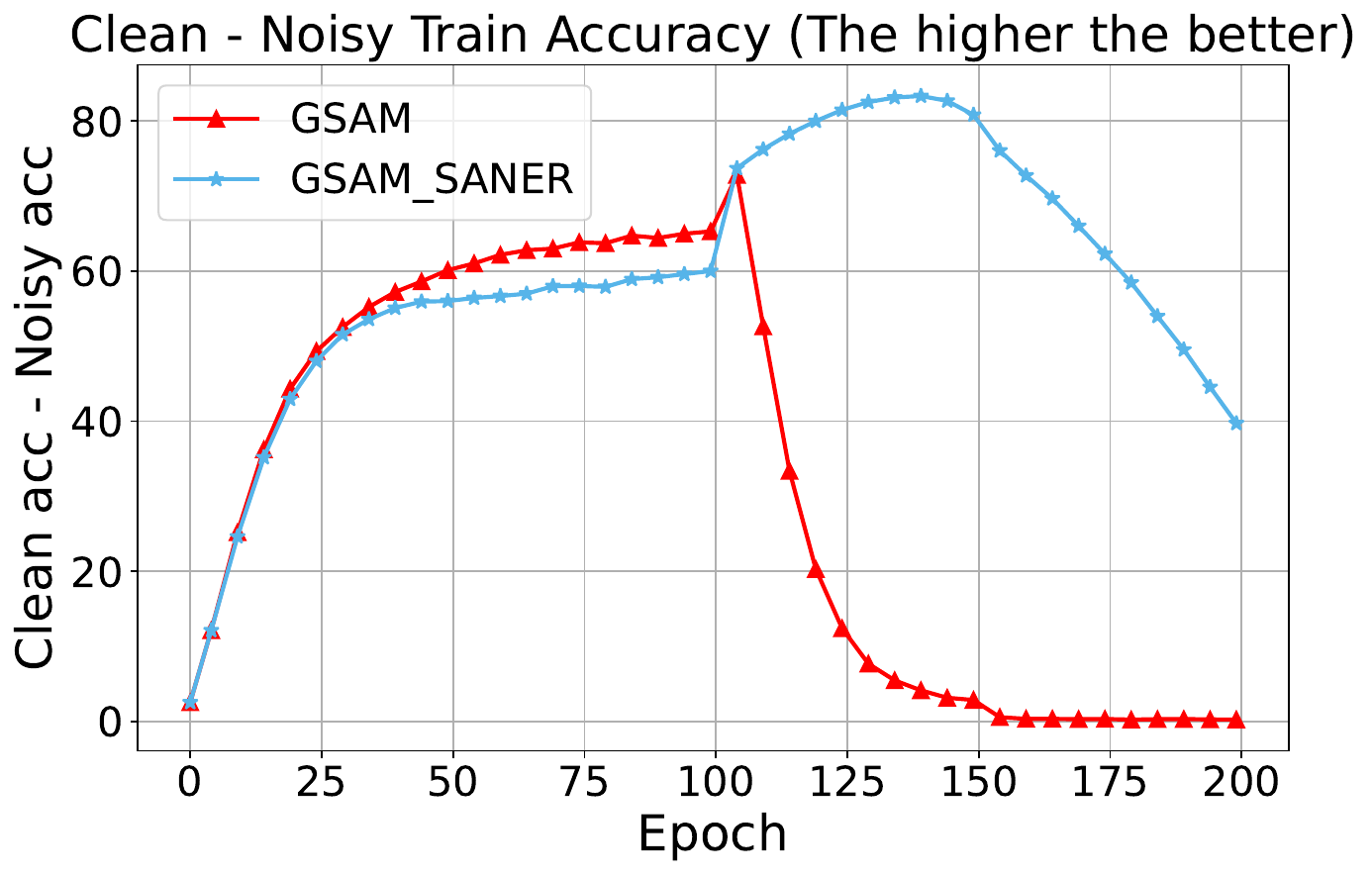}
        \includegraphics[width=0.24\linewidth]{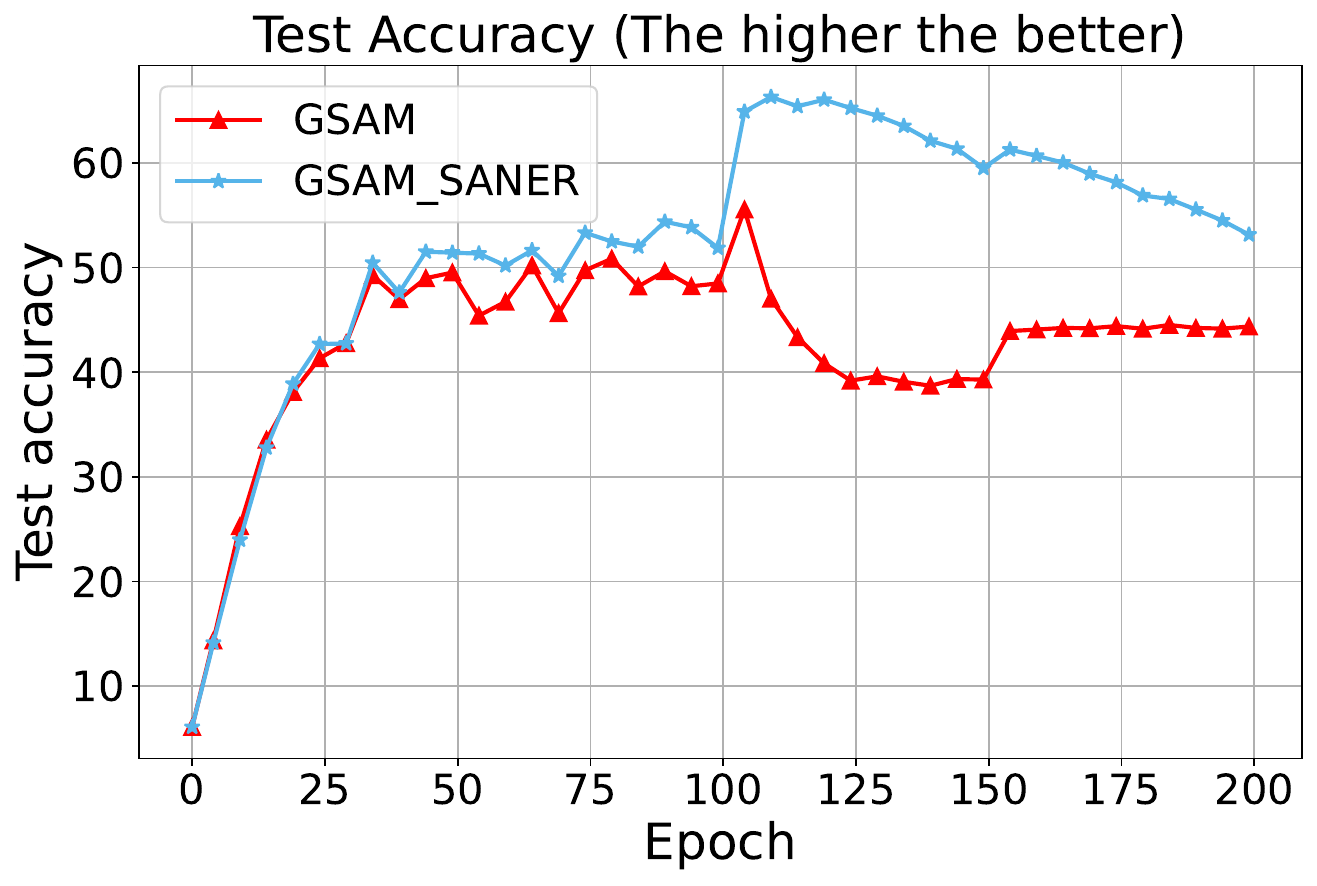}
        \includegraphics[width=0.24\linewidth]{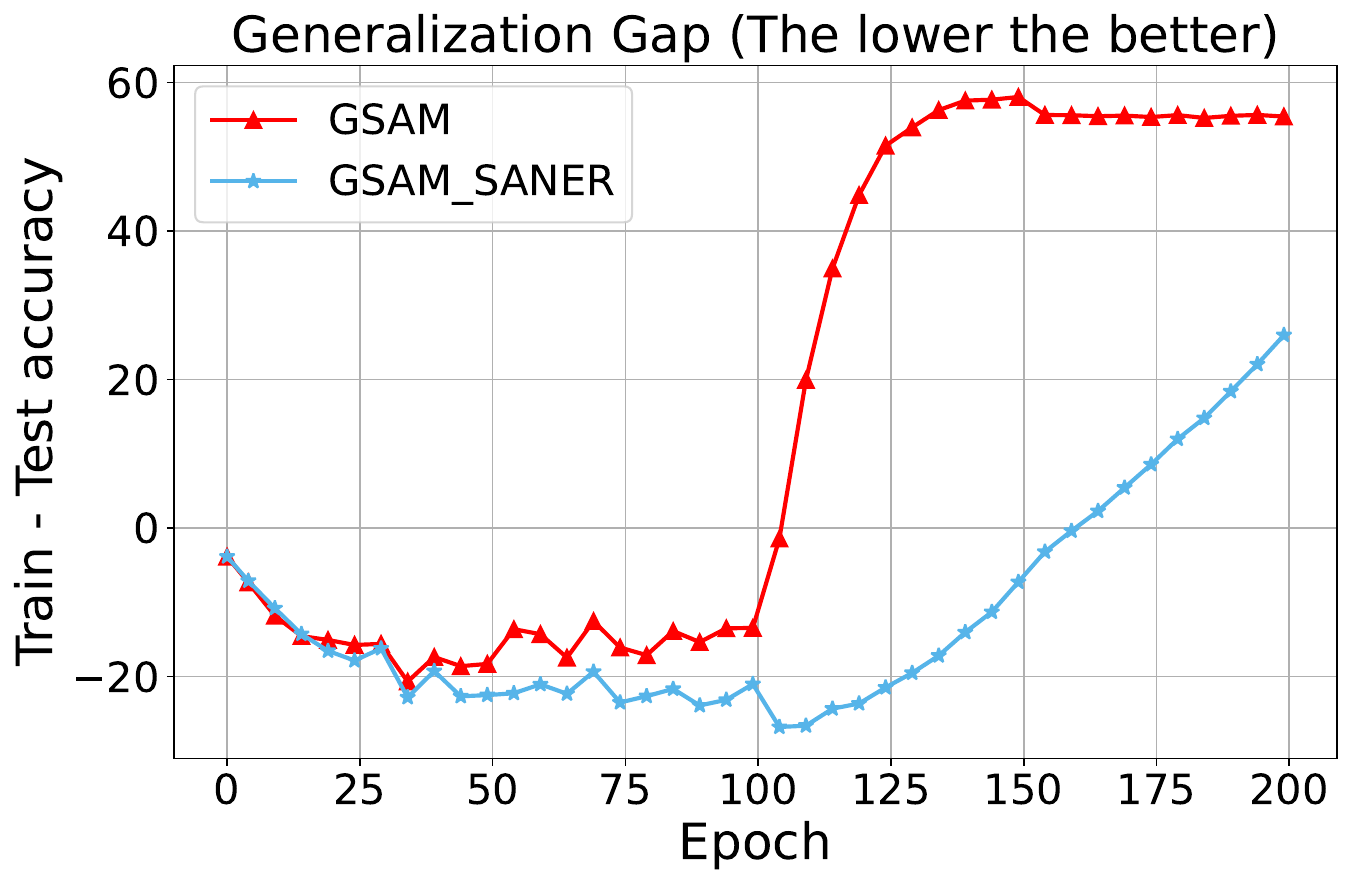}
        \caption{GSAM}
    \end{subfigure}
    \vspace{3pt}

    \begin{subfigure}{\linewidth}
        \centering
        \includegraphics[width=0.24\linewidth]{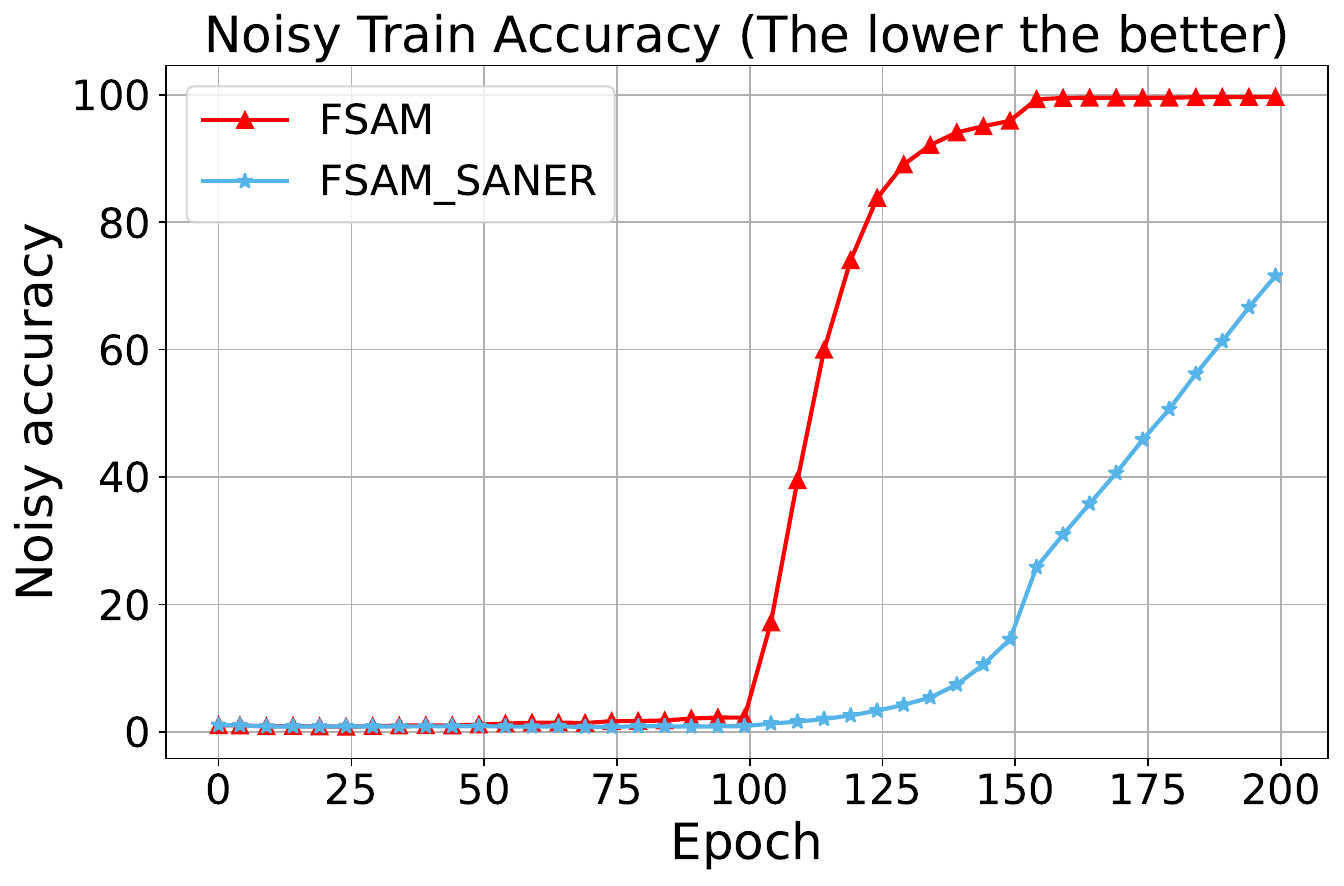}
        \includegraphics[width=0.24\linewidth]{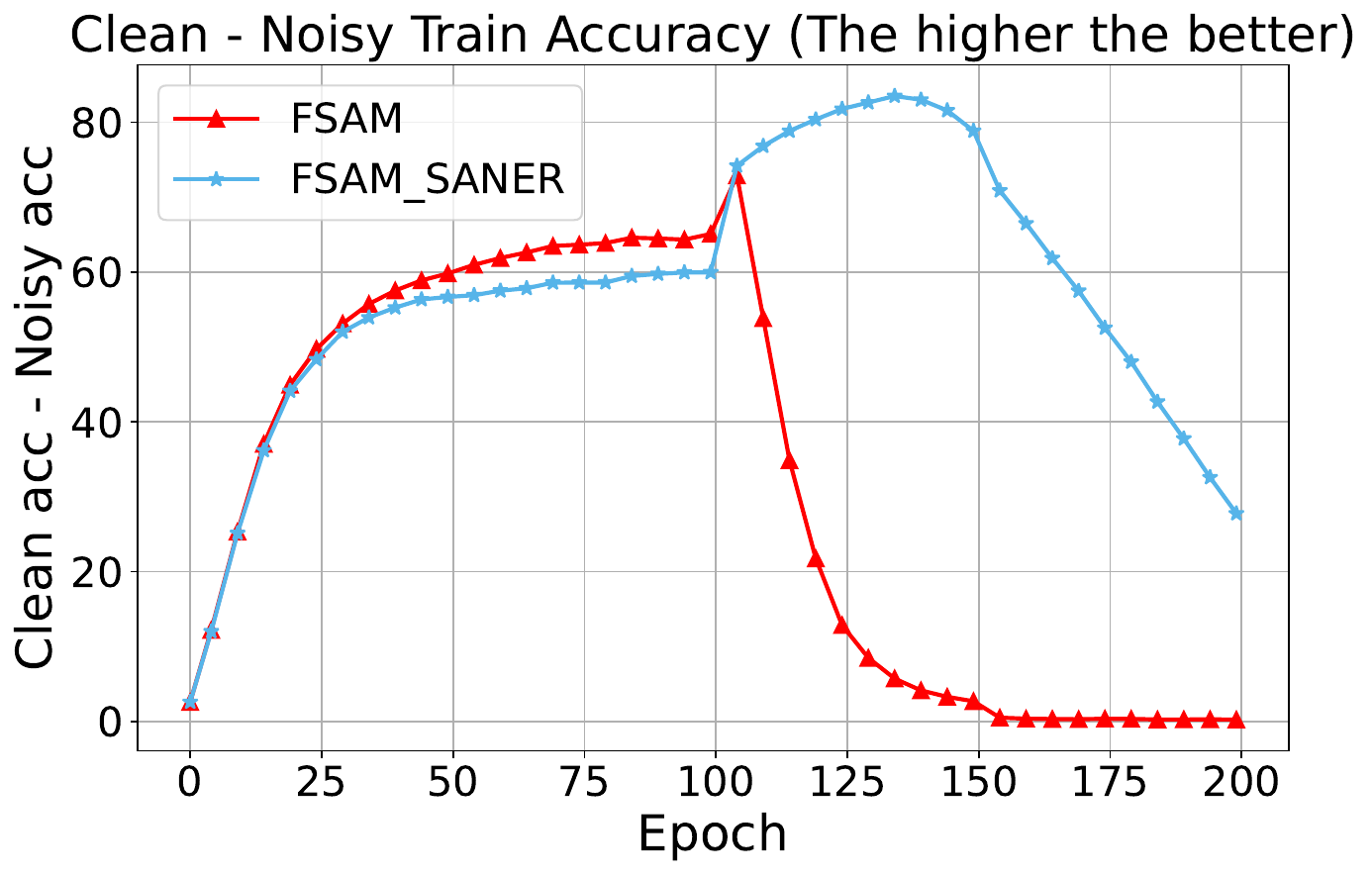}
        \includegraphics[width=0.24\linewidth]{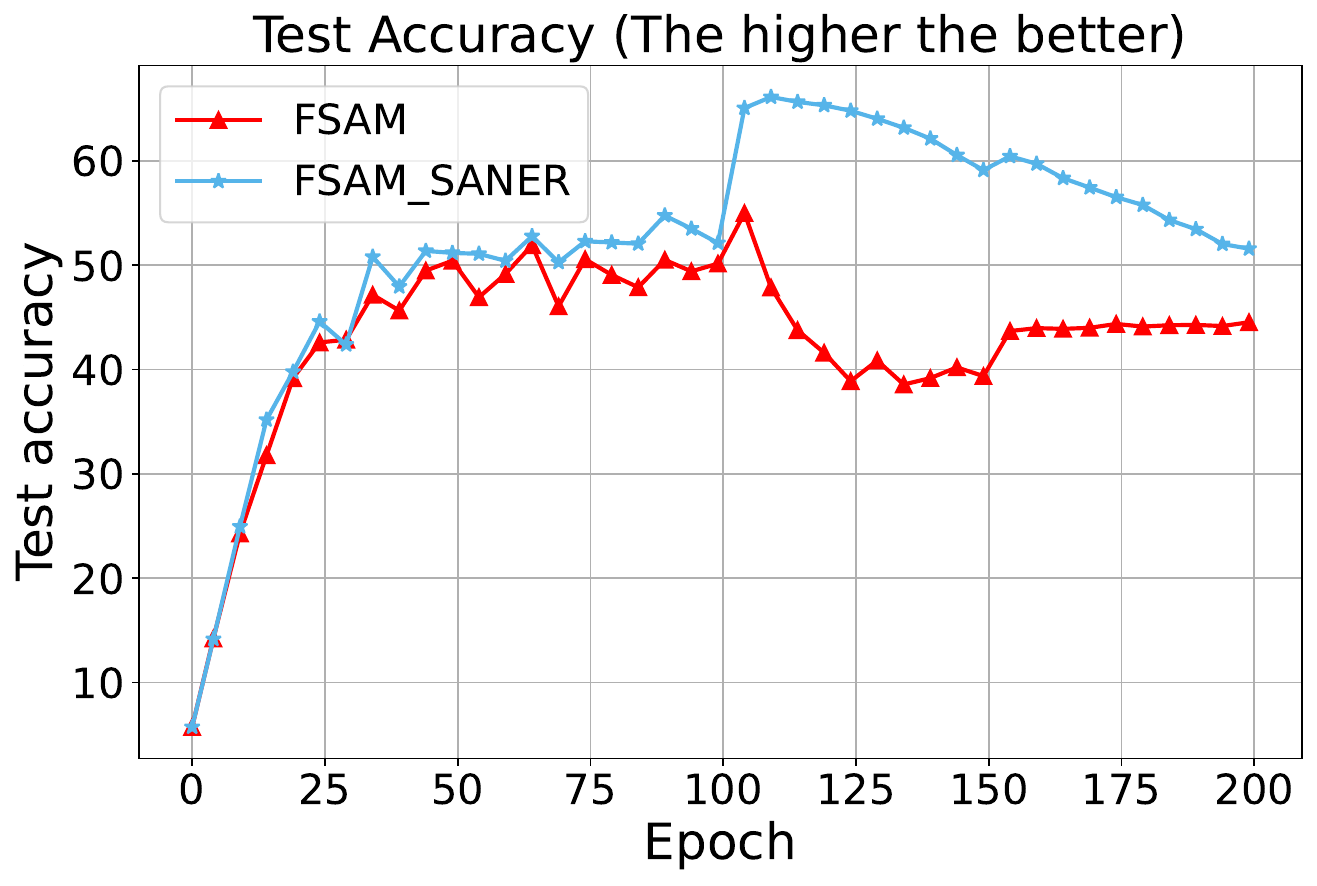}
        \includegraphics[width=0.24\linewidth]{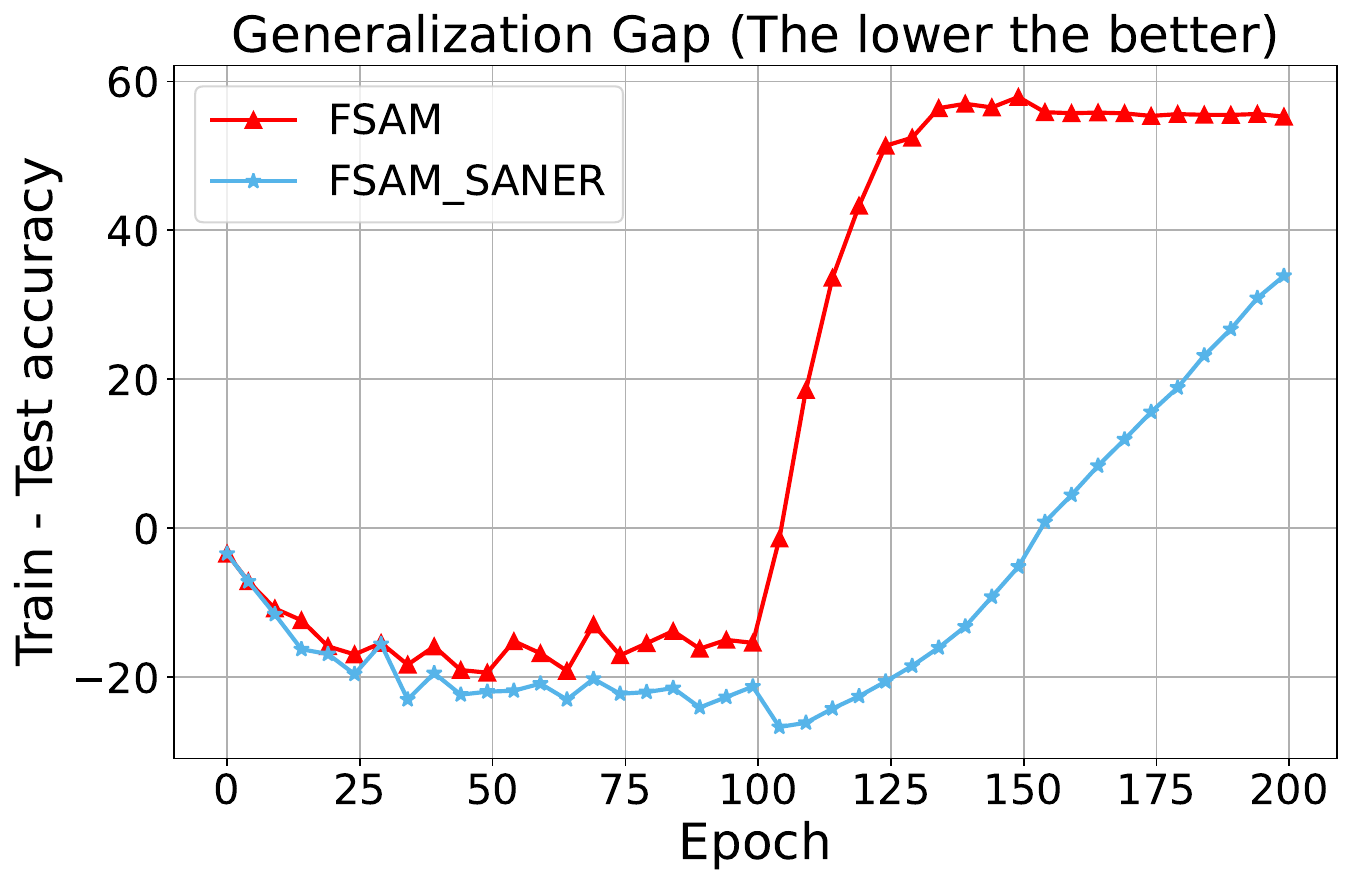}
        \caption{FSAM}
    \end{subfigure}
    \vspace{3pt}

    \begin{subfigure}{\linewidth}
        \centering
        \includegraphics[width=0.24\linewidth]{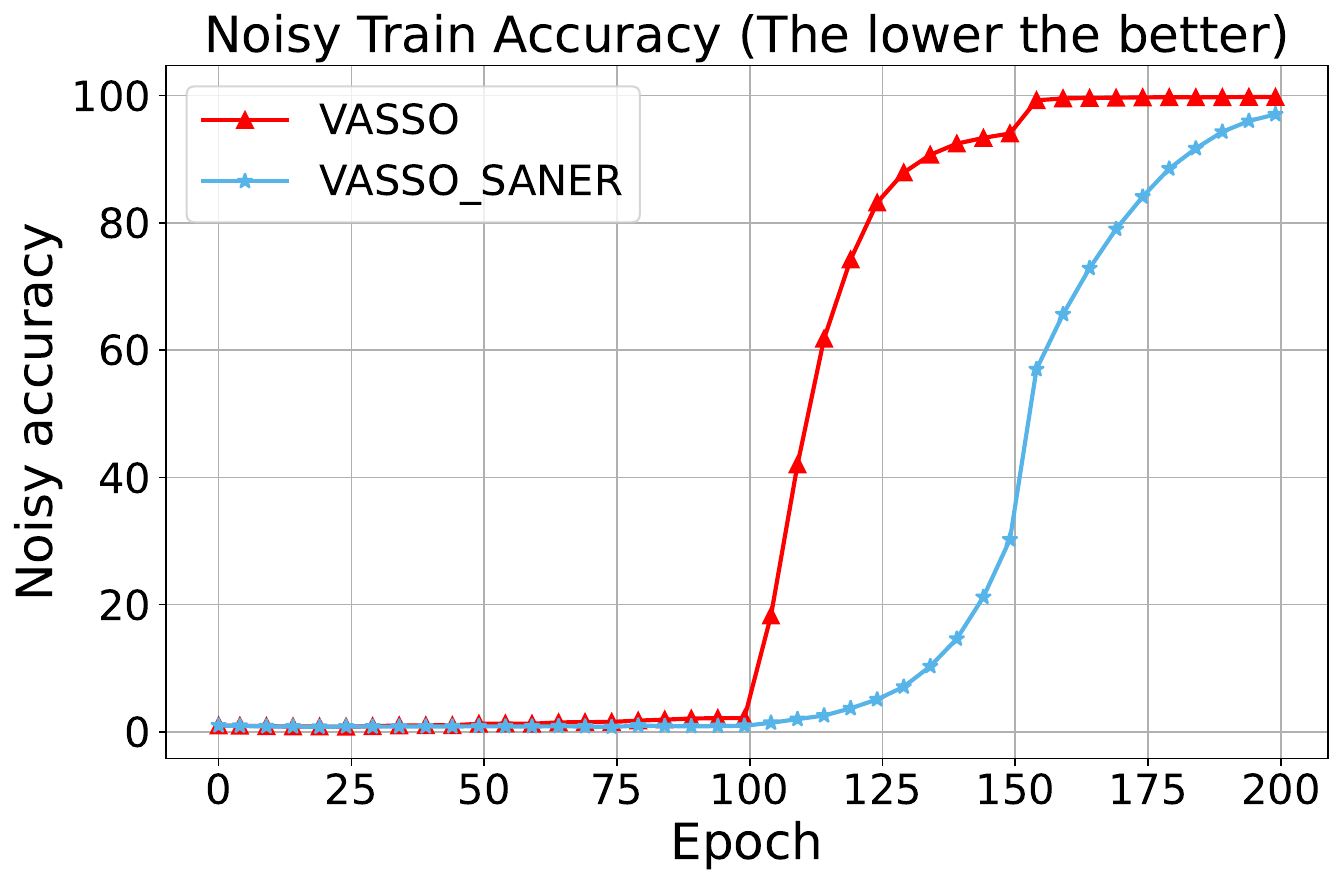}
        \includegraphics[width=0.24\linewidth]{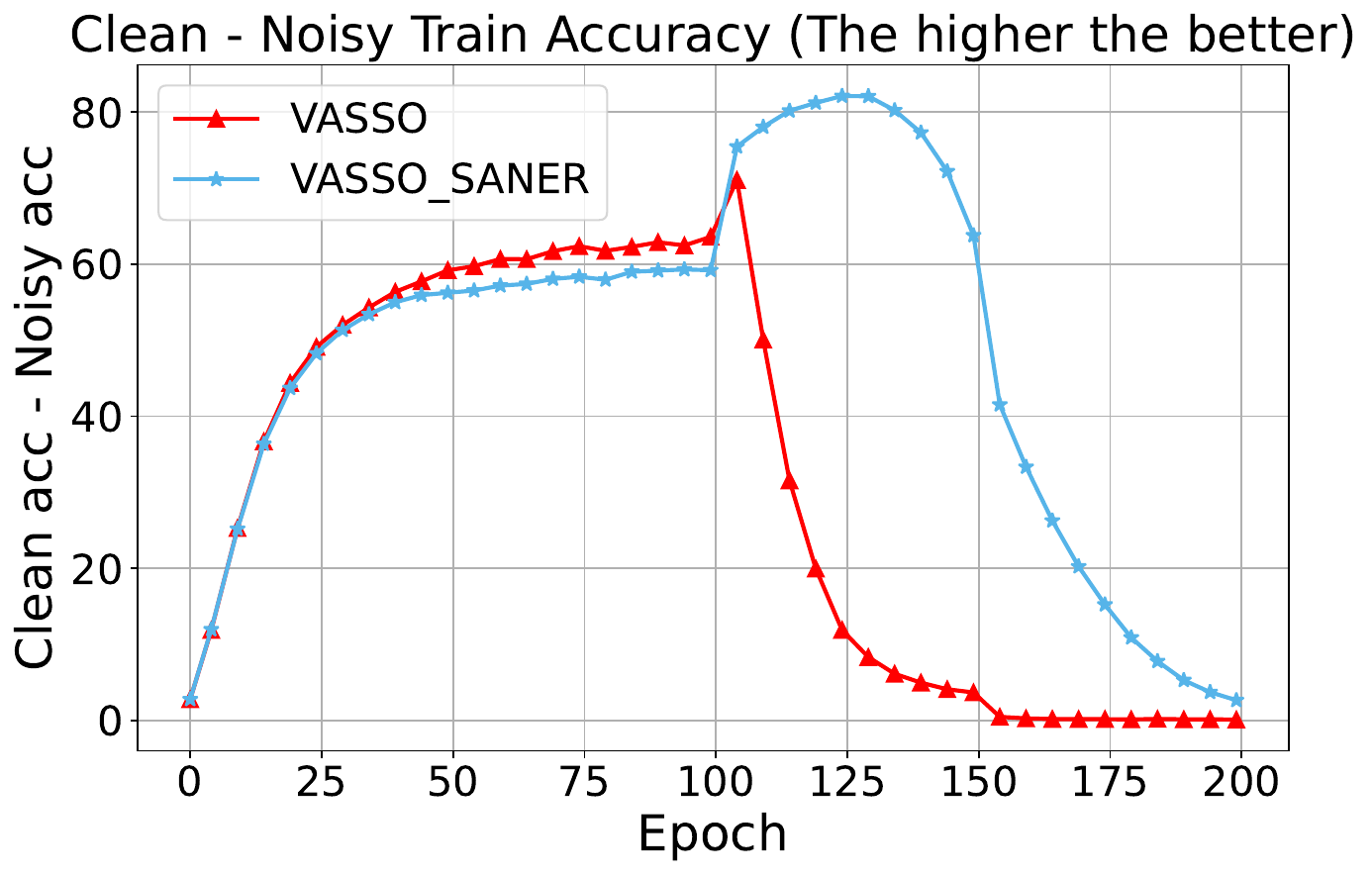}
        \includegraphics[width=0.24\linewidth]{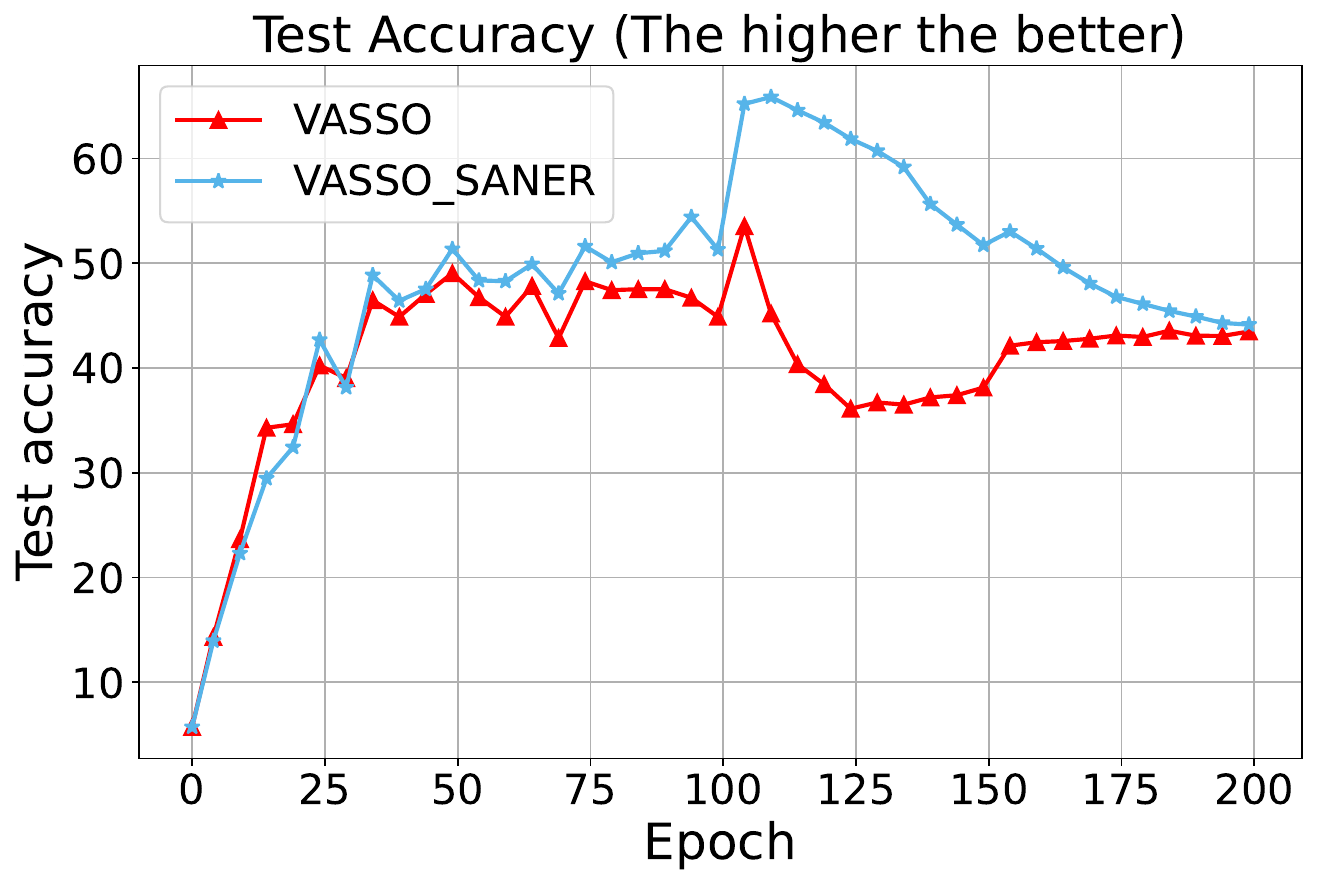}
        \includegraphics[width=0.24\linewidth]{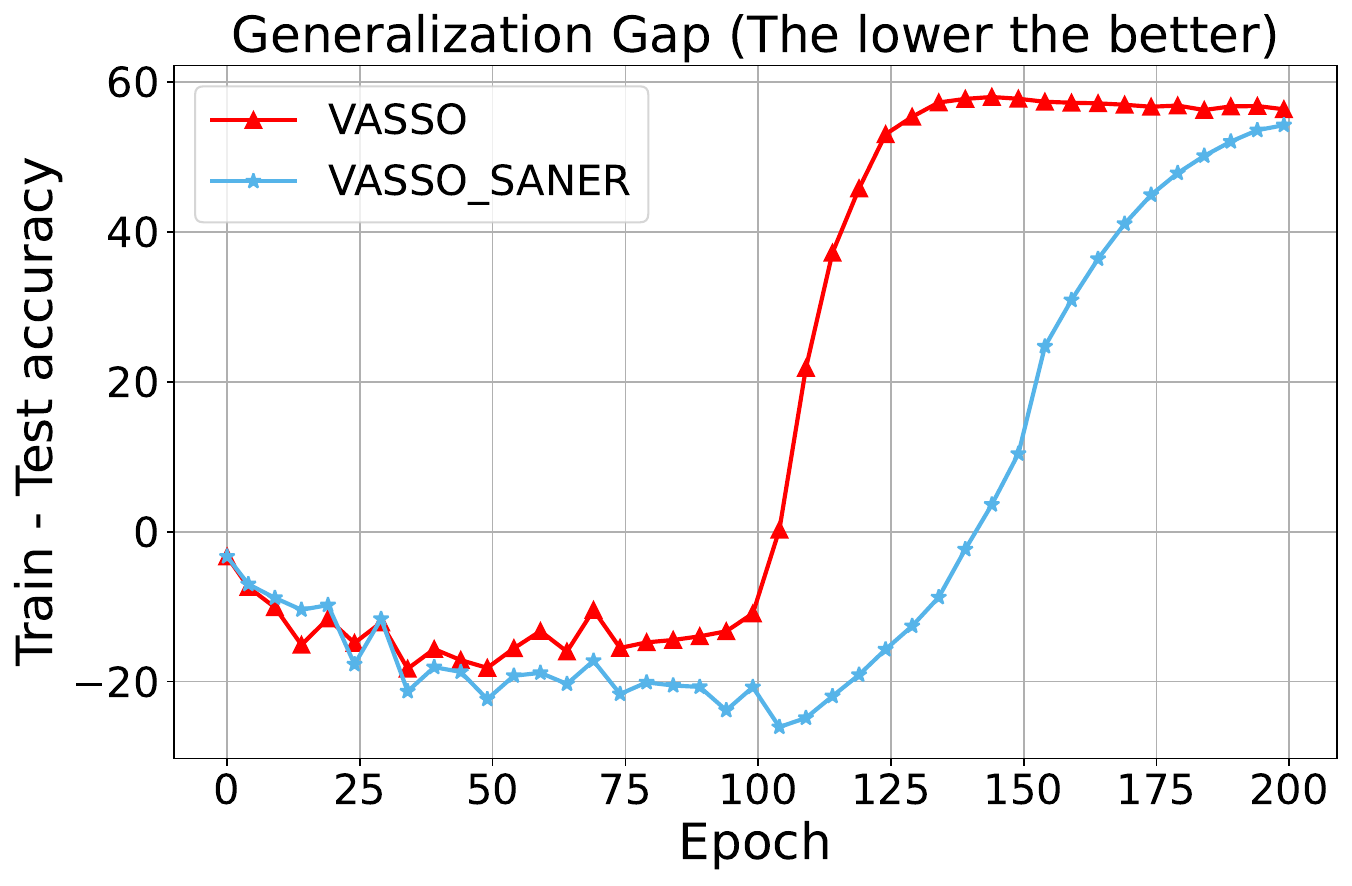}
        \caption{VaSSO}
    \end{subfigure}

    \caption{Performance comparison of ASAM, GSAM, FSAM, and VaSSO (with and without SANER) trained on ResNet18 with CIFAR-100 under 25\% label noise. The columns represent (from left to right): noisy training accuracy, gap between clean and noisy accuracy, test accuracy, and generalization gap. Overall, integrating SANER with these SAM variants slows the noisy fitting rate while preserving the clean fitting rate.}
    \label{fig:sam_like_comparison}
\end{figure*}

\textbf{Noisy train accuracy.} As illustrated in Figure~\ref{fig:sam_like_comparison}, the integration of SANER into SAM variants significantly reduces the number of noisy examples that are memorized during training compared to their original variants. This is particularly evident in high-noise scenarios such as 50\%, where the noisy fitting curve rises more gradually in SANER-integrated models compared to their original counterparts. This indicates that SANER helps slow down the memorization of noisy labels, allowing the models to focus more on clean data, which leads to better generalization.

\subsection{Effect of Increasing ResNet18 Width}
\begin{figure*}[ht]
    \centering
    \begin{subfigure}{\linewidth}
        \centering
        \includegraphics[width=0.24\linewidth]{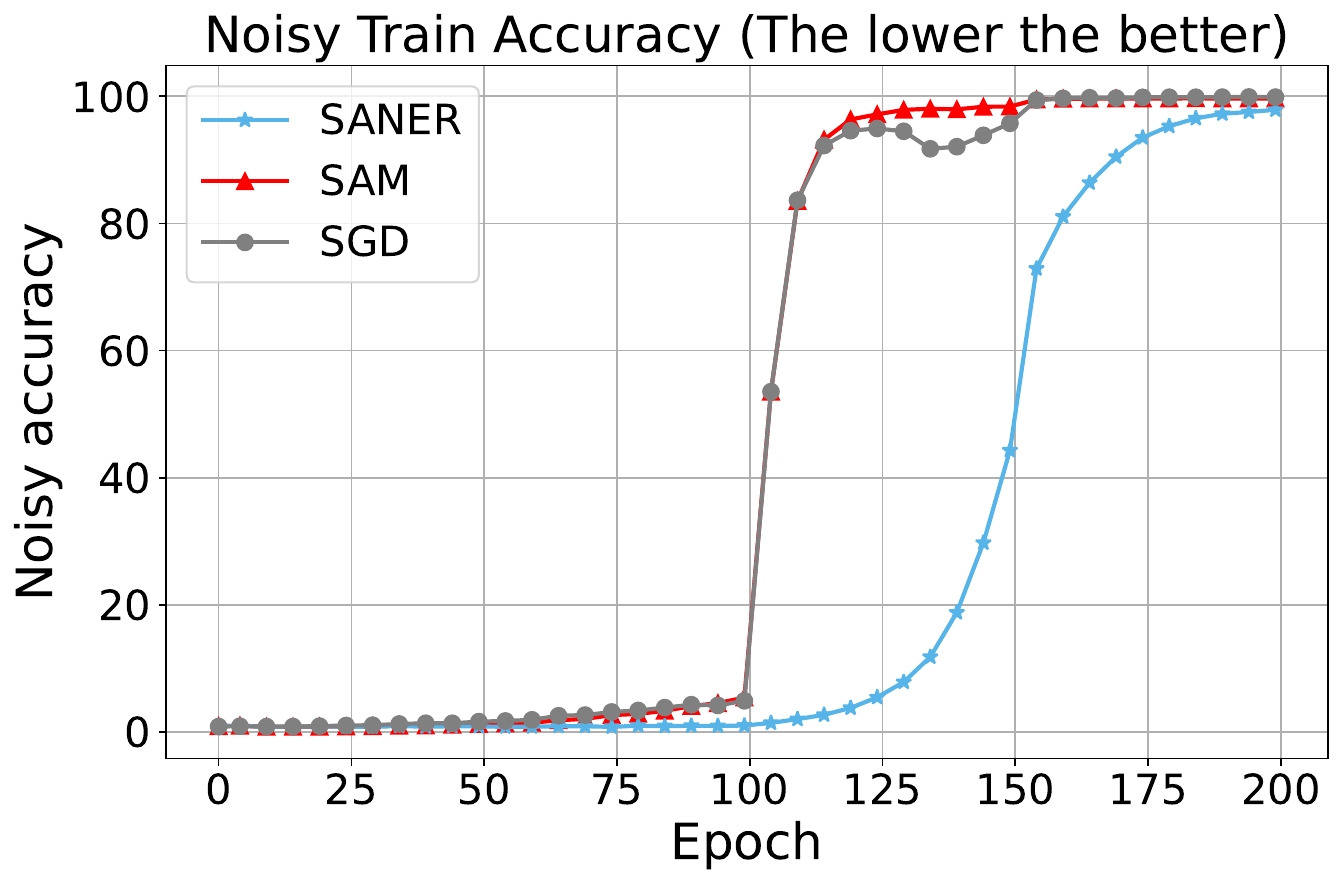}
        \includegraphics[width=0.24\linewidth]{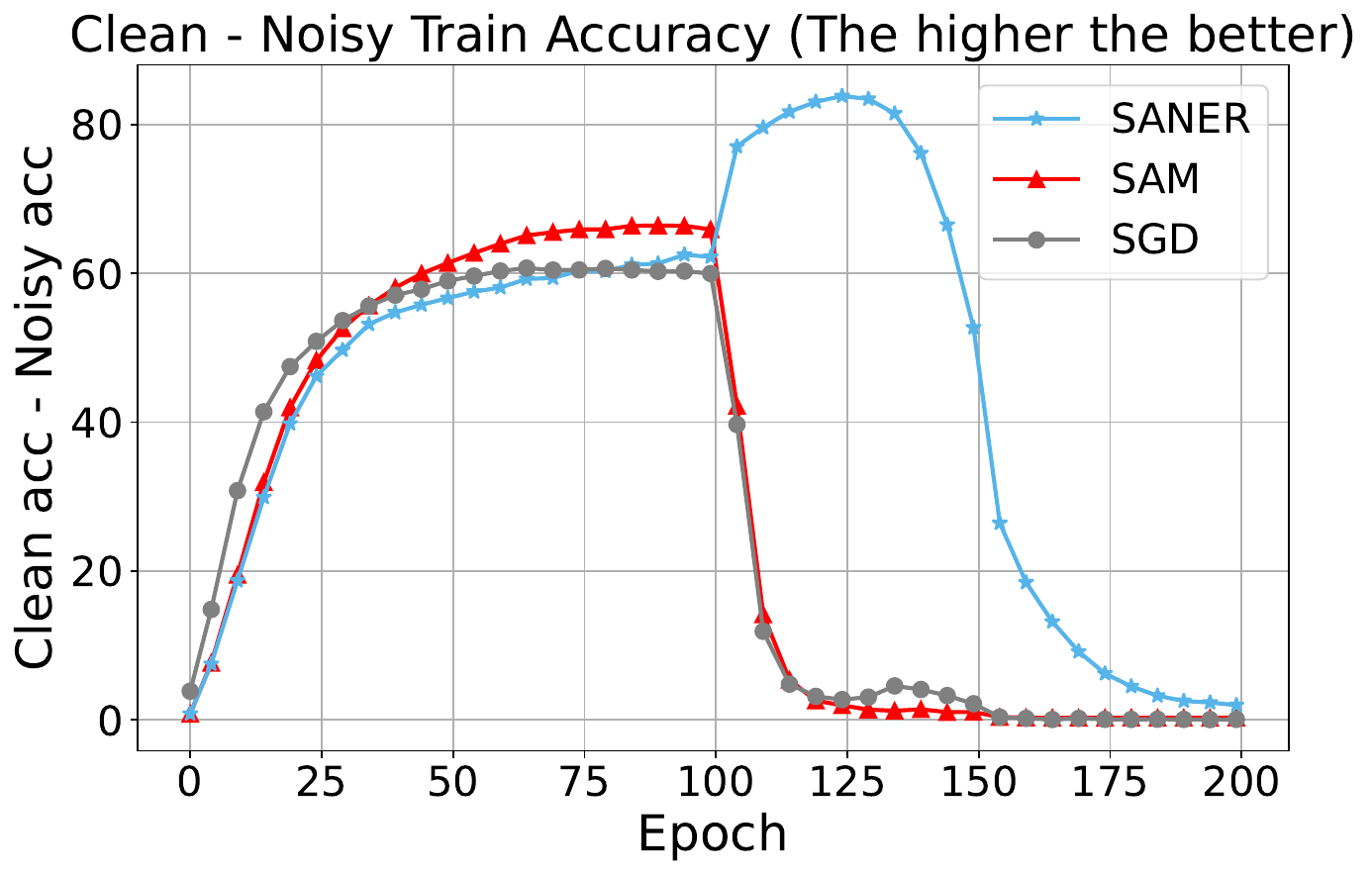}
        \includegraphics[width=0.24\linewidth]{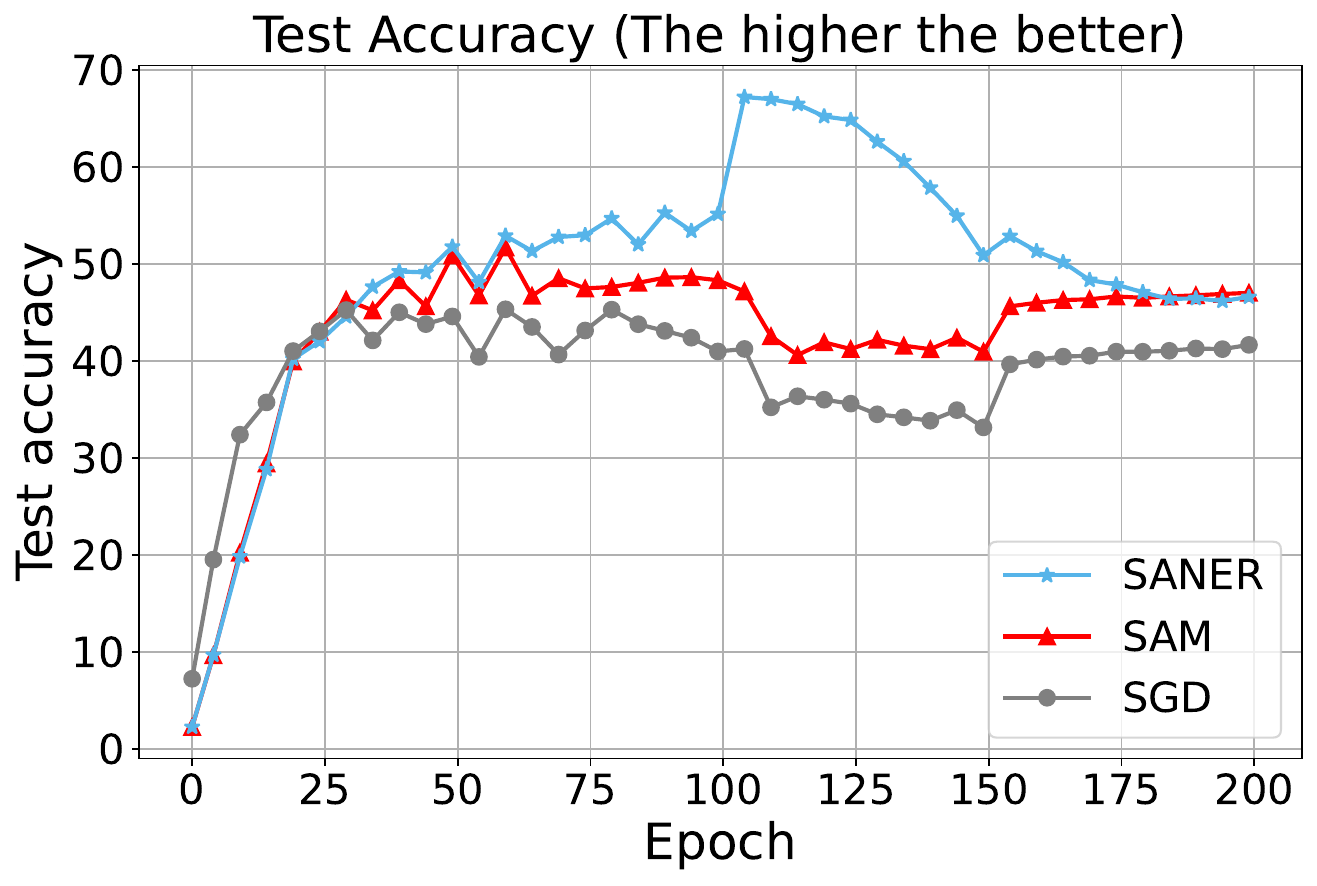}
        \includegraphics[width=0.24\linewidth]{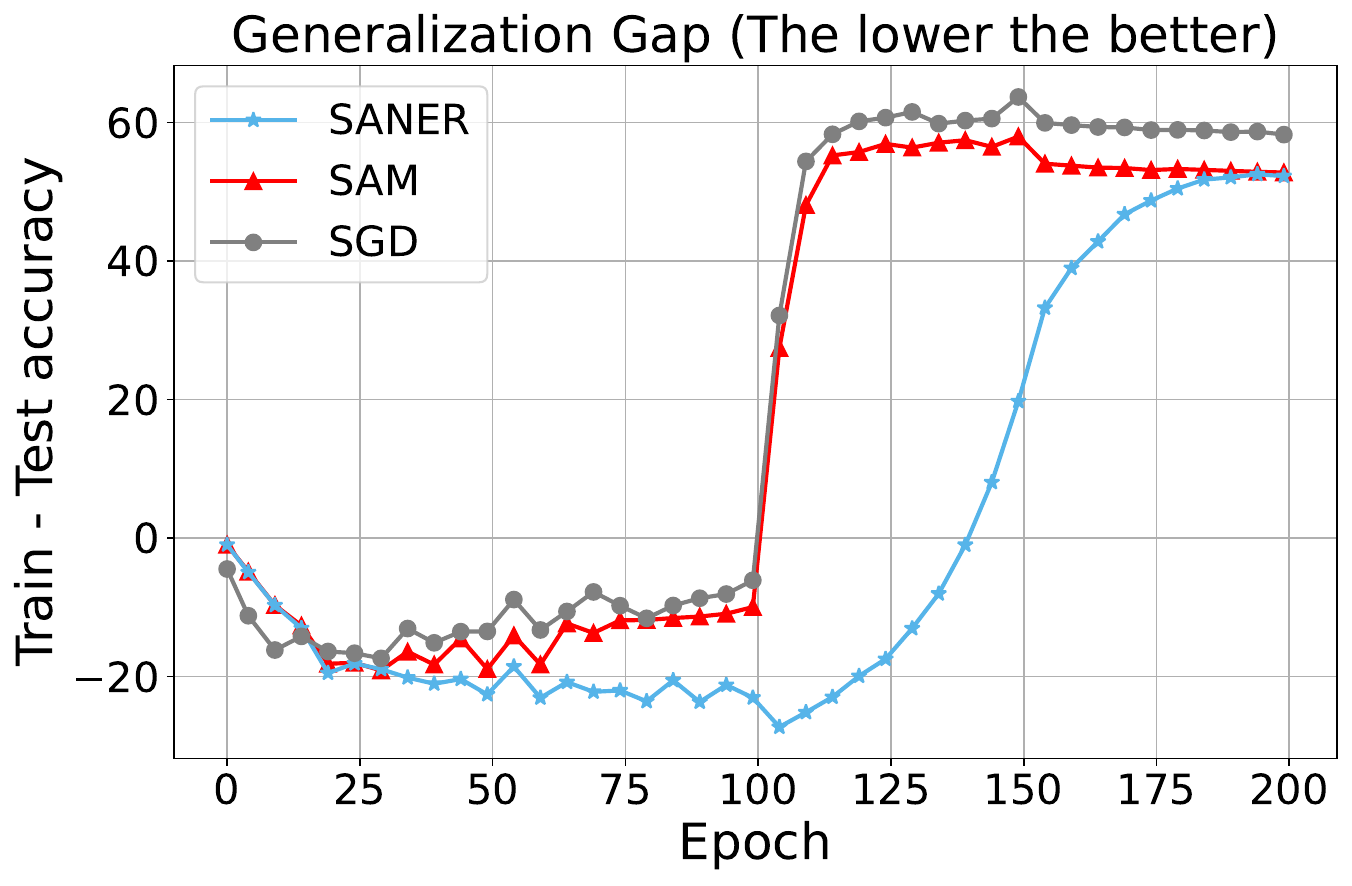}
        \caption{ResNet18 (Width = 1.5)}
        \label{fig:sam_vs_sgd_wi=1.5}
    \end{subfigure}
    \vspace{3pt}
    
    \begin{subfigure}{\linewidth}
        \centering
        \includegraphics[width=0.24\linewidth]{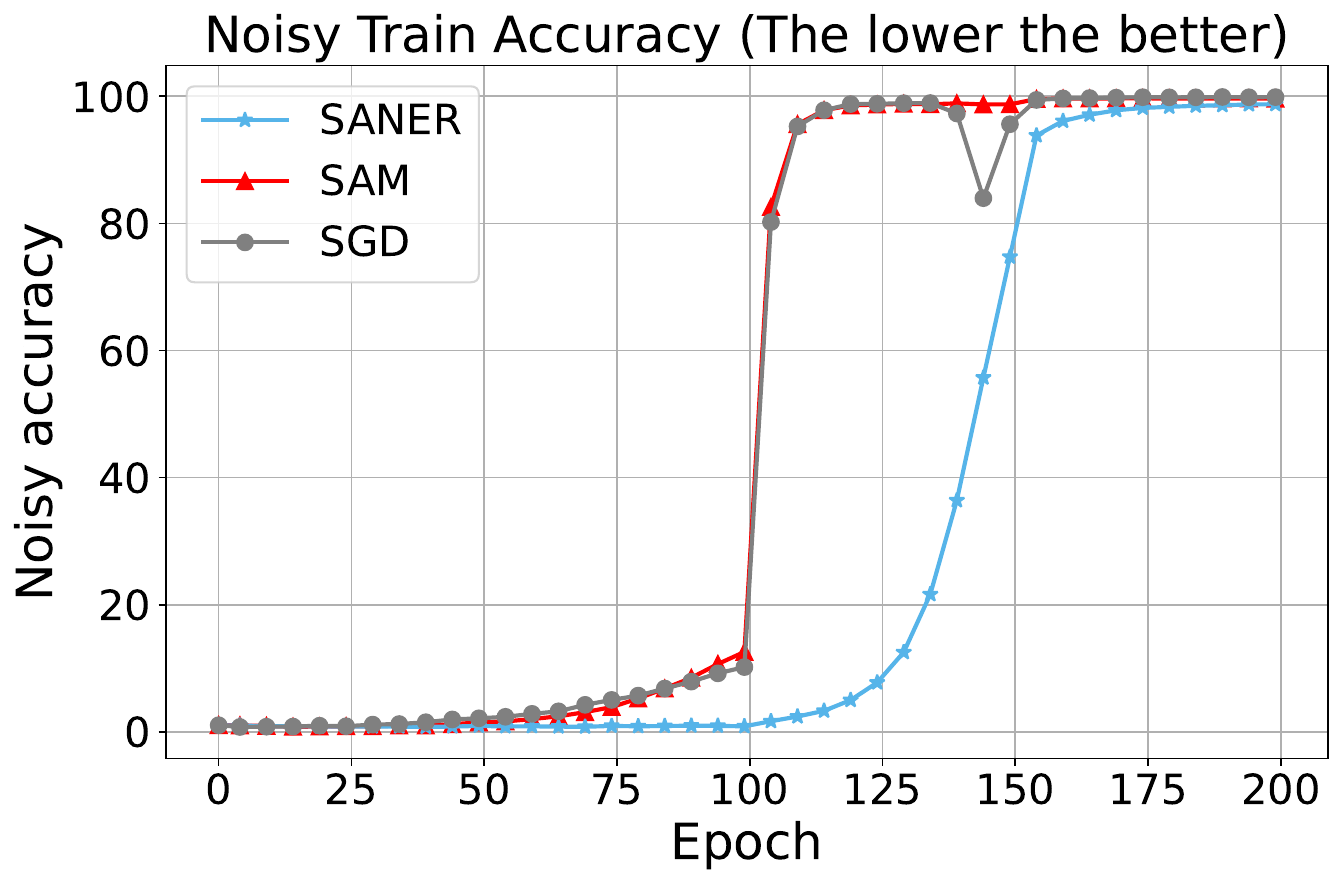}
        \includegraphics[width=0.24\linewidth]{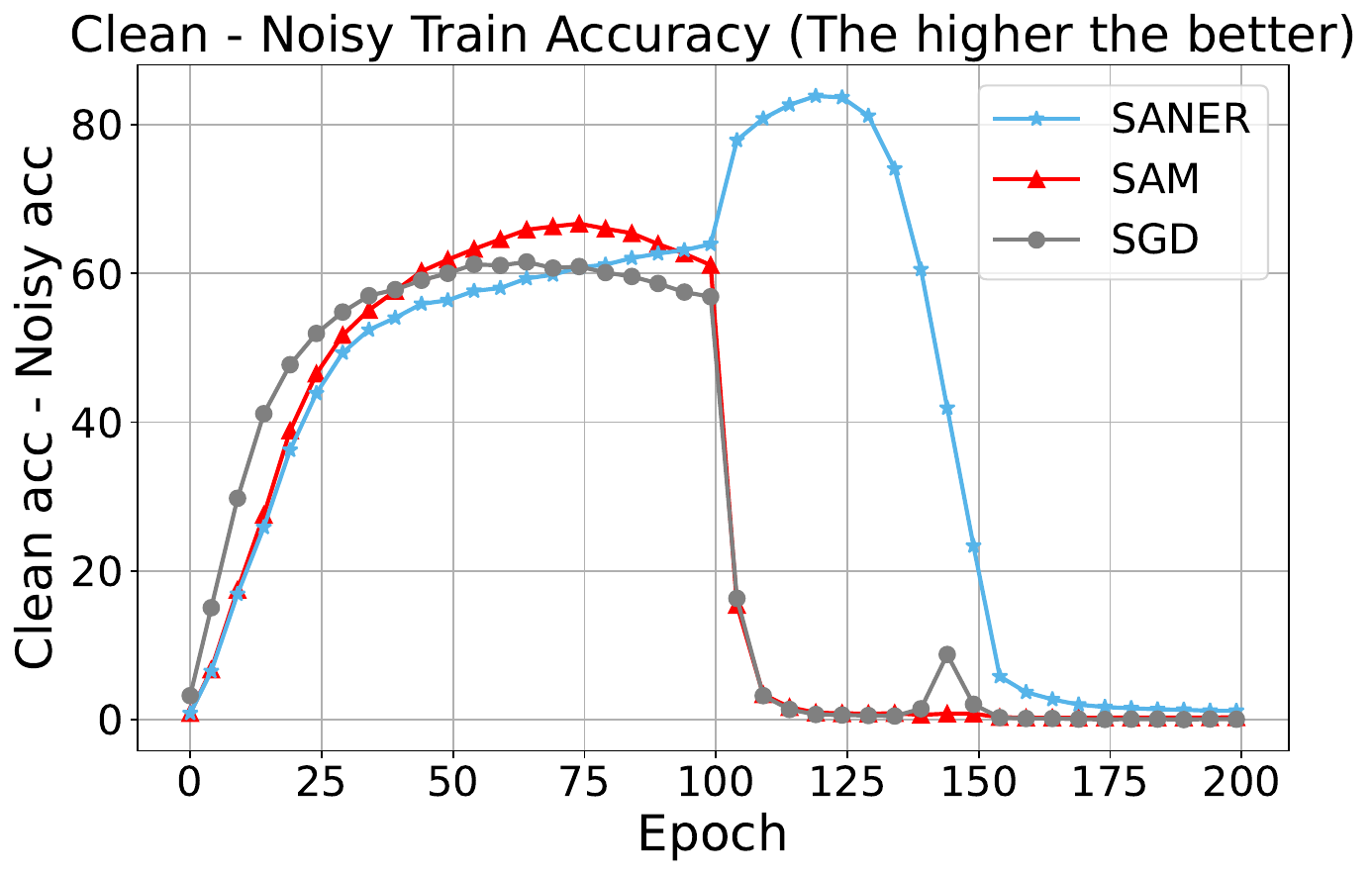}
        \includegraphics[width=0.24\linewidth]{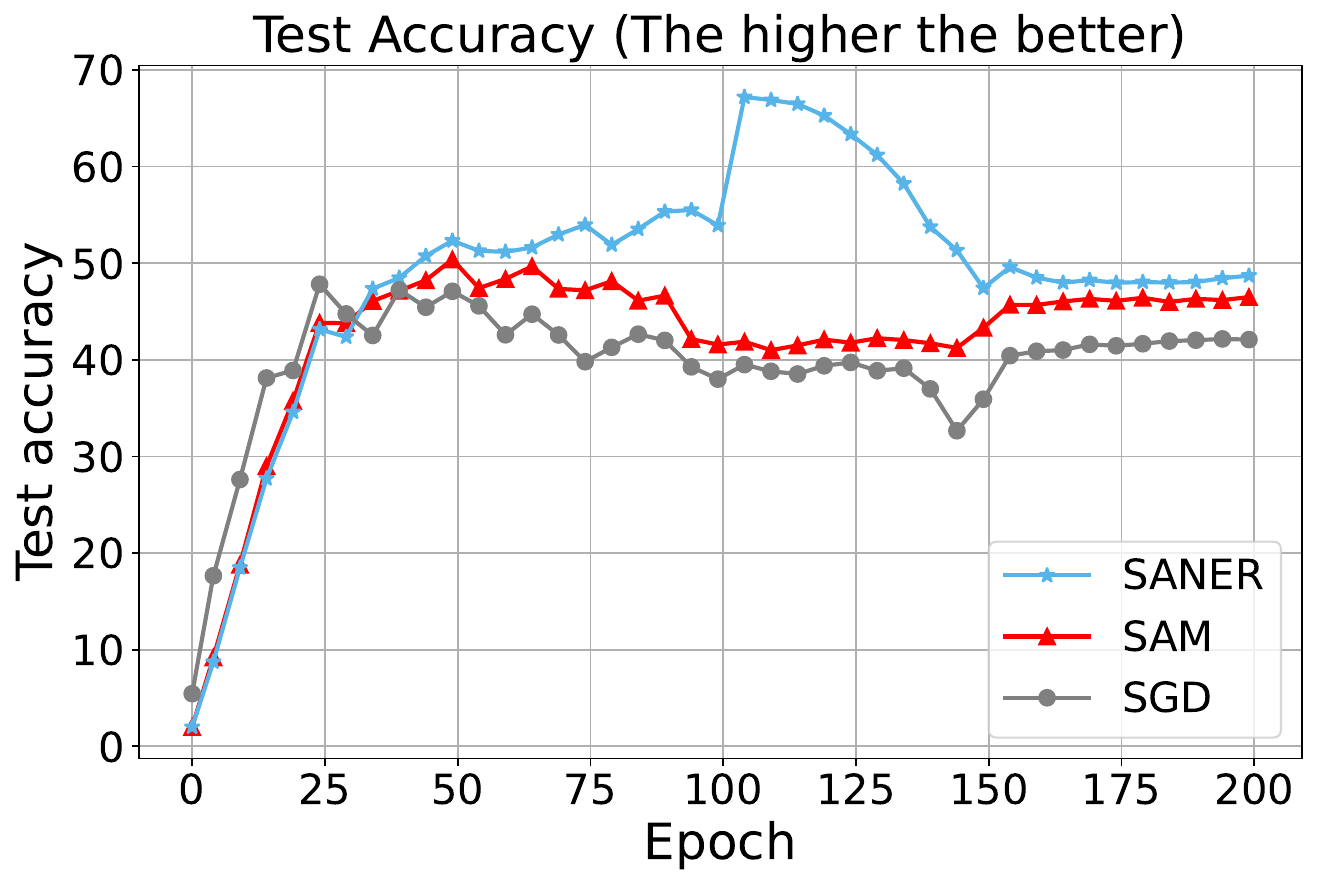}
        \includegraphics[width=0.24\linewidth]{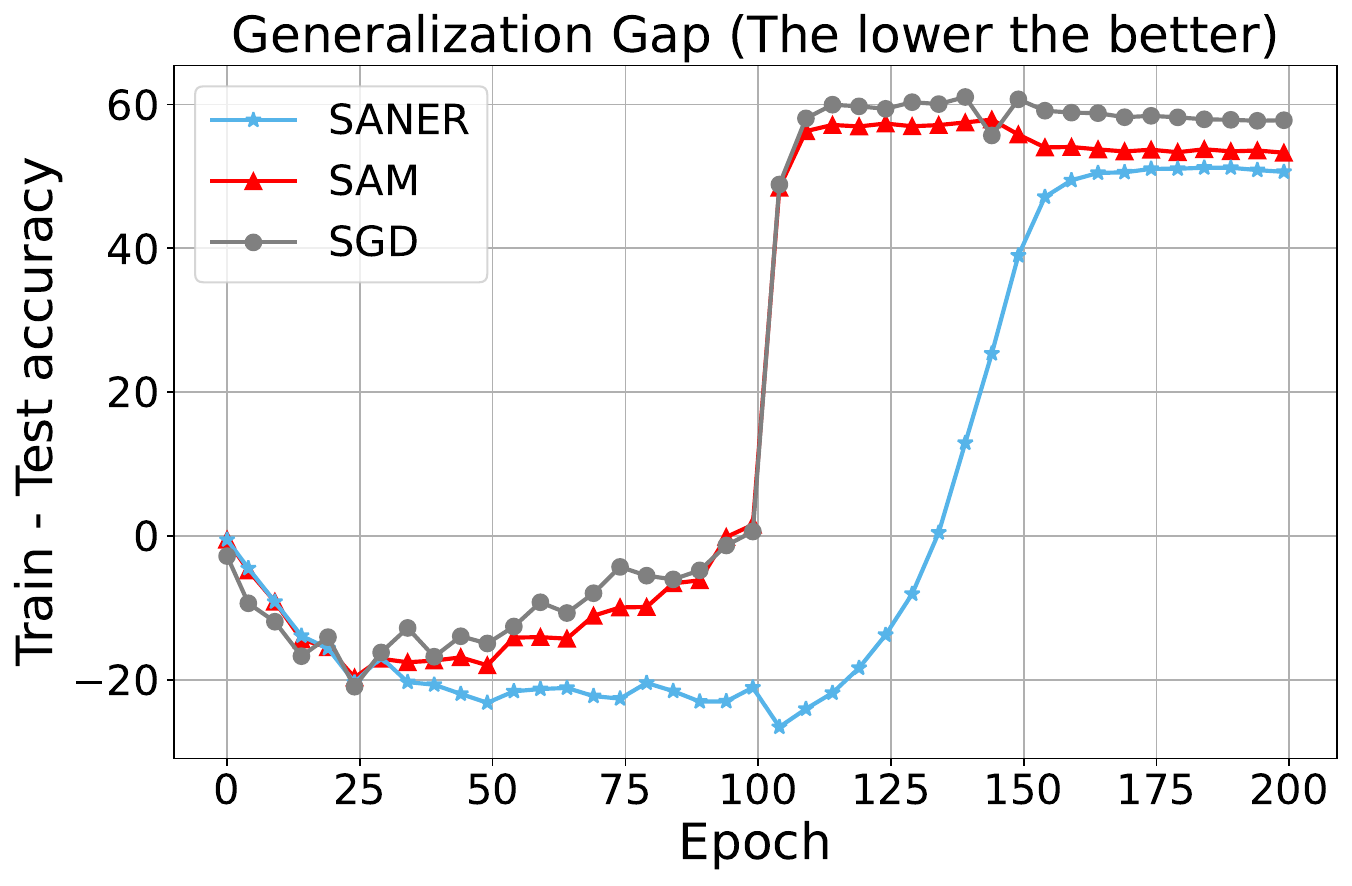}
        \caption{ResNet18 (Width = 2)}
        \label{fig:sam_vs_sgd_wi=2}
    \end{subfigure}

    \caption{Performance comparison of SAM, SGD, and SANER (ours) when \textbf{increasing width} of ResNet18 trained on CIFAR-100 under 50\% label noise. The columns represent (from left to right): noisy training accuracy, gap between clean and noisy accuracy, test accuracy, and generalization gap. The noisy fitting rate of SAM reaches that of SGD, whereas SANER keeps it low for a longer duration, resulting in better performance.}
    \label{fig:different-width-training-process}
\end{figure*}

To demonstrate the effectiveness of our method, we conduct experiments in overfitting-prone scenarios by increasing model parameters, as detailed in main paper. In this section, we visualize the training process under overparameterization by increasing the width of ResNet18 to provide further insights into the fitting rates of SGD, SAM, and SANER. As shown in Figure~\ref{fig:different-width-training-process}, increasing model width enhances overfitting, causing SAM to match the noisy fitting rate of SGD. In contrast, SANER maintains a slower noisy fitting rate while preserving the clean fitting rate, allowing the model to better leverage overparameterization and achieve higher test accuracy.

\subsection{Various Architectures}

\begin{figure*}[ht]
    \centering
    \begin{subfigure}{\linewidth}
        \centering
        \includegraphics[width=0.24\linewidth]{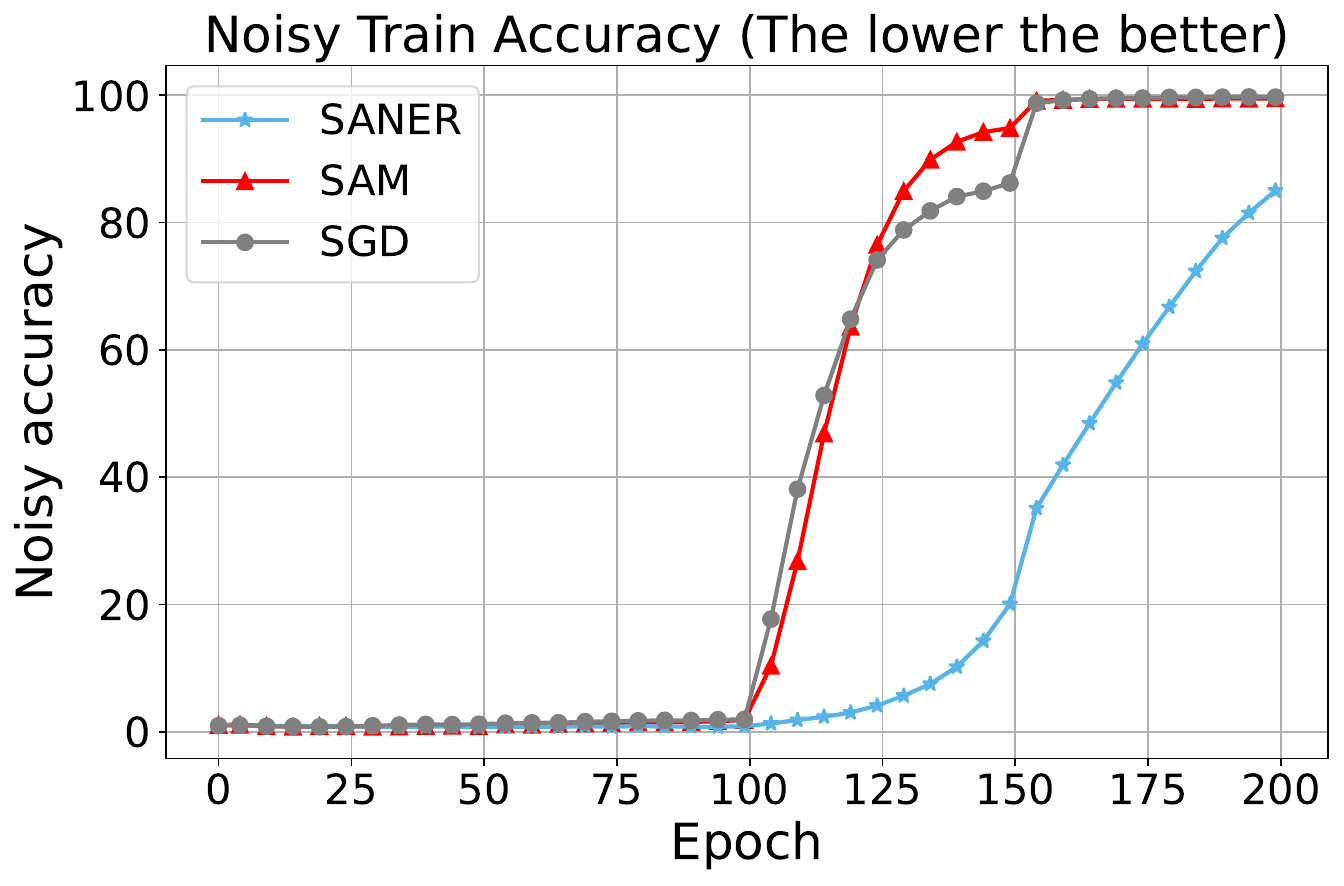}
        \includegraphics[width=0.24\linewidth]{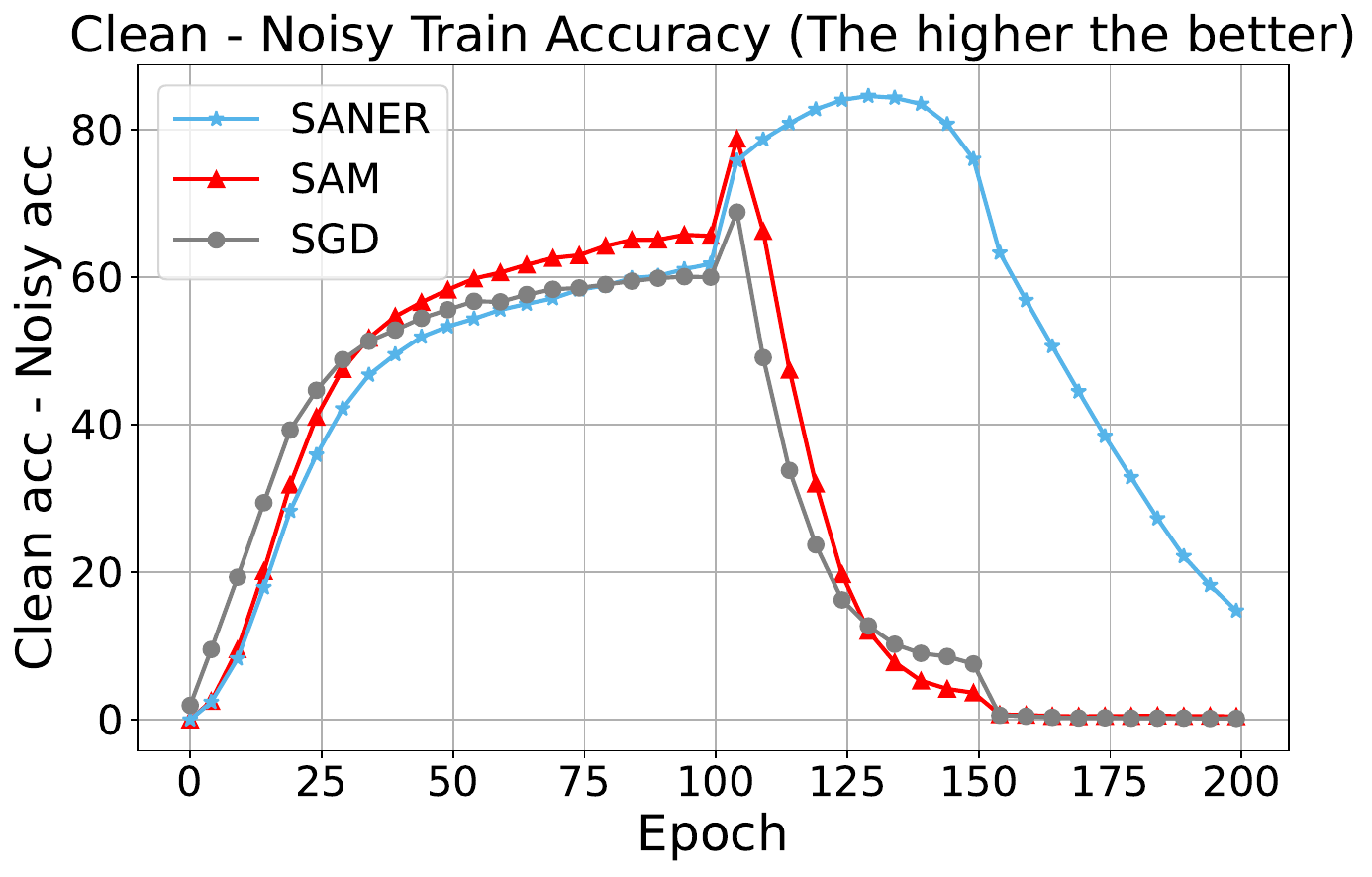}
        \includegraphics[width=0.24\linewidth]{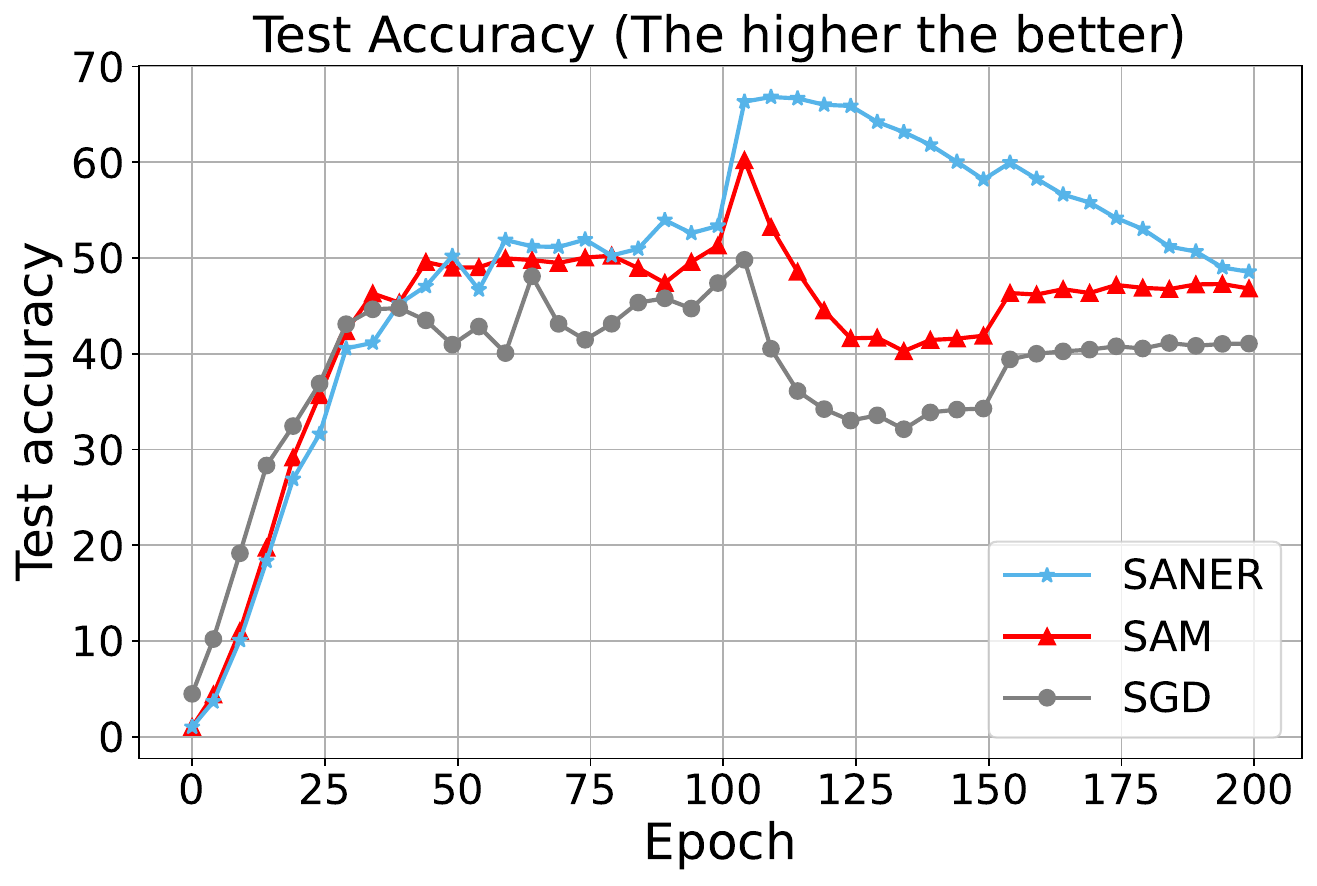}
        \includegraphics[width=0.24\linewidth]{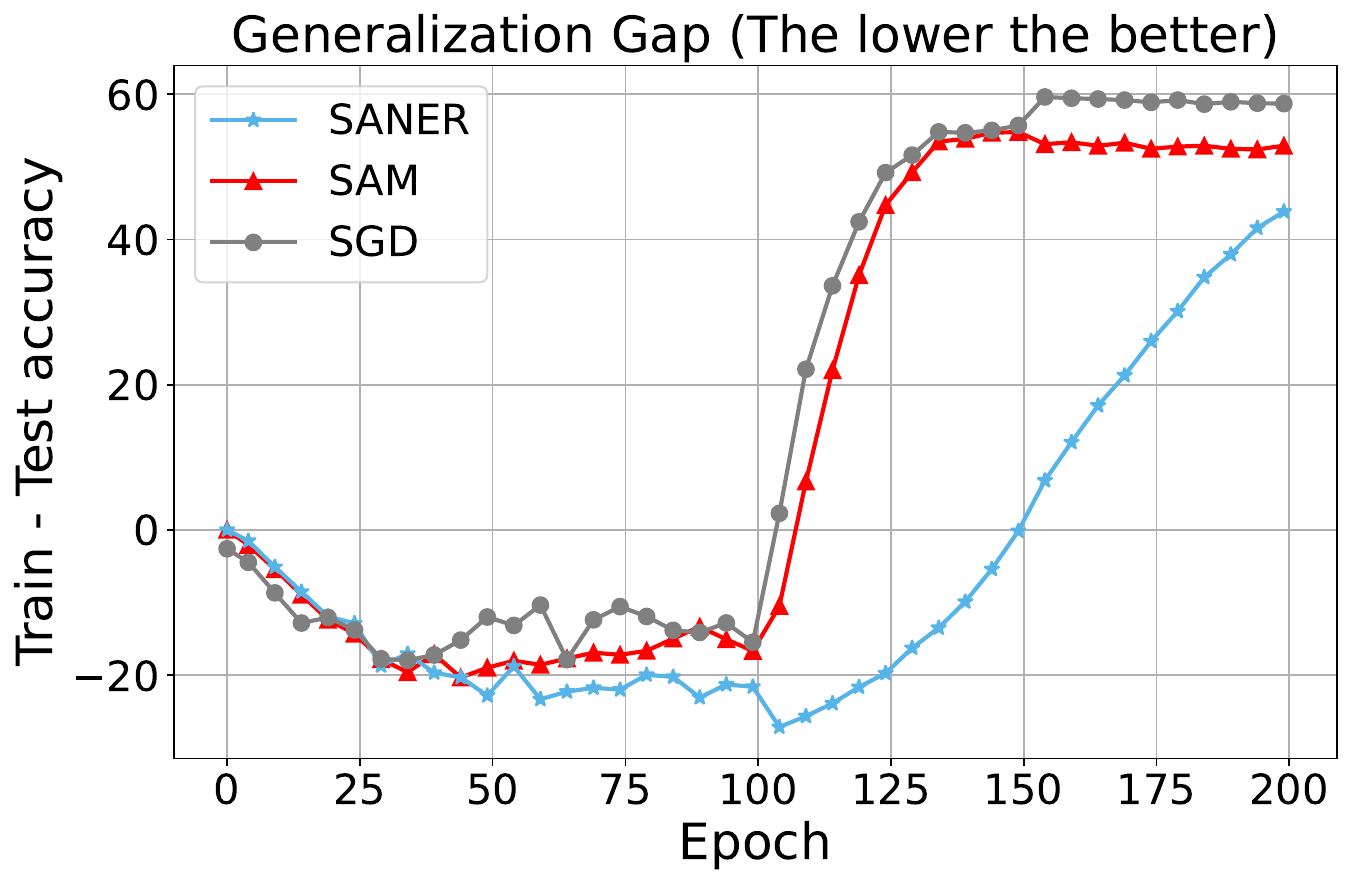}
        \caption{ResNet34}
        \label{fig:sam_vs_sgd_resnet34}
    \end{subfigure}
    \vspace{3pt}
    
    \begin{subfigure}{\linewidth}
        \centering
        \includegraphics[width=0.24\linewidth]{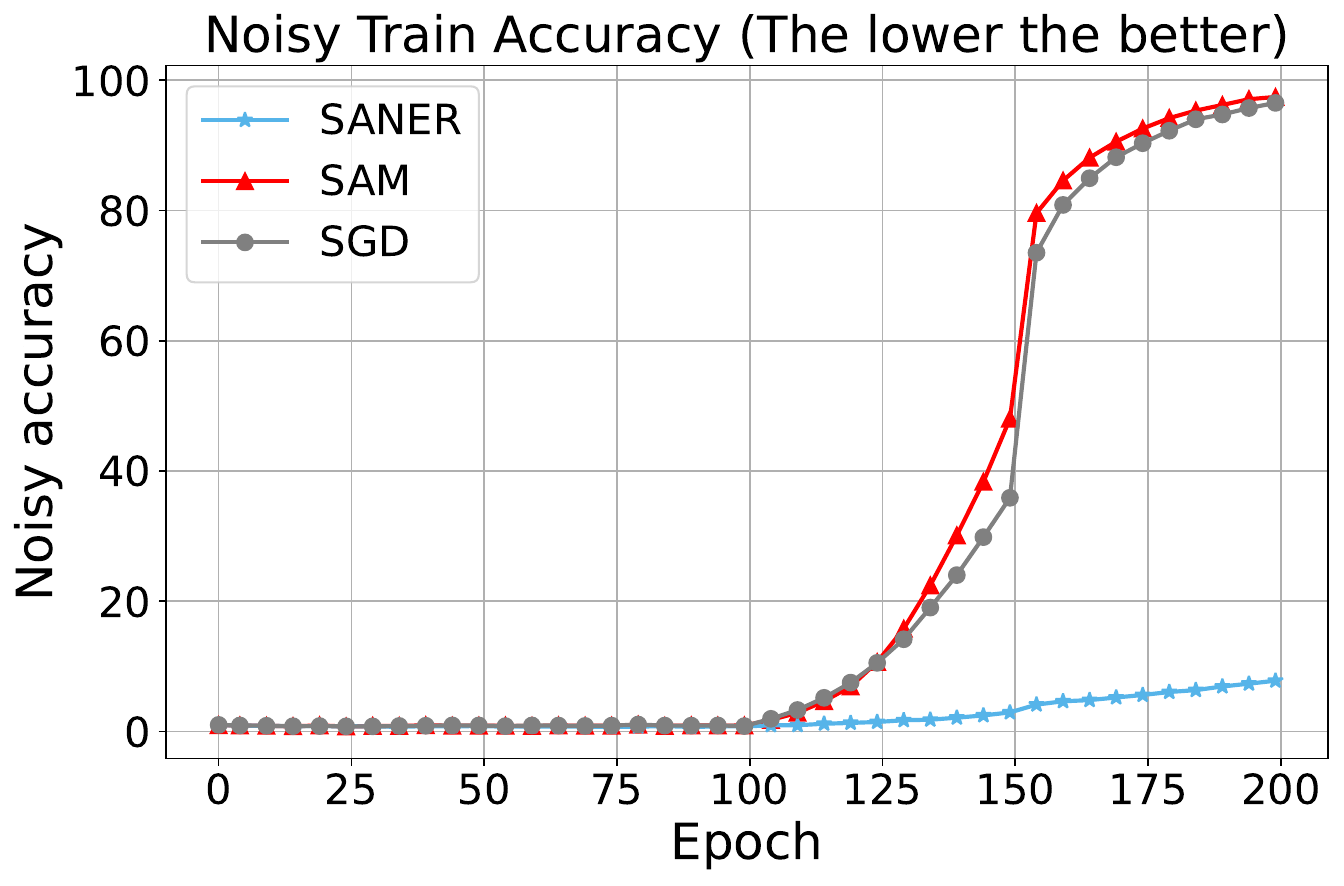}
        \includegraphics[width=0.24\linewidth]{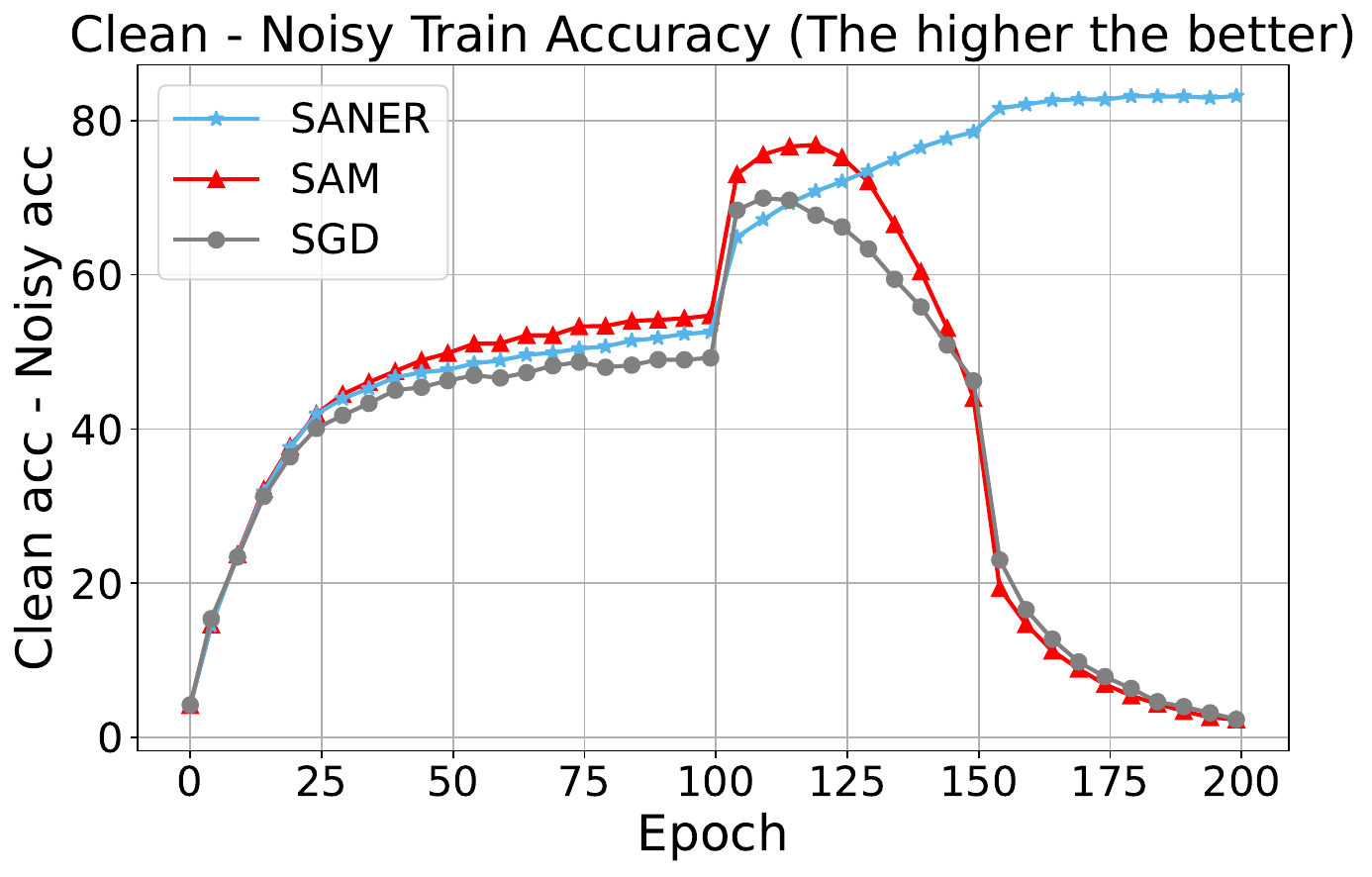}
        \includegraphics[width=0.24\linewidth]{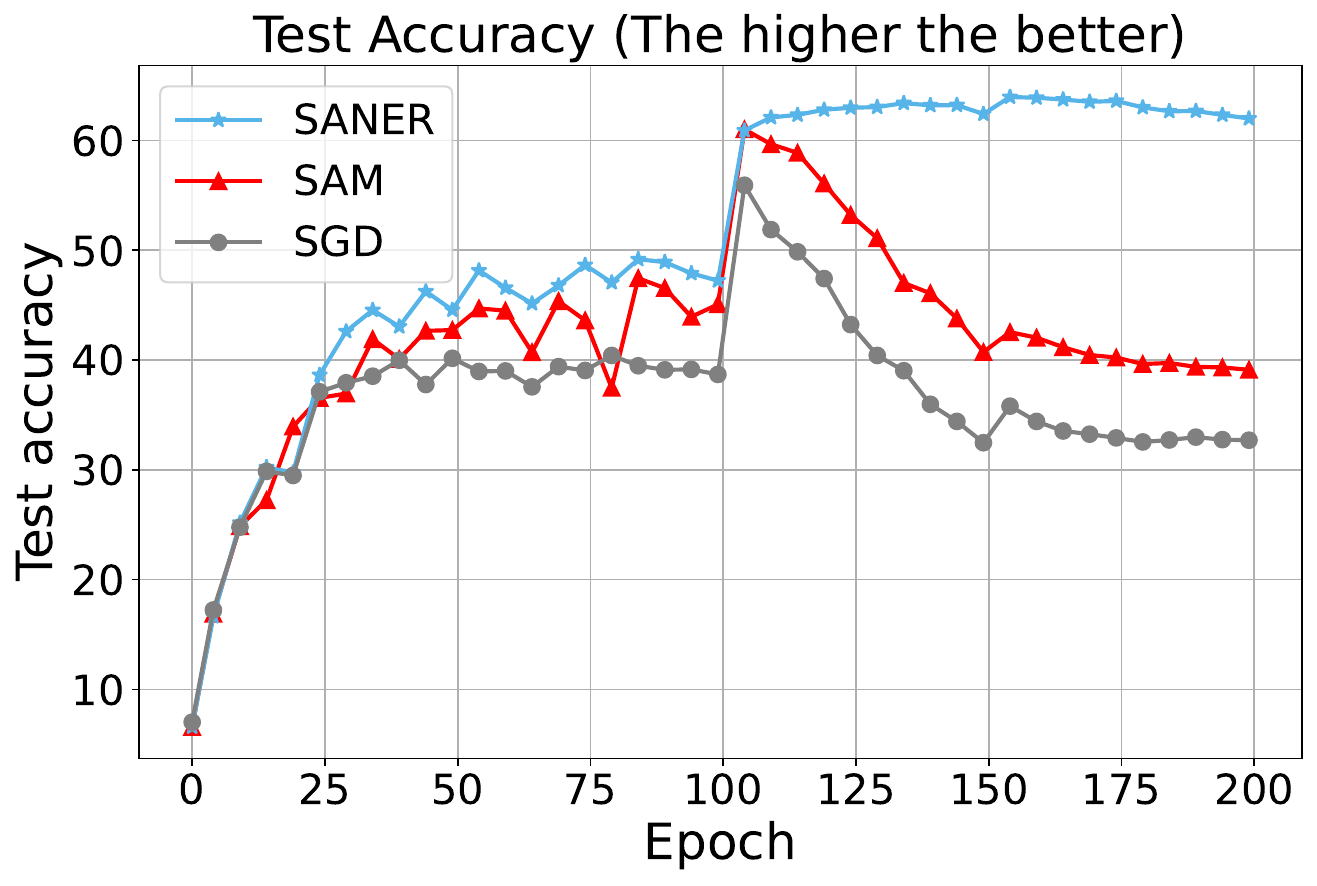}
        \includegraphics[width=0.24\linewidth]{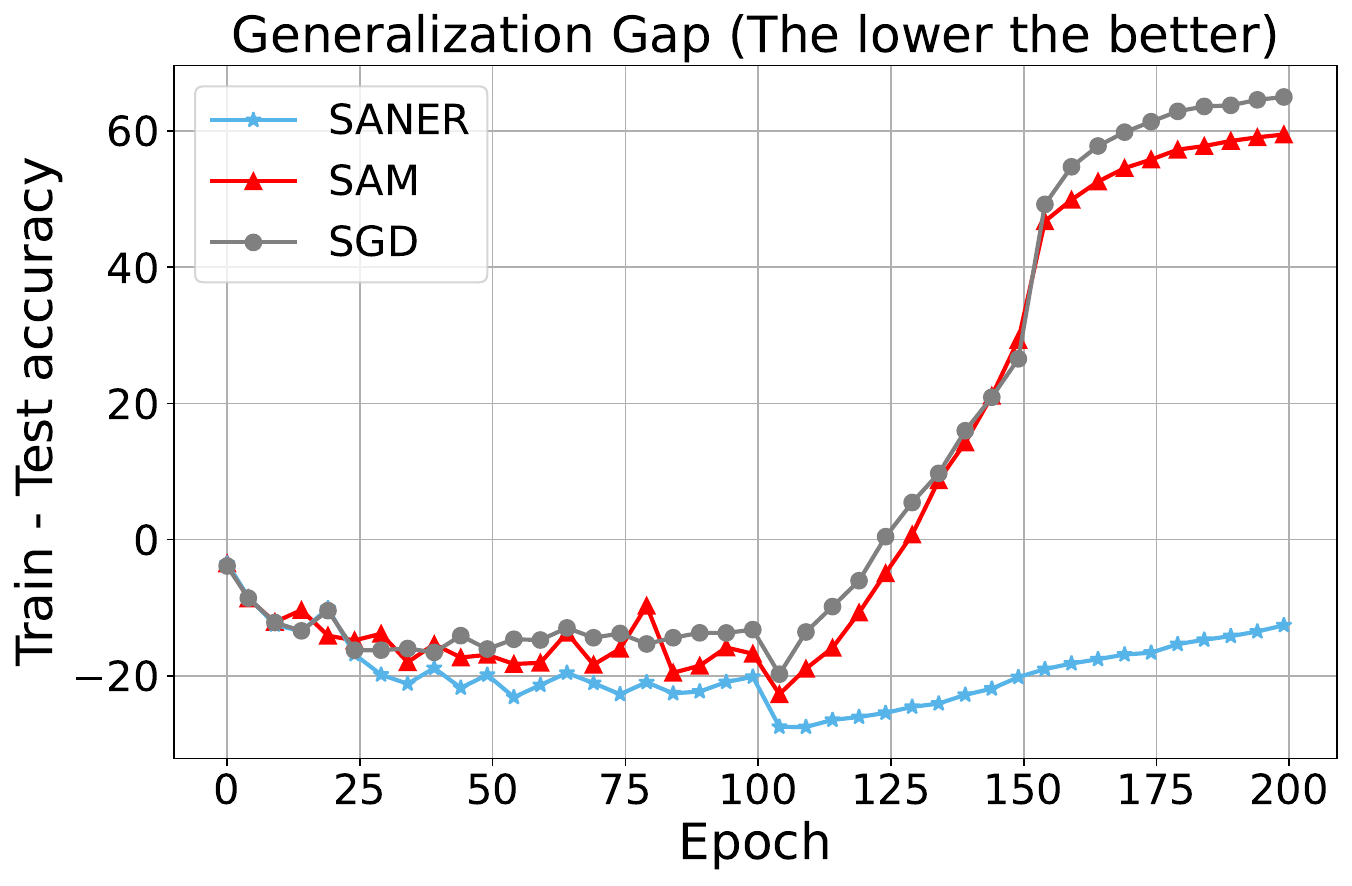}
        \caption{DenseNet121}
        \label{fig:sam_vs_sgd_densenet121}
    \end{subfigure}
    \vspace{3pt}

    \begin{subfigure}{\linewidth}
        \centering
        \includegraphics[width=0.24\linewidth]{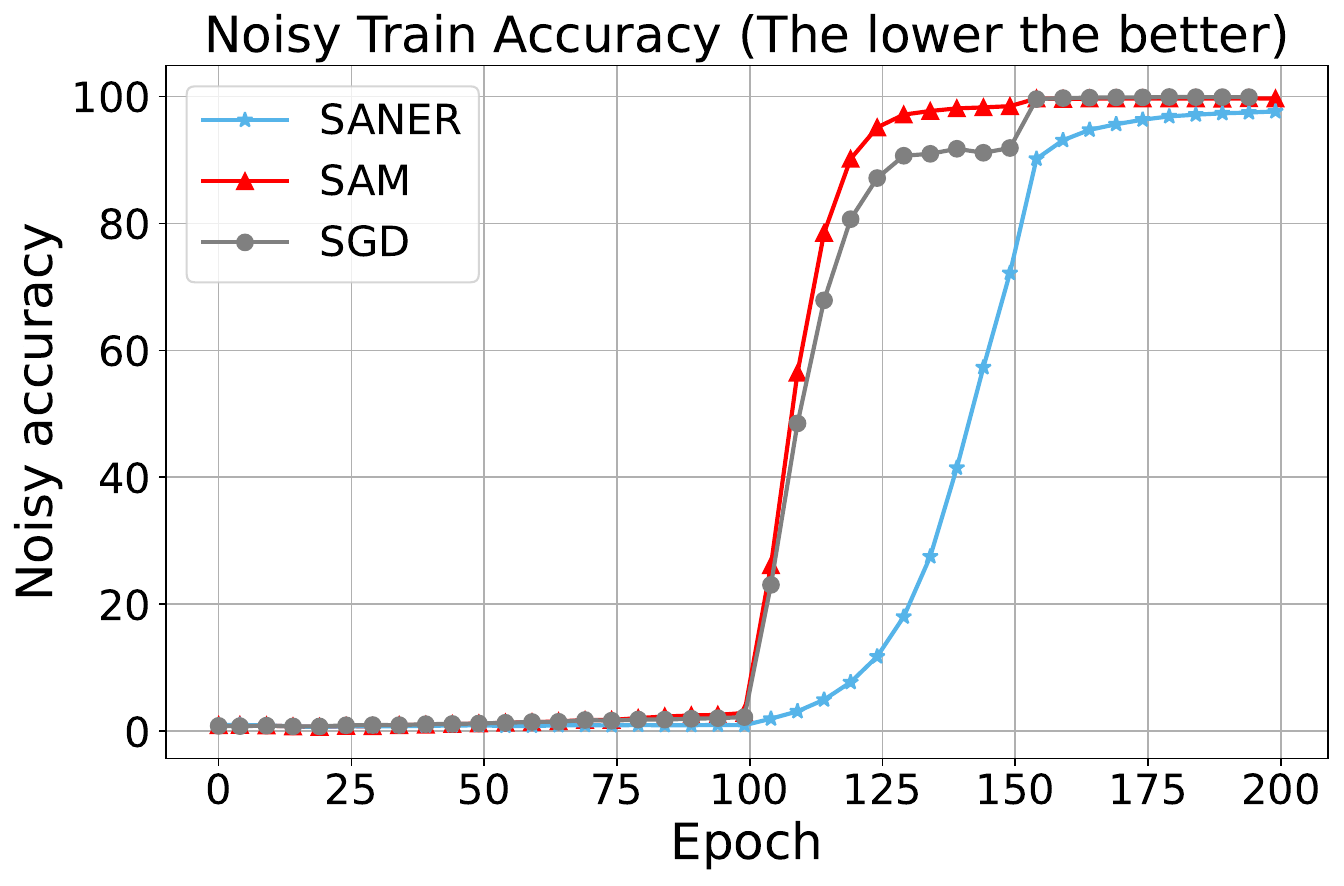}
        \includegraphics[width=0.24\linewidth]{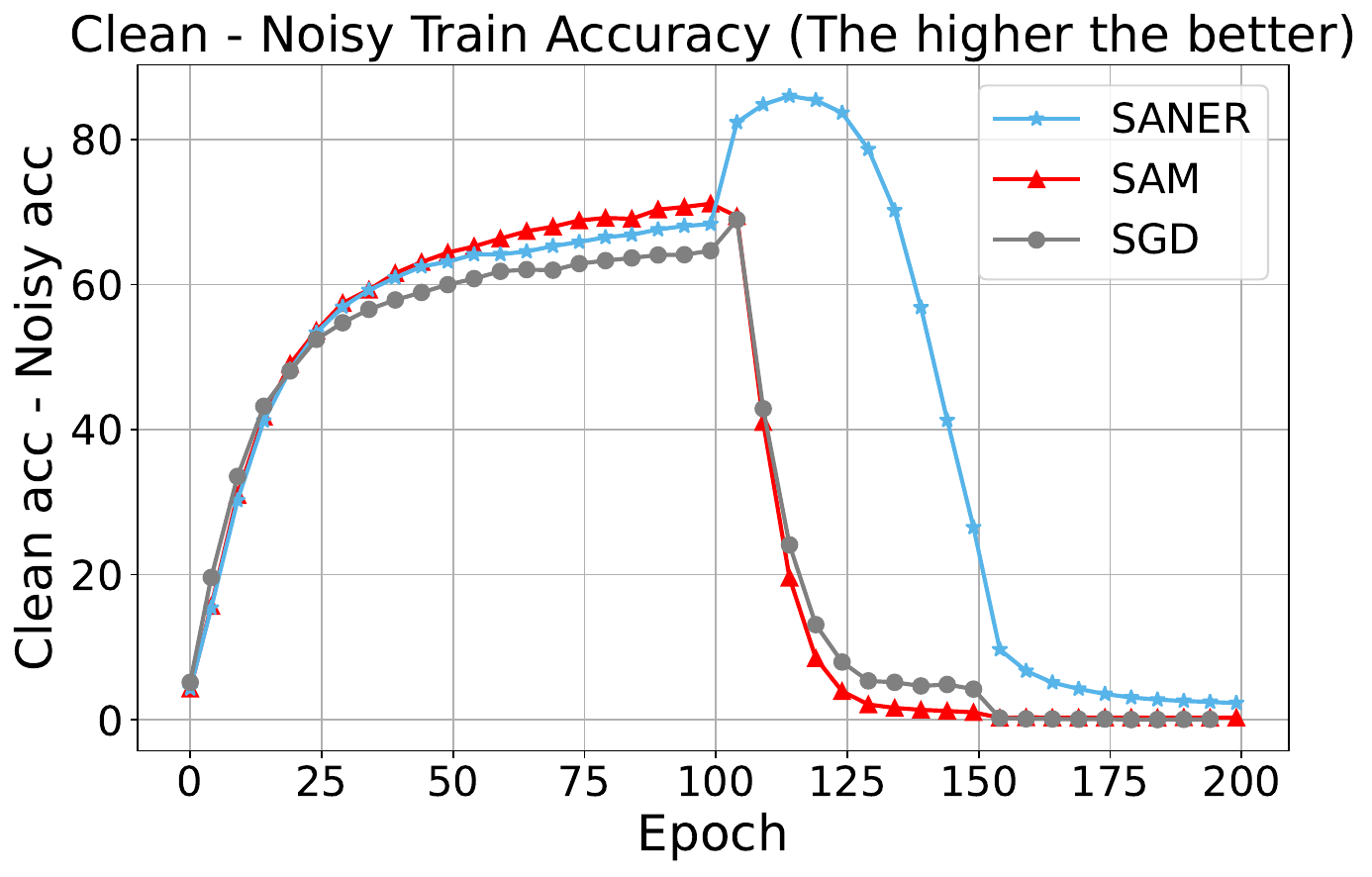}
        \includegraphics[width=0.24\linewidth]{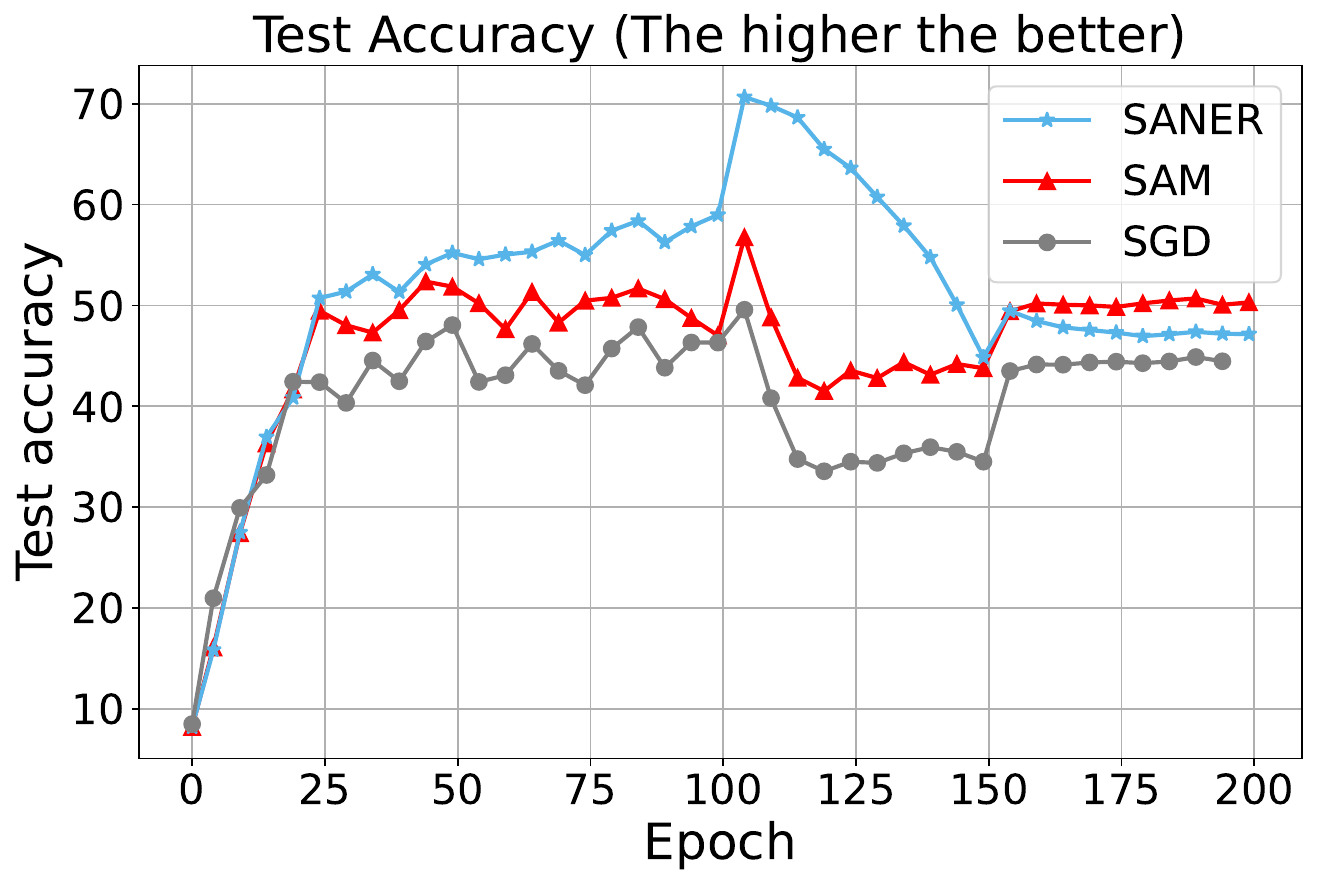}
        \includegraphics[width=0.24\linewidth]{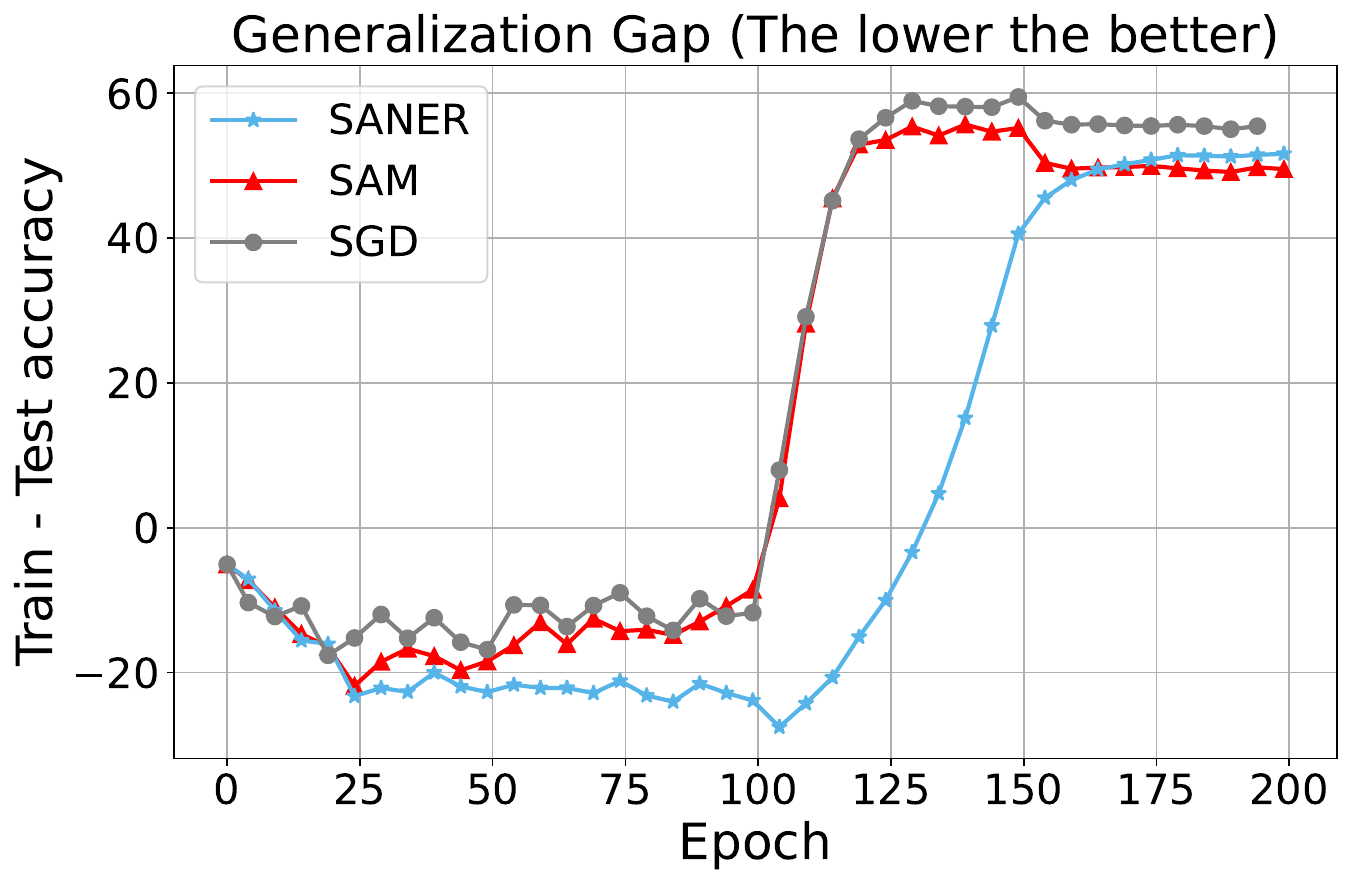}
        \caption{WideResNet28-10}
        \label{fig:sam_vs_sgd_wrn28_10}
    \end{subfigure}

    \caption{Performance comparison of SAM, SGD, and SANER (ours) across \textbf{different models} trained on CIFAR-100 under 50\% label noise. The columns represent (from left to right): noisy training accuracy, gap between clean and noisy accuracy, test accuracy, and generalization gap. SANER outperforms SAM in both noisy accuracy and the clean-noisy accuracy gap, demonstrating better generalization through higher test accuracy.}
    \label{fig:architecture-training-process}
\end{figure*}

To evaluate SANER's robustness across different neural network architectures, we conducted experiments using ResNet34, DenseNet121, and WideResNet28-10 on CIFAR datasets with 25\% and 50\% label noise. We analyze the impact of SANER on the training process, specifically its ability to regulate noisy fitting rates, as shown in Figure~\ref{fig:architecture-training-process}. SANER consistently achieves better control over noisy fitting, thereby reducing overfitting and enhancing generalization performance. These results demonstrate SANER's effectiveness in handling noisy data across diverse architectures, yielding significant improvements over both SGD and SAM.

\end{document}